\documentclass[journal]{IEEEtran} 

\usepackage{amsmath,amsfonts}
\usepackage{algorithmic}
\usepackage{algorithm}
\usepackage{array}
\usepackage[caption=false,font=normalsize,labelfont=sf,textfont=sf]{subfig}
\usepackage{textcomp}
\usepackage{stfloats}
\usepackage{url}
\usepackage{verbatim}
\usepackage{graphicx}
\usepackage{cite}
\usepackage{graphics} 
\usepackage{epsfig}
\usepackage{mathptmx}
\usepackage{times}
\usepackage{amssymb}
\usepackage{float}
\usepackage{arydshln}
\usepackage{multirow}
\usepackage{diagbox}
\usepackage{bbm}
\usepackage{makecell}
\usepackage{lscape}
\usepackage{mathbbol}

\usepackage{hyperref}
\hypersetup{colorlinks,breaklinks,
linkcolor=[rgb]{0.5,0.,0.},
citecolor=[rgb]{0.000,0.427,0.173},
urlcolor=[rgb]{0.031,0.318,0.612}}

\newtheorem{definition}{Definition}

\newcommand{\etal}{\textit{et al}.}
\newcommand{\ie}{\textit{i}.\textit{e}.}
\newcommand{\eg}{\textit{e}.\textit{g}.}

\newcommand{\blfootnote}[1]{%
  \begingroup
  \renewcommand*{\thefootnote}{}% Remove footnote number
  \footnotetext{#1}% Add the footnote text without hyperlink
  \endgroup
}

\usepackage[font=footnotesize]{caption}
\begin{document}

\title{Integrating One-Shot View Planning with a Single Next-Best View via Long-Tail Multiview Sampling}

\author{
Sicong Pan, Hao Hu, Hui Wei, Nils Dengler, Tobias Zaenker, Murad Dawood and Maren Bennewitz
}

\markboth{IEEE Transactions on Robotics. Preprint Version. Accepted October, 2024}{Pan \MakeLowercase{\textit{et al.}}: Integrating One-Shot View Planning with a Single Next-Best View via Long-Tail Multiview Sampling}

\twocolumn[{%
\renewcommand\twocolumn[1][]{#1}%

\maketitle

\vspace{-1.5cm}

\begin{figure}[H]
\hsize=\textwidth
\centering
\includegraphics[width=1.0\textwidth]{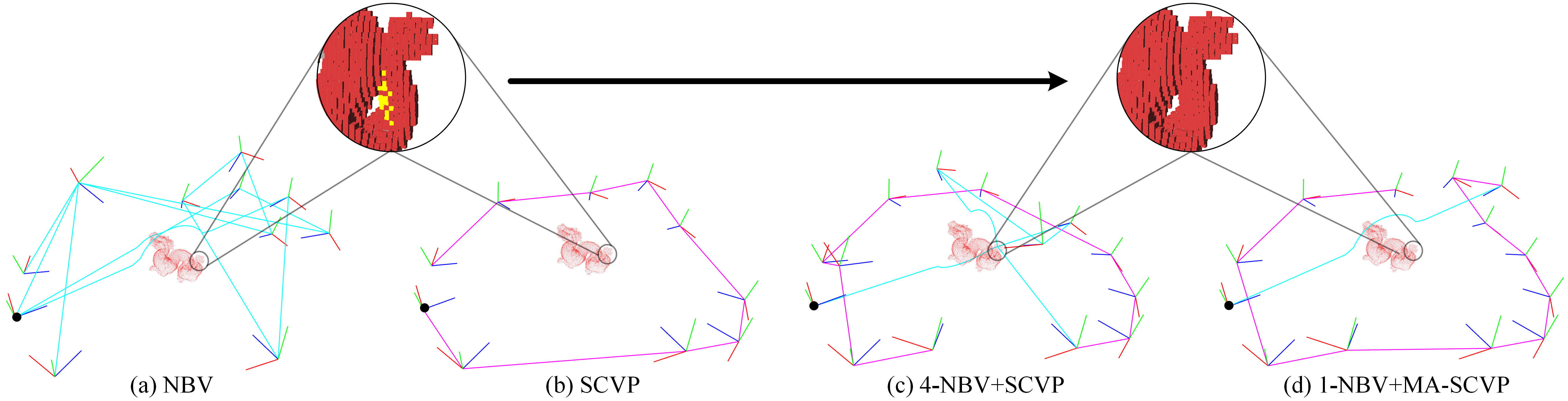}
\caption{
Comparative results of reconstructing an unknown object: Each method is depicted through reconstructed 3D models (red point clouds), local paths (cyan), global paths (purple), views (red-green-blue coordinate systems), and the same initial view (black circle). In (a) and (b), it is observed that both the iterative NBV method and the one-shot set-covering view-planning (SCVP) network fail to provide complete surface details (yellow voxels in enlarged areas). To address these missing surfaces, we introduce a novel combined pipeline that incorporates four NBVs before activating the SCVP network, as shown in (c). However, this approach requires more NBVs and paths (4 NBVs with 4 long local paths). Therefore, we propose the multiview-activated MA-SCVP network, trained on our innovative long-tail multiview dataset as shown in (d), requiring only 1 NBV with 1 local path. Our novel pipeline preserves high-quality reconstruction while reducing the number of necessary NBVs and minimizing path length.
} 
\label{fig_paper_cover}
\end{figure}
}]

\begin{abstract}
Existing view planning systems either adopt an iterative paradigm using next-best views (NBV) or a one-shot pipeline relying on the set-covering view-planning (SCVP) network. However, neither of these methods can concurrently guarantee both high-quality and high-efficiency reconstruction of 3D unknown objects. To tackle this challenge, we introduce a crucial hypothesis: with the availability of more information about the unknown object, the prediction quality of the SCVP network improves. There are two ways to provide extra information: (1)~leveraging perception data obtained from NBVs, and (2)~training on an expanded dataset of multiview inputs. In this work, we introduce a novel combined pipeline that incorporates a single NBV before activating the proposed multiview-activated \mbox{(MA-)SCVP} network. The MA-SCVP is trained on a multiview dataset generated by our long-tail sampling method, which addresses the issue of unbalanced multiview inputs and enhances the network performance. Extensive simulated experiments substantiate that our system demonstrates a significant surface coverage increase and a substantial 45\% reduction in movement cost compared to state-of-the-art systems. Real-world experiments justify the capability of our system for generalization and deployment.
\end{abstract}


\vspace{-0.5cm}

\blfootnote{Manuscript received: December 25, 2023; Revised: October 8, 2024; Accepted: October 16, 2024. This work was supported by the NSFC Project under Grant 61771146 (Corresponding author: Hui Wei).}
\blfootnote{Sicong Pan, Hao Hu, and Hui Wei are with the Laboratory of Algorithms for Cognitive Models, School of Computer Science, Fudan University, China (email: 18210240033@fudan.edu.cn, 20210240286@fudan.edu.cn, weihui@fudan.edu.cn).}
\blfootnote{Sicong Pan, Nils Dengler, Tobias Zaenker, Murad Dawood, and Maren Bennewitz are with the Humanoid Robots Lab, University of Bonn, Germany (email: span@uni-bonn.de, dengler@cs.uni-bonn.de, tzaenker@cs.uni-bonn.de, dawood@cs.uni-bonn.de, maren@cs.uni-bonn.de).}

\IEEEpeerreviewmaketitle

%%%%%%%%%%%%%%%%%%%%%%%%%%%%%%%%%%%%%%%%%%%%%%%%%%%%%%%%%%%%%%%%%%%%%%%%%%%%%%%%

% \vspace{+0.5cm}

\section{Introduction} \label{secI}

\IEEEPARstart{A}{} complete 3D model of an object is important for various autonomous tasks, such as crop monitoring in the agriculture domain~\cite{zaenker2021viewpoint}, pushing in confined spaces~\cite{dengler2023viewpoint}, and grasping in tabletop scenarios~\cite{breyer2022closed}. Facing an unknown environment, active vision systems based on view planning~\cite{chen2011active} are widely studied due to their ability to automatically reconstruct a \textit{prior unknown} object. In this work, we address the view planning challenge in robotic active vision, specifically tailored for achieving both high-quality and high-efficiency 3D reconstruction of unknown objects in tabletop scenarios.

Existing view planning systems fall into either an \textit{iterative} paradigm utilizing the next-best-view~(NBV) evaluation~\cite{zeng2020view} or a \textit{one-shot} pipeline relying on the set-covering view-planning (SCVP) network~\cite{pan2022scvp}. In the iterative paradigm, the robot uses stochastic state analysis~\cite{delmerico2018comparison} or pre-learned knowledge~\cite{zeng2020pc} to plan an NBV after each sensor scan, navigates to that NBV, and repeats this process until a specific stop criterion is met. However, this iterative paradigm lacks global path planning capabilities as shown in Fig.~\ref{fig_paper_cover}(a). Specifically, the robot can only plan a long local path to the NBV~\cite{zeng2020pc} in each iteration. To improve the reconstruction efficiency, we have previously suggested one-shot view planning~\cite{pan2022scvp}, that utilizes SCVP to predict a set of views and compute a global path only once as shown in Fig.~\ref{fig_paper_cover}(b). By accessing all required views at once, the robot performs much faster than iterative systems.

However, neither the iterative nor the one-shot method guarantees a high-quality reconstruction. As shown in the enlarged area in Fig.~\ref{fig_paper_cover}(a) and (b), the 3D models reconstructed by these methods lack surface details that may be crucial features for subsequent tasks, such as grasping and watertight meshing. The insufficiency in surface details within iterative methods primarily arises from the hand-crafted heuristic stop criterion, which determines the required number of views. Due to the resolution constraints, these methods tend to stop prematurely with an insufficient number of views. In contrast, the one-shot SCVP method introduces a built-in auto-stop criterion learned from set covering, which is to find the smallest set of views for full surface coverage. Although SCVP does take into account the stop criterion, the insufficiency in surface details is attributed to its dependence on a singleview input, which provides inadequate information about the object to be reconstructed in an unknown environment.

In the unknown environment, one obvious hypothesis is that the more information you have, the more accurate your predictions will be. Based on this insight, we assume that with the availability of more information about the object to be reconstructed, the prediction quality of the SCVP network improves. In this work, we explore two distinct solutions to provide extra information to SCVP: (1)~\textit{online} leveraging more perception data obtained from NBVs as SCVP input, and (2)~\textit{offline} training on an expanded dataset of multiview inputs. For the first solution, we propose a combined auto-stop pipeline that selects a few NBVs before activating the SCVP network. As shown in Fig.~\ref{fig_paper_cover}(c), this pipeline does increase the surface coverage, but extra NBVs and paths, especially four long NBV paths, are required to finish the reconstruction.

To further improve efficiency while maintaining high-quality reconstruction, we adopt the second solution because multiview inputs contain more surface information. Different from singleview inputs, multiview inputs show an imbalance in importance to network training. We observe a vital phenomenon in terms of the gain of surface coverage: the first 1-3 views cover approximately 80\% of the surface, whereas subsequent views contribute only to the remaining 20\%, resembling a classic long-tail distribution~\cite{zhang2021deep}. Consequently, multiview inputs containing perception data from fewer views should be deemed more important. For instance, a 15-view input case may already cover 97\% surfaces, making it less valuable for training. Therefore, we devise a long-tail sampling strategy to align the number of views in multiview inputs with the long-tail distribution of surface coverage gain. Moreover, we suggest a multiview-activated (MA)-SCVP network architecture, which adds a view state as input, to enhance training on the long-tail multiview dataset. The advantages of employing the long-tail dataset and the MA-SCVP network architecture are validated through experimental studies on network training.

In simulated reconstruction tests, the MA-SCVP network trained on our long-tail dataset exhibits superior overall performance compared to traditional sampling methods~\cite{zeng2020pc,mendoza2020supervised}. Notably, training on our long-tail dataset also improves the surface coverage of the first few NBVs obtained from NBV networks. This enhancement facilitates a more seamless integration of the MA-SCVP with NBV networks. Ablation studies conducted on our combined pipeline reveal that, with the assistance of training on our long-tail multiview dataset, a single NBV is required to provide additional information before activating the MA-SCVP network. Fig.~\ref{fig_paper_cover}(d) illustrates an example of the integration of one-shot view planning with a single NBV through our long-tail multiview sampling, reducing the required NBVs and paths while maintaining high-quality reconstruction.

Extensive simulated experiments provide compelling evidence that our system outperforms both iterative baselines (including the heuristic stop criterion) and the one-shot baseline, yielding a significant increase in surface coverage and a substantial 45\% reduction in movement cost. Subsequently, since the object center and size are not known in advance in a real-world environment, we assess our system using the dynamic object bounding box to justify its generalization capability. In real-world comparative tests, our system consistently validates the claims established in our simulations. Finally, we analyze how our system of providing extra information enhances the reconstruction quality in terms of object test case complexity.

The contributions of this work are summarized as follows:
\begin{itemize}
\item Unlike singleview inputs in the SCVP network, we propose a long-tail sampling method to generate a multiview dataset, offering extra information during the training phase. This long-tail sampling addresses unbalanced importance among multiview inputs, presenting an efficient training dataset compared to traditional sampling.
\item Unlike the heuristic stop criterion commonly employed in NBV methods, our system includes an auto-stop criterion learned from set covering. We propose the MA-SCVP network architecture to train on our long-tail multiview dataset, ensuring a sufficient number of predicted views (stop criterion) for a high-quality reconstruction.
\item We propose a novel combined pipeline that incorporates a few NBVs before embarking on one-shot view planning. Ablation studies confirm that a single NBV is adequate for providing extra information during online data collection. This helps us reduce the onboard resource demands for the required NBVs and paths.
\item We deploy our system on a robotic arm equipped with an RGB-D camera in a real-world environment. Comparative experiments demonstrate the robust generalization capabilities and reconstruction performance of our system. Our system concurrently achieves both high-quality and high-efficiency unknown object reconstruction.
\end{itemize}

To support reproducibility, our implementation is released at \url{https://github.com/psc0628/MA-SCVP}.

% \subsection{Contributions and Outline} \label{secI.A}
% The rest of this paper is organized as follows. Section~\ref{secII} presents related work on active vision systems and view planning algorithms. Section~\ref{secIII} describes the problem formulation and the combined pipeline. The proposed long-tail sampling method with the MA-SCVP network is detailed in Sec.~\ref{secIV}. Section~\ref{secV} describes the simulation and real-world experiments. Finally, we discuss the future work of our study in Sec.~\ref{secVI} and summarize our findings in Sec.~\ref{secVII}.

\vspace{-0.2cm}

\section{Related Work} \label{secII}

Active vision systems and view planning problems for 3D reconstruction, dating back to the 1980s~\cite{connolly1985determination}, have been extensively studied~\cite{tarabanis1995survey,scott2003view,chen2011active} and continue to be a frontier area of robotic research~\cite{zeng2020view,maboudi2022review}. Depending on the moving area of robots, 3D reconstruction targets can be categorized as scenes or objects~\cite{zeng2020view}. Scene reconstruction, also referred to as 3D exploration~\cite{bircher2018receding,monica2019humanoid,song2020online} requires mobile robots to \textit{explore inside} a limited 3D volume and aims to reconstruct large structures. Since mobile robots experience cumulative errors and battery issues during long-distance travel, view planning focuses more on the path planning problems~\cite{song2021view}. On the other hand, object reconstruction~\cite{vasquez2013hierarchical,aleotti2014global,monica2018contour} requires robots to \textit{observe outside} a limited 3D volume and aims to reconstruct a relatively small object with complex surface details. In this study, we concentrate on the reconstruction of small tabletop objects characterized by complex self-occluded surfaces that are an essential consideration for applications like industrial manufacturing~\cite{peng2020viewpoints} and agricultural plant reconstruction~\cite{burusa2022attention}.

\vspace{-0.2cm}

\subsection{Next-Best-View Planning} \label{secII.A}

Early research on view planning methods for object reconstruction used synthesis methods, whereby sensor views are directly chosen based on specific constraints represented by analytical functions~\cite{connolly1985determination,wong1999next}. The synthesis method can be faster than the search-based method, but it lacks the capability to address occlusion issues, which is important for the modeling of complex surfaces. 

Zeng \etal~\cite{zeng2020view} summarized modern search-based view planning methods as a generate-and-test procedure. Since recently more learning-based methods emerged, we broaden this summary to the generalized generate-and-test procedure, as follows: (1)~Generate a candidate view space (a number of candidate views) under some constraints or task requirements, such as the robot workspace and regions of interest~\cite{zaenker2021combining}. (2)~Test the candidate view space via a certain utility function to achieve the goal of view planning.

The fundamental differences between view planning methods lie in the definition of the utility function: goal (output), 3D representation (input), and optimization technique.

\subsubsection{Goal}

Active vision systems are typically designed for autonomous applications in unknown environments~\cite{chen2011active}. An iterative pipeline that selects an NBV maximizing the utility function for each sensor update is therefore commonly used. This makes sense due to the ability to monitor information gains in an unknown environment in real-time. 

In addition, some works are based on multi-robot collaboration~\cite{wu2019plant,lauri2020multi}, which assigns an NBV for each robot in an iteration by overlap awareness. Generally, in active vision, the goal of the utility function is defined as finding the NBV(s), and the view planning methods are also known as iterative NBV planning methods. 

\subsubsection{3D Representation} 

Active vision systems continuously collect detected information about an unknown environment to deal with uncertainties. The information is the representation of the 3D environment and the object. As an input to the utility function, it is clear that the choice of data structures strongly influences the view planning strategy~\cite{zeng2020view}.

\textbf{Point cloud}. Point cloud is the directly accessible data from sensor observations, such as $(x,y,z)$ coordinates and color information of depth sensors~\cite{keselman2017intel}. It is straightforward to restore the surface details of an object~\cite{border2018surface}.

\textbf{Surface}. Surface representation is the common form of 3D models~\cite{wu20153d} in computer graphics. The most popular discrete surface is the triangle mesh~\cite{pito1999solution}. The continuous Poisson field can also be used for view planning~\cite{wu2014quality}. The surface representation can capture fine details but more details have higher computational costs.

\textbf{Voxel}. Voxel representation is the discrete grid of the 3D space, usually stored in an octree or occupancy map~\cite{hornung2013octomap}. Unlike point clouds and surfaces, voxels can describe unknown and empty spaces and are therefore the most commonly used in view planning~\cite{massios1998best}. Some related voxel representations, such as TSDF~\cite{menon2022viewpoint} and surfel~\cite{monica2018surfel}, can be used to extract surfaces for better view planning. The principal disadvantage is its wastefulness of storage.

\textbf{NeRF}. Neural Radiance Field (NeRF) is an implicit representation of 3D space~\cite{mildenhall2021nerf}, stored in the network function by taking a point on a ray direction as inputs and its color value with density as outputs. Researchers have increased interest in using NeRF for view planning~\cite{pan2022activenerf,ran2023neurar,jin2023neu} because it can reduce the storage costs of explicit representations.

\subsubsection{Optimization Technique}

There are two common optimization techniques to define the way to find NBV(s): search and learning. Another important difference is the additional constraints on the candidate view space.

\textbf{Search}. Search-based methods define heuristic functions on the 3D representations and select the best view of maximum utility. The candidate views have no other constraints and can be dynamic. Some of them are random or spherical sampled and can sample a large number while others are defined by heuristic functions and selected by certain requirements such as robot movement cost. Although the 3D representation varies, the heuristic functions have some common high-level features: frontier areas, boundary and shape analysis, and entropy and occlusion awareness.

\textit{For point cloud-based search}, Border \etal~\cite{border2018surface} proposed density representation to define a frontier between fully and partially observed surfaces and later they~\cite{border2020proactive,border2022surface} handle occlusions to select NBVs that most improve an observation. 

\textit{For surface-based search}, Krigel \etal~\cite{kriegel2011surface} purposed boundary and curve estimation of mesh for new surface information. Wu \etal~\cite{wu2014quality} suggested gradient and smoothness confidence to quantify Poisson surfaces and Lee \etal~\cite{lee2020automatic} improved with primitive shape analysis to select NBVs.

\textit{For voxel-based search}, Krainin \etal~\cite{krainin2011autonomous} modeled the information gain in terms of entropy and suggested the boundaries between occupied and unknown voxels. Vasquez-Gomez \etal~\cite{vasquez2014volumetric,vasquez2017view} proposed the area factor for perceiving unknown areas and handling overlap with previous scans. Daudelin and Campbell~\cite{daudelin2017adaptable} proposed to use frontier voxels to model the object probability of belonging to the object surfaces. Delmerico \etal~\cite{delmerico2018comparison} summarized ray casting-based volumetric information gains and proposed rear side entropy, assuming that unknown areas behind the observed object surface are more likely to be the new object surface. Pan and Wei~\cite{pan2022aglobal} proposed a max-flow-based global optimization for volumetric covering of predicted surfaces and later they~\cite{pan2023global} improved NBV selections by generalized maximum coverage.

\textit{For NeRF-based search}, Lee \etal~\cite{lee2022uncertainty} proposed ray-based volumetric uncertainty and S{\"u}nderhauf \etal~\cite{sunderhauf2023density} proposed density-based ensemble uncertainty to select NBVs for better NeRF online learning.

\textbf{Learning}. Diverging from search-based methods, learning-based methods typically acquire the view-planning ability through end-to-end learning from the environment or data.

\textit{For reinforcement learning,} these methods get the reward of the utility of a view directly from the environment and are usually trained in simulation. The candidate views are implicitly defined by states and actions. Peralta \etal~\cite{peralta2020next} proposed the reward function of the percentage of observed surface points over the total number of surface points in the ground truth model. Zeng \etal~\cite{zeng2022deep} proposed the reward function of the coverage of the unknown area for fruit detection and mapping.

\textit{For deep learning,} NBV planning is usually treated as classification problems, \ie, an optimal view class is selected for the current scene, which can be trained by a simulated dataset. The candidate views can be re-scaled with object size but they should be pre-defined, \eg, on a sphere. With the development of voxel-based networks~\cite{zhou2018voxelnet}, Mendoza \etal~\cite{mendoza2020supervised} proposed NBVNet and later they analyzed the NBV regression~\cite{vasquez2021next}. Along with advances in networks of point cloud~\cite{qi2017pointnet}, Zeng \etal~\cite{zeng2020pc} proposed point-cloud-based PCNBV for scoring the views and Han \etal~\cite{han2022double} proposed double-brunch-based DBNBV for ranking the views instead of scoring.

To sum up, in the field of NBV planning, search-based methods are the most developed and straightforward. However, facing the problem of time efficiency due to the potential large search space, a balance between accuracy and efficiency needs to be considered. Therefore, since learning-based methods do not require hand-crafted heuristics and can infer very fast after sufficient training, they are receiving more attention from current researchers. In this work, we also use deep learning techniques for fast inference.

\vspace{-0.2cm}

\subsection{One-Shot View Planning} \label{secII.B}

Deep learning essentially learns prior knowledge from data and stores it in the network function. In particular, the NBV-based networks learn object local geometry and plan a view to locally maximize the coverage of the object surface~\cite{mendoza2020supervised,zeng2020pc,vasquez2021next,han2022double}. However, neural networks can be designed in a certain way to learn the global features of the object geometry. Wu \etal~\cite{wu2019plant} proposed to use a point completion network to predict the shape of the plant and update a voxel grid to select the NBV. Monica and Aleotti~\cite{monica2021probabilistic} proposed a voxel-based network to predict the uncertainty of the environment (surface distribution of the object) and also plan NBV by voxel-based search. These methods confirm the ability of deep learning to learn global information. 

However, these methods use deep learning to learn globally but plan views locally. Why not use networks to learn global information and directly predict a set of global views? This motivation brings us back to the goal of view planning. In addition to active vision systems, view planning is also used for inspection and reverse engineering tasks~\cite{peuzin2021survey} where the environment and the object are known in advance, \eg, given a prior 3D model. The goal is to find a view subset of the minimum size that simultaneously covers all object surfaces. Kaba \etal~\cite{kaba2017reinforcement} formulate this model-based view planning as the set covering optimization problem (SCOP) and solve it through reinforcement learning. The variations of SCOP are further studied to improve the system performance in aerial modeling~\cite{hepp2018plan3d} and industrial manufacturing~\cite{jing2018computational}. 

It seems promising to use deep learning to learn from model-based global view planning and also predict the smallest set of views covering all object surfaces. To achieve this goal, we have previously introduced the SCVP network~\cite{pan2022scvp} and the automatic supervised label generation by solving SCOP in simulation. We use a voxel-based network because it is easier to define the SCOP on structured surfaces (voxels) rather than unstructured surfaces (point clouds). Using deep learning, we bring a novel one-shot method to the view planning problem, \ie, facing partial reconstruction but predicting a set of views from pre-learned global knowledge about object geometry. 

However, when facing an unknown environment, the information about the object remains only partial, which considerably affects the prediction quality of the SCVP. Although SCVP is highly efficient for reconstruction, it struggles to capture sufficient surface details. In this work, we try to answer a crucial question: \textit{How can we enhance the prediction quality of one-shot view planning by providing extra information about the object to be reconstructed?}

We answer this question by analyzing two solutions to providing extra information. One is to \textit{online} provide more perception data from NBVs, \ie, input with more object surfaces, as detailed in Sec.~\ref{secIII}. The other is to \textit{offline} provide more geometric knowledge about the unknown object, \ie, training from the long-tail multiview dataset instead of the singleview, as detailed in Sec.~\ref{secIV}. Extensive experiments detailed in Sec.~\ref{secV} validate that the integration of one-shot view planning with a single NBV through the long-tail sampling results in both high-quality and high-efficiency reconstruction.

\vspace{-0.2cm}

\section{Problem Formulation and System Overview} \label{secIII}

We describe the view planning problem in an active vision system designed for unknown object reconstruction tasks. Subsequently, we formulate the combined pipeline that integrates one-shot view planning with NBVs. Finally, the definition of views and path planning are elucidated.

\vspace{-0.2cm}

\subsection{View Planning for Unknown Object Reconstruction} \label{secIII.A}

The unknown object reconstruction task considered in this work is reconstructing a high-quality 3D model of an object in an unknown and spatially limited 3D space through a sequence of sensor views navigated by a robotic arm. The six-degrees-of-freedom (6-DOF) robotic arm is equipped with an \mbox{RGB-D} camera at its end effector to acquire color point clouds. We adopt an OctoMap $M$ to represent the environment and its occupancy, which is composed of three states of layered voxels, including occupied, free, and unknown voxels~\cite{hornung2013octomap}. Our system constructs the 3D model of the object as a high-resolution color point cloud model $P$. 

This work assumes that one side of the object surface can be seen from the initial view $v_0$ before reconstruction. Therefore, there is a maximum size bound to the estimated object geometry. Note that the size of the object can be arbitrary, as long as it does not exceed the field of view of the camera. The assumption is reasonable in a tabletop environment, where the size of the objects is relatively small~\cite{pan2022scvp}. 

A view in the candidate view space $V\subset\mathbb{R}^3\times SO(3)$ of the eye-in-hand camera is 6-DOF, \ie, the 3D position and the pose. The position is defined on a hemisphere around the placed object, and the pose is pointing to the center of the object, as detailed in Sec.~\ref{secIII.C}. The view planning problem is to generate the optimal sequence of views $V^\ast$ that simultaneously satisfies the following two objectives: reconstruction of a high-quality point cloud model $P$, and efficient observation of these surfaces within short paths.

Existing solutions for the view planning problem include iterative NBV planning and one-shot view planning. Brief definitions of each are provided below:
\begin{definition} \label{def1}
The iterative NBV view planning is to find the NBV $v_i^\ast\in V$, where $i \in \mathbb{N}$ is the current reconstruction iteration. The planned view sequence $V^\ast=\{v_0,v_1^\ast,v_2^\ast,...,v_i^\ast,...\}$ starts with the initial view $v_0$ and appends the NBV $v_i^\ast$ until a stop criterion is met.
\end{definition}
\begin{definition} \label{def2}
The one-shot view planning is to find the smallest subset $V^\ast_{cover}\subset V$ that can ideally cover all the remaining object surfaces, and then compute $V_{path}$ by global path planning on $V^\ast_{cover}$ (detailed in Sec.~\ref{secIII.C}). The planned view sequence $V^\ast=\{v_0,v_1^\ast,v_2^\ast,...,v_{|V_{path}|}^\ast\}$ starts with the initial view $v_0$ and appends views in $V_{path}$, where $|\ |$ stands for the size of the set.
\end{definition}

\vspace{-0.2cm}

\subsection{Combined View Planning System} \label{secIII.B}

To achieve a higher reconstruction quality, in this work, we propose the combined solution for view planning, which is the synthesis of the one-shot solution and the NBV solution above, defined as follows:
\begin{definition}  \label{def3}
The combined view planning is to select a few NBVs until an iteration number $k$ and then compute the global path $V_{path}$ by one-shot view planning. The planned view sequence $V^\ast=\{v_0,v_1^\ast,v_2^\ast,...,v_k^\ast,v_{k+1}^\ast,v_{k+2}^\ast,...,v_{k+|V_{path}|}^\ast\}$ starts with the initial view $v_0$, and appends NBVs and views in $V_{path}$.
\end{definition}

\begin{figure}[!t]
\centering
\includegraphics[width=0.75\columnwidth]{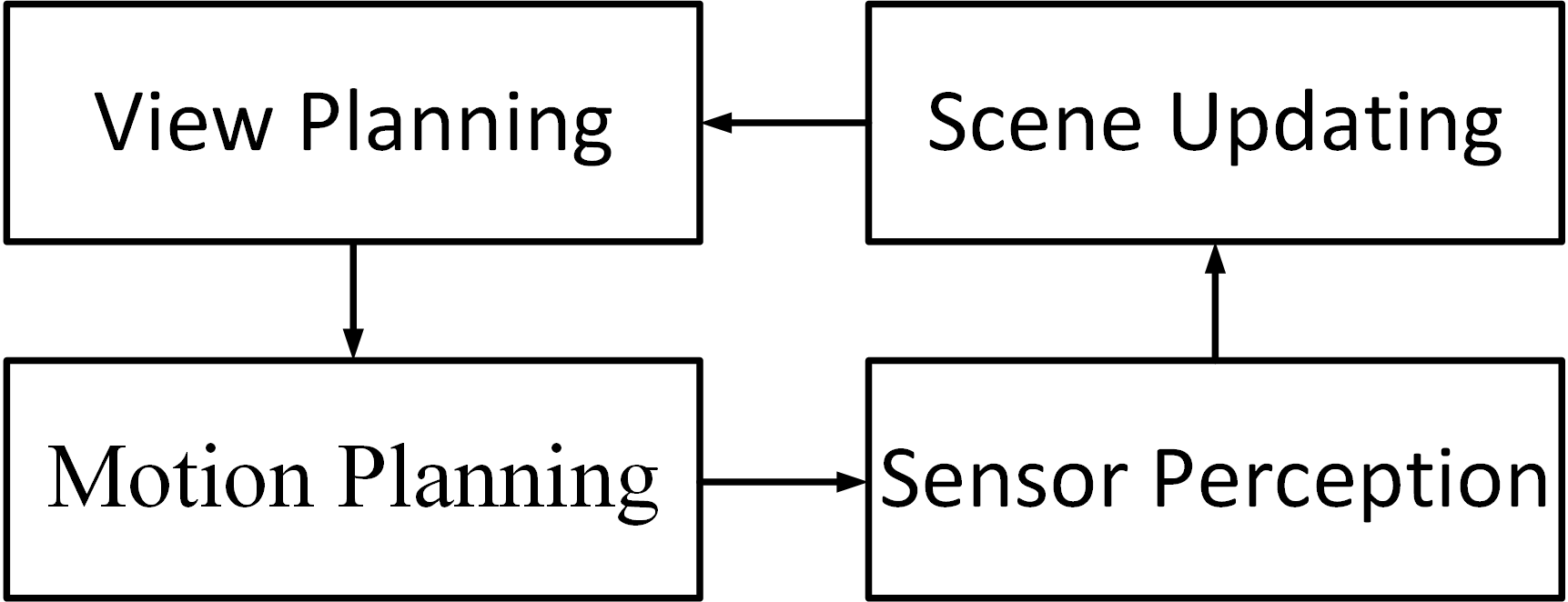}
\caption{
Combined view planning system structure.
} 
\label{fig_active_vision_system}
\vspace{-0.5cm}
\end{figure}

To achieve this combined pipeline, we follow the typical closed-loop control process of the active vision system~\cite{zeng2020view}. As shown in Fig.~\ref{fig_active_vision_system}, the system comprises four modules, which are detailed as follows.

\subsubsection{View Planning}

This module takes the current environment and the candidate view space as input and outputs the NBV or the sequential global view path $V_{path}$, including global path planning. For the initial view $v_0$, it can be a random view of the candidate view space or user-defined. For the NBV planning submodule, a variety of methods can be chosen as described in Sec.~\ref{secV.A}. For the one-shot view planning submodule, the function of our MA-SCVP network is described in Sec.~\ref{secIV}. To illustrate the main workflow of view planning, an example of one NBV ($k=1$) combining our MA-SCVP network is shown in Fig.~\ref{fig_workflow}.

\subsubsection{Motion Planning}

This module navigates the robot to a planned view, including workspace, local path planning, and trajectory planning. The robot workspace contains the areas in which the robot can move, which depend on the robot kinematics equation~\cite{patidar2016survey}. For local view path planning, the obstacle avoidance method is defined in 3D space as described in Sec.~\ref{secIII.C}. For trajectory planning, the point-to-point motion is performed by the kinematics solver~\cite{chitta2016moveit}. 

\subsubsection{Sensor Perception}

This module outputs the registered point cloud from the current view. For RGB-D images from the camera, we use the Default preset and the Align operation~\cite{keselman2017intel} to generate a high-resolution color point cloud. We use the hand-eye calibration~\cite{enebuse2021comparative} to coarsely register the point clouds. Then the iterative closest point algorithm~cite{wang2017survey} is used to finely reduce the error.

\subsubsection{Scene Updating}

This module updates the state of the environment from point clouds, including the 3D representations, the object size and center, and the candidate view space. The OctoMap updates through the default Insertion operation of a point cloud with its view. The $V_{state}$ is a binary vector describing whether a view has been visited or not.

\begin{figure}[!t]
\vspace{-0.5cm}
\begin{algorithm}[H]
\caption{Combined View Planning Method} 
\label{alg1}
\vspace{-0.1cm}
\begin{algorithmic}[1] 
\REQUIRE $\mathit{Initial\ View\ v_0,\ Maximum\ NBV\ Iteration\ k}$
\STATE $V^\ast \gets \{v_0\}$
\STATE $MotionPlanning(v_0)$
\STATE $P_0 \gets SensorPreception(v_0)$
\STATE $P,M,V_{state} \gets \mathit{SceneUpdating(P_0)}$
\FOR{$i$ {\bf from} $1$ {\bf to} $k$}
    \STATE $v_i^\ast \gets \mathit{NBVPlanning(P,M,V_{state})}$
    \STATE $V^\ast \gets \{V^\ast,v_i^\ast\}$
    \STATE $MotionPlanning(v_i^\ast)$
    \STATE $P_i \gets SensorPreception(v_i^\ast)$
    \STATE $P,M,V_{state} \gets \mathit{SceneUpdating(P_i)}$
\ENDFOR
\STATE $V^\ast_{cover} \gets \mathit{f_{MA-SCVP}(M,V_{state})}$
\STATE $V_{path} \gets GlobalPath(V^\ast_{cover} \cup \{v_k^\ast\})$
\FOR{$v_{k+i}^\ast \in V_{path},\ i$ {\bf from} $1$ {\bf to} $|V_{path}|$}
    \STATE $V^\ast \gets \{V^\ast,v_{k+i}^\ast\}$
    \STATE $MotionPlanning(v_{k+i}^\ast)$
    \STATE $P_{k+i} \gets SensorPreception(v_{k+i}^\ast)$
    \STATE $P \gets \mathit{SceneUpdating(P_{k+i})}$
\ENDFOR
\STATE $P \gets BackgroundFilter(P)$
\RETURN $P,V^\ast$
\end{algorithmic} 
\end{algorithm}
\vspace{-0.8cm}
\end{figure}

Algorithm~\ref{alg1} summarizes the auto-stop object reconstruction procedure based on the combined view planning system. The number $k \in \mathbb{N}$ of the maximum NBV iteration is studied in Sec.~\ref{secV.D}. Note that $k=0$ means no NBV selected and performs directly one-shot view planning with the initial view $v_0$. This work assumes that there is only one object to be reconstructed in a tabletop scenario. Therefore, at the end of the reconstruction process, the 3D point cloud model can be obtained by filtering the background of the table (line 20) by the height or a filtering method~\cite{han2017review}. It is worth mentioning that the OctoMap $M$ does not need updates during Scene Updating in the global path execution (line 18) because the 3D model is constructed using point clouds. This reduction in the number of OctoMap updates contributes to the computation speed of our system, as detailed in Sec.~\ref{secV.E}.

\vspace{-0.2cm}

\subsection{View Path Planning} \label{secIII.C}

\begin{figure*}[!t]
\centering
\includegraphics[width=1.0\textwidth]{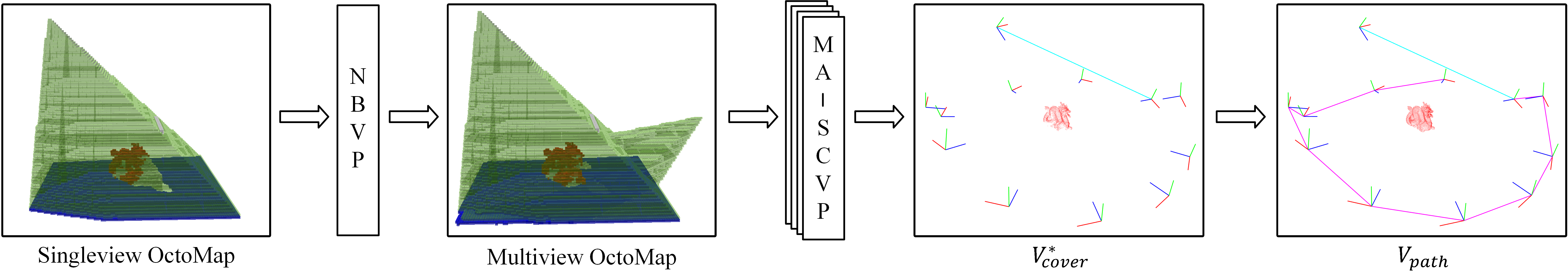}
\caption{
Illustration of the combined view planning pipeline, which performs one NBV planning (NBVP) before activating the MA-SCVP network: The inputs are displayed in OctoMaps to clearly show free areas (green), occupied areas (red object and blue table), and unknown areas (gray voxels near the object). In brief, point clouds and view states are not shown here, which may be required by some view planning methods. The partially reconstructed object models are shown in point cloud space as well as views (red-green-blue), local paths (cyan), and global paths (purple). The output $V^\ast_{cover}$ is the predicted smallest subset covering the remaining object surfaces (views excluding two visited views), which is sequenced to $V_{path}$ by global path planning. 
} 
\label{fig_workflow}
\vspace{-0.5cm}
\end{figure*}

\begin{figure}[!t]
\centering
\includegraphics[width=1.0\columnwidth]{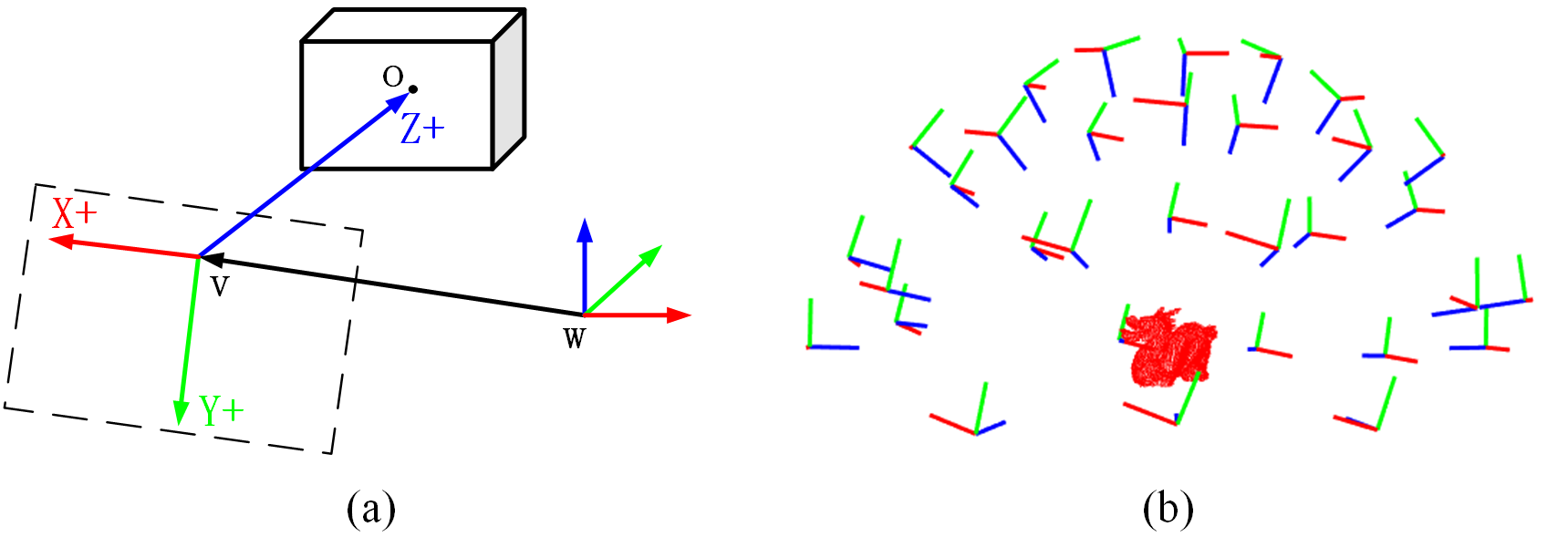}
\caption{
Illustration of candidate views: (a) The 5-DOF view pose. The last DOF can be regarded as rotating the view pose along the Z+ by any degree (rotating the XvY plane, black rectangle). (b) The view space of 32 candidates.
} 
\label{fig_view_space}
\vspace{-0.5cm}
\end{figure}

Here we define the candidate view space before presenting the view path planning. The position of the candidate views is defined by Spherical Codes~\cite{conway2013sphere}, which solves the problem of placing a few points on a sphere to maximize the minimal distance between them. This helps to minimize the intersection of perception on views. Given our focus on the tabletop scenario, only views on the upper hemisphere are considered. Note that in real-world applications, the radius of the view space sphere may vary according to the size of the object because the RGB-D camera needs a certain distance from the object to obtain correct depth images.

The pose of each candidate view is defined as pointing to the center of the object, which is similar to NBVNet~\cite{mendoza2020supervised}. Assume that $o\left(o_x,o_y,o_z\right)$ is the object center position, $v\left(v_x,v_y,v_z\right)$ is the view position, and $w\left(w_x,w_y,w_z\right)$ is the origin of world coordinate system. Thus, we define that Z+ (our camera points along the Z+ axis) is $\vec{vo}$ (units into $r_z$), the Y+ of the view pose is $\vec{vo}\times\vec{wv}$ (units into $r_y$), and the X+ of the view pose is $r_x=r_y\times r_z$. The homogeneous view matrix $H_v$ is defined as:
\begin{equation} 
\label{equ1}
H_v=
\left(
\begin{matrix}
r_x &   r_y &   r_z &   0  \\
0   &   0   &   0   &   1  \\
\end{matrix}
\right)^{-1}
\ast
\left(
\begin{matrix}
I   &   -\vec{wv}   \\
0   &   1           \\
\end{matrix}
\right)
\end{equation}
As discussed in NBVNet~\cite{mendoza2020supervised}, pointing to the center of the object defines the 5-DoF pose, and the last degree of freedom has almost no effect on the reconstruction. We illustrate this 5-DoF definition in Fig.~\ref{fig_view_space}(a) and many 6-DoF poses are valid here. In general, the last degree of freedom can be chosen as the most upward camera pose or a pose with minimal change from the previous pose. It is worth mentioning that in real-world applications, as the reconstruction proceeds, the size and the center of the object change. Therefore, our view pose also adapts to the change, and the effects are studied in Sec.~\ref{secV.F}.

The size $|V|$ of the candidate view space is set to 32 as in our previous work SCVP~\cite{pan2022scvp}. In most cases, having 32 views on the upper hemisphere in a tabletop scenario is sufficient to observe all visible areas. This choice is influenced by the fact that NBVNet~\cite{mendoza2020supervised} assumes 14 views on the upper hemisphere, while PCNBV~\cite{zeng2020pc} assumes 33 views on the whole sphere. We illustrate the candidate view space in Fig.~\ref{fig_view_space}(b).

\begin{figure}[!t]
\centering
\includegraphics[width=1.0\columnwidth]{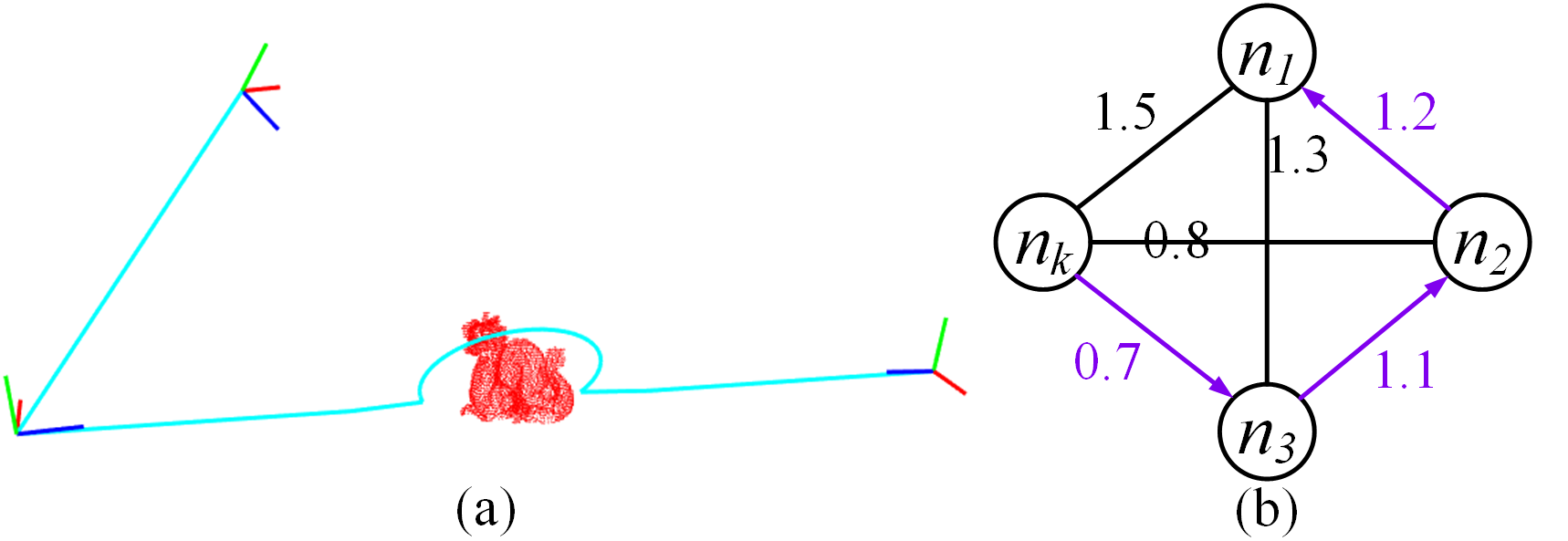}
\caption{
Illustration of the view path planning: (a) local path planning (cyan) of a straight path and an obstacle avoidance path; (b) global path planning (purple) by finding the shortest Hamiltonian path on the undirected complete graph of views.
} 
\label{fig_path_planning}
\vspace{-0.5cm}
\end{figure}

\subsubsection{Local View Path Planning}

A local view path is a path between two views. To navigate the robot arm from one view to another, we need to consider the obstacles of the object and the tabletop. For safety reasons, we define an obstacle as a bounding sphere around the object, where the center is the center of the object and the radius is the size of the object. We can ignore the tabletop because the candidate views are all above it. Formally, the local path length between any two views is given as:
\begin{equation}
\label{equ2}
\resizebox{0.9\hsize}{!}{$
Path(v_i,v_j)=
\left\{
\begin{aligned}
& ||v_i-v_j||^2,\ \ \ \ \ \ \ if\ rt_1\ or\ rt_2\ does\ not\ exist  \\
& ||v_i-rt_1||^2+Arc(rt_1,rt_2)+||rt_2-v_j||^2,\ else
\end{aligned}
\right.
$}
\end{equation}
where $||\ ||^2$ is the Euclidean distance between the 3D positions of two views, ${rt}_1$, ${rt}_2$ are the two intersection points between the obstacle sphere and the 3D line, and $Arc({rt}_1,{rt}_2)$ is the arc length on the sphere. In most cases, the arc is above the object and computes as the spherical distance, \ie, the length of the minor arc of the great circle. In the case of the minor arc below the object, we instead use the major arc to avoid collisions. As shown in Fig.~\ref{fig_path_planning}(a), if there is no obstacle, the local path will be straight, otherwise, it will go around it. We next sample some points on the local path and traverse them with better trajectory planning.

\subsubsection{Global View Path Planning}

The global path consists of some local paths between a set of views. The goal of global path planning is to find the shortest local paths to traverse all the views, formally defined as:
\begin{definition} \label{def4}
Given the set of views $V^\ast_{cover} \cup \{v_k^\ast\}$ and the local path between any two views, the global path planning problem is to find the sequence $V_{path}$, whose total path length $\sum_{i=1}^{|V^\ast_{cover}|}Path\left(v_i,v_{i+1}\right)$ is minimized. Note that the sequence starts from the current view $v_k^\ast$, i.e., $v_1=v_k^\ast$.
\end{definition}

This problem is known as the shortest Hamiltonian path problem, which is similar to the traveling salesman problem but without returning to the starting point~\cite{held1962dynamic}. A Hamiltonian path is a path in an undirected graph that visits each vertex exactly once. Note that here the path is a set of edges in the graph. The shortest Hamiltonian path is a Hamiltonian path with minimized total distance weight of edges.

To define it, we generate an undirected complete graph $G=\left(N,E\right)$, whose vertex $n_i\in N$ is the view $v_i \in V^\ast_{cover} \cup \{v_k^\ast\}$, and whose edge $<n_i,n_j> \in E$ has a distance weight of the local path length $Path(v_i,v_j)$. The global path planning problem is now to find the shortest Hamiltonian path from the vertex $n_k$ ($v_k^\ast$) to another vertex. Fig.~\ref{fig_path_planning}(b) shows an example of the shortest Hamiltonian path between vertices $n_k$ and $n_1$, which has a minimum total distance $3.0$ of edges.

Solving the shortest Hamiltonian path is known to be NP-hard, but state compression dynamic programming~\cite{held1962dynamic} can be used to solve it with time complexity $O(n^2 2^n)$. Since the number of vertices is usually not more than 20, it can be solved in an acceptable time (much less than a second).

\vspace{-0.2cm}

\section{Learning One-Shot View Planning Via Long-Tail Multiview Sampling} \label{secIV}

This section introduces the network function $\mathit{f_{MA-SCVP}}$ in Algorithm~\ref{alg1} (line 12), responsible for predicting the smallest subset $V^\ast_{cover}$ that covers the remaining object surfaces. First, we discuss the long-tail distribution phenomena related to surface coverage gain and propose the long-tail sampling method to create multiview input cases. Second, the SCOP is defined for these input cases to create our long-tail view-planning dataset. Finally, we propose the multiview-activated SCVP network architecture to enhance training performance.

\vspace{-0.2cm}

\subsection{Long-Tail Sampling of Multiview Input Cases} \label{secIV.A}

An input case $(c_{obj},c_{view})$ for our network is defined as a pair of object case and view case, where $c_{obj}=(o,r)$ comprises an object $o$ with its rotation angle $r$, and $c_{view}\in (0,2^{32})$ is an integer. The binary bits of $c_{view}$ indicate whether a candidate view is selected or not. The multiview OctoMap in Fig.~\ref{fig_workflow} illustrates an example of a multiview input case. Specifically, the object case is the Dragon object placed at a certain rotation angle on the tabletop, while the view case consists of two 1 bits representing an initial view and an NBV. The partially reconstructed data that serve as input to our network are determined by this case. In other words, one input case can only generate a partially reconstructed object OctoMap, and vice versa.

\subsubsection{Traditional Input Case Sampling}

\begin{figure}[!t]
\centering
\includegraphics[width=1.0\columnwidth]{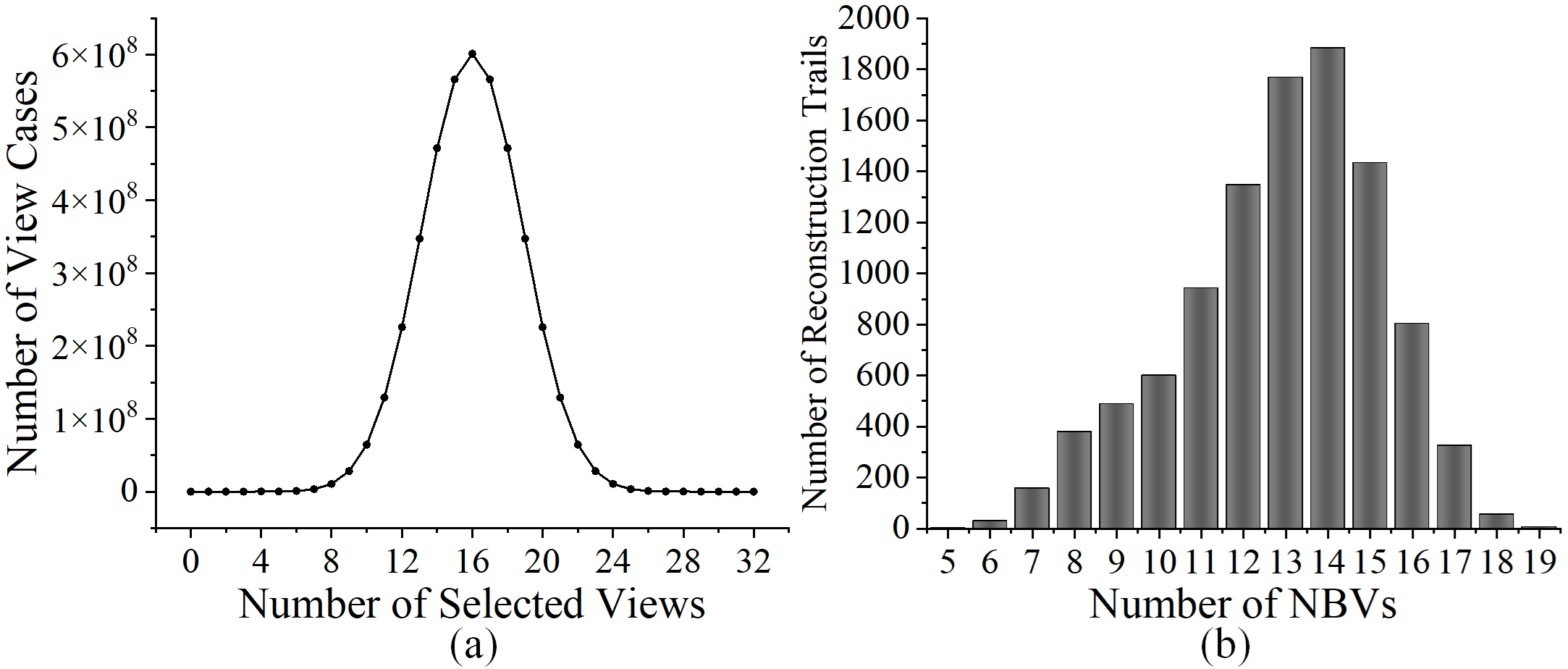}
\caption{
Unsuitability of random sampling: (a) Distribution of the number of view cases over the number of selected views. (b) Distribution of the number of reconstruction trails over the number of NBVs for achieving 100\% surface coverage. These two distributions illustrate that random sampling results in many input cases with already full coverage, rendering them unable to have any views labeled.
} 
\label{fig_multiview_features}
\vspace{-0.4cm}
\end{figure}

Since $c_{obj}$ is easy to determine, a straightforward sampling method is to random sample a set of $c_{view}$. However, this method fails to obtain valuable data for view planning. Fig.~\ref{fig_multiview_features}(a) shows the binomial distribution of the number of $c_{view}$ over the number of selected views in each $c_{view}$ (the total number of 1 bit). Random sampling from this distribution results in a concentration of 14-18 selected views that is enough to cover most of the object surfaces, making the data useless for view planning. In more extreme cases, where 20 or more selected views cover the entire surface, the need for view planning is eliminated entirely, \ie, no views can be labeled for these input cases.

Therefore, instead of direct random sampling, existing NBV networks~\cite{mendoza2020supervised,vasquez2021next,zeng2020pc,han2022double} employ the NBV-reconstruction (NBVR) sampling method. This traditional method performs several simulated NBV reconstruction trials on different $c_{obj}$, recording each $c_{view}$ in each iteration until achieving full coverage of object surfaces. It is reasonable to consider only NBVs in $c_{view}$ because a well-trained network is assumed to work on the input of ideal NBV perception data.

As it guarantees that each input case has at least one view to be labeled, we also follow the NBV reconstruction process to construct our whole sampling space $C_{whole}$ as detailed in Appendix~\ref{Appendix_NBVRsampling}. Along with constructing $C_{whole}$, we depict how many NBVs are required to finish these objects in Fig.~\ref{fig_multiview_features}(b). The results confirm that most objects in reconstruction trials require 12-15 NBVs for 100\% coverage, indicating the unsuitability of random sampling.

\begin{figure}[!t]
\centering
\includegraphics[width=1.0\columnwidth]{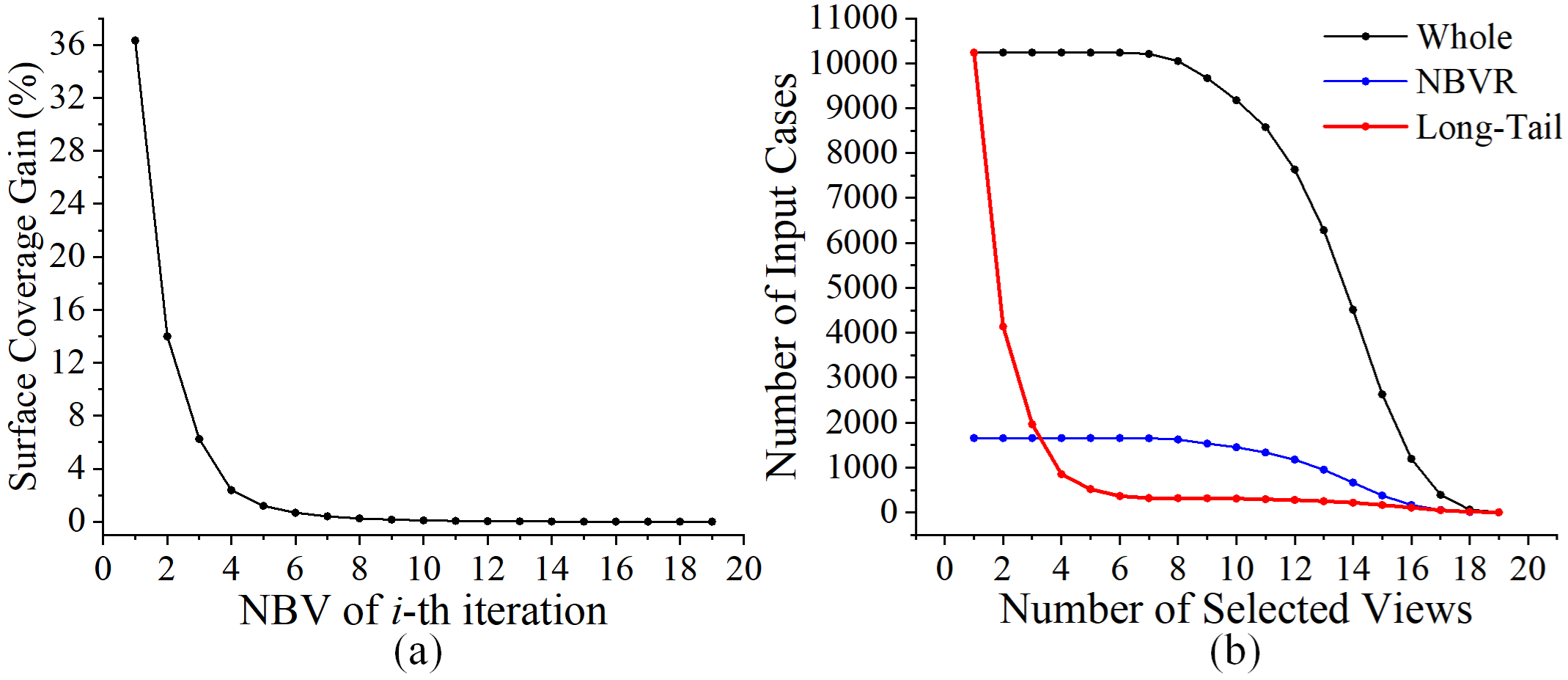}
\caption{
Long-tail phenomenon and sampling methods: (a) Distribution of the surface coverage gain over the NBV corresponding to the $i$-th iteration. Note that the coverage of the initial view (iteration 0) is not relevant and omitted here, which is about 38.02\%. (b) Distributions of the number of sampled input cases over the number of selected views, where Whole stands for the whole sampling space while NBVR and Long-Tail denote two sampling methods. The importance of an input case is assessed by the surface coverage gain of its NBV. Therefore, we design the long-tail sampling method to align with the characteristic long-tail curve shape.
} 
\label{fig_longtail_sample}
\vspace{-0.5cm}
\end{figure}

\subsubsection{Distribution of Surface Coverage Gain}

Although each input case in $C_{whole}$ can be utilized for network training, their importance varies. For instance, a 15-view input case may already cover 97\% surfaces, making it less valuable. Therefore, a natural question arises: \textit{how many NBVs are needed to cover most portion (\eg, 80\%) of the object surfaces?}

Fig.\ref{fig_longtail_sample}(a) calculates the surface coverage gain of the NBV corresponding to the $i$-th iteration, revealing a classic long-tail distribution frequently observed in deep learning~\cite{zhang2021deep}. It illustrates that the first 1-3 views (an initial view and 2 NBVs) cover more than 80\% of the object surfaces, while subsequent views only contribute to the remaining 20\%. Intuitively, an input case with fewer selected views should be considered more important. In other words, we need to prioritize sampling more input cases with fewer selected views.

\begin{figure}[!t]
\vspace{-0.5cm}
\begin{algorithm}[H]
\caption{Long-Tail Input Case Sampling Method} 
\label{alg2}
\vspace{-0.1cm}
\begin{algorithmic}[1] 
\REQUIRE $\mathit{Whole\ Input\ Cases\ C_{whole}}$
\STATE $\mathit{RandomShuffle(C_{whole})}$
\FORALL{$(c_{obj},c_{view}) \in C_{whole}$}
    \STATE $n_{select} \gets \mathit{PopCount(c_{view})}$
    \IF {$|C_{(c_{obj},n_{select})}| < L(c_{obj},n_{select})$}
        \STATE $C_{(c_{obj},n_{select})} \gets C_{(c_{obj},n_{select})} \cup \{(c_{obj},c_{view})\}$
    \ENDIF
\ENDFOR
\FORALL{$c_{obj}$}
    \FOR{$n_{select}$ {\bf from} $1$ {\bf to} $|V|$}
        \STATE $C_{longtail} \gets C_{longtail} \cup C_{(c_{obj},n_{select})}$
    \ENDFOR
\ENDFOR
\RETURN $C_{longtail}$
\end{algorithmic} 
\end{algorithm}
\vspace{-0.8cm}
\end{figure}

\subsubsection{Long-Tail Sampling Method}

We thereby design a long-tail sampling method to make our sampled distribution the long-tailed shape curve. Our method is summarized as Algorithm~\ref{alg2}. $\mathit{RandomShuffle}$ is used to randomly sort the set (line 1). We denote $n_{select}$ as the number of selected views in $c_{view}$, which is easy to obtain by $\mathit{PopCount}$ operation counting 1 bit (line 3). Given the fact that each object case $c_{obj}$ has its own long-tail distribution because objects can vary in size, occlusion, and surface complexity. Thus, we tailor the number of maximum sampled cases $L(c_{obj},n_{select})$ for each object:
\begin{equation}
\label{equ3}
\resizebox{0.9\hsize}{!}{$
L(c_{obj},n_{select}) = 
\left\{
\begin{aligned}
&n_{single}, & n_{select}=1 \\
&\lceil n_{single} \times \frac{NS(c_{obj},n_{select})}{NS(c_{obj},1)} \rceil, & n_{select}\neq1
\end{aligned}
\right.
$}
\end{equation}
where $n_{single} \in [1,32]$ is a user-defined integer of singleview cases to control the total number of samples (by default, $n_{single}=32$ equals the number of candidate views). $NS(c_{obj},n_{select})$ is the surface coverage gain of the object $c_{obj}$ under $n_{select}$-th NBV. The detailed computation of $NS(c_{obj},n_{select})$ can be found in Appendix~\ref{Appendix_NBVRsampling}. We apply the ceiling operation to ensure that there is at least one case for any number of selected views. It is worth mentioning that we generate datasets using different $n_{single}$ as it changes the long-tailed shape, as detailed in Sec.~\ref{secV.B}.

To illustrate our long-tail sampling method, we plot the number of sampled input cases over the number of selected views shown in Fig.~\ref{fig_longtail_sample}(b). The black curve represents the distribution of $C_{whole}$, where it gradually goes to zero due to the different number of NBVs required for full coverage. The red curve illustrates our distribution of $C_{longtail}$, characterized by a long-tail shape. Additionally, the blue curve shows the distribution of $C_{nbvr}$ from the NBVR sampling method, sharing a similar shape with the $C_{whole}$ as it uses uniform sampling. We validate the superiority of our long-tail sampling over the NBVR sampling method in Sec.~\ref{secIV.B} and \ref{secIV.C}.

\vspace{-0.2cm}

\subsection{Multiview Dataset Generation through SCOP} \label{secIV.B}

\begin{figure}[!t]
\vspace{-0.5cm}
\begin{algorithm}[H]
\caption{View Planning Dataset Generation} 
\label{alg3}
\vspace{-0.1cm}
\begin{algorithmic}[1] 
\REQUIRE $\mathit{Long\text{-}Tail\ Input\ Cases\ C_{longtail}}$
\FORALL{$(c_{obj},c_{view})\in C_{longtail}$}
    \FORALL{$v \in V$}
        \STATE $\mathbb{v} \gets VirtualImaging(c_{obj},v)$
        \STATE $U \gets U \cup \mathbb{v}$
    \ENDFOR
    \STATE $V_{state} \gets Decompress(c_{view})$
    \STATE $M \gets Initialize\ with\ Tabletop$
    \FORALL{$v=1 \wedge v \in V_{state}$}
        \STATE $M \gets Insert\ with\ \mathbb{v}$
        \STATE $U_{cover} \gets U_{cover} \cup \mathbb{v}$
    \ENDFOR
    \STATE $U_{rest} \gets U \setminus U_{cover}$
    \FORALL{$v \in V$}
        \STATE $\mathbb{v}_{rest} \gets \mathbb{v} \setminus U_{cover}$
        \STATE $\mathbb{V}_{rest} \gets \mathbb{V}_{rest} \cup \mathbb{v}_{rest}$
    \ENDFOR
    \STATE $V_{label}^\ast \gets \mathit{SetCoveringOptimizer(U_{rest},\mathbb{V}_{rest})}$
        \STATE $Save(M,V_{state},V_{label}^\ast)$
\ENDFOR
\end{algorithmic}
\end{algorithm}
\vspace{-0.8cm}
\end{figure}

Given multiview input cases from $C_{longtail}$, our goal is to automatically generate the view planning dataset by solving SCOP and considering sim-to-real gaps.

\subsubsection{Automatic Dataset Generation}

\begin{figure*}[!t]
\centering
\includegraphics[width=0.85\textwidth]{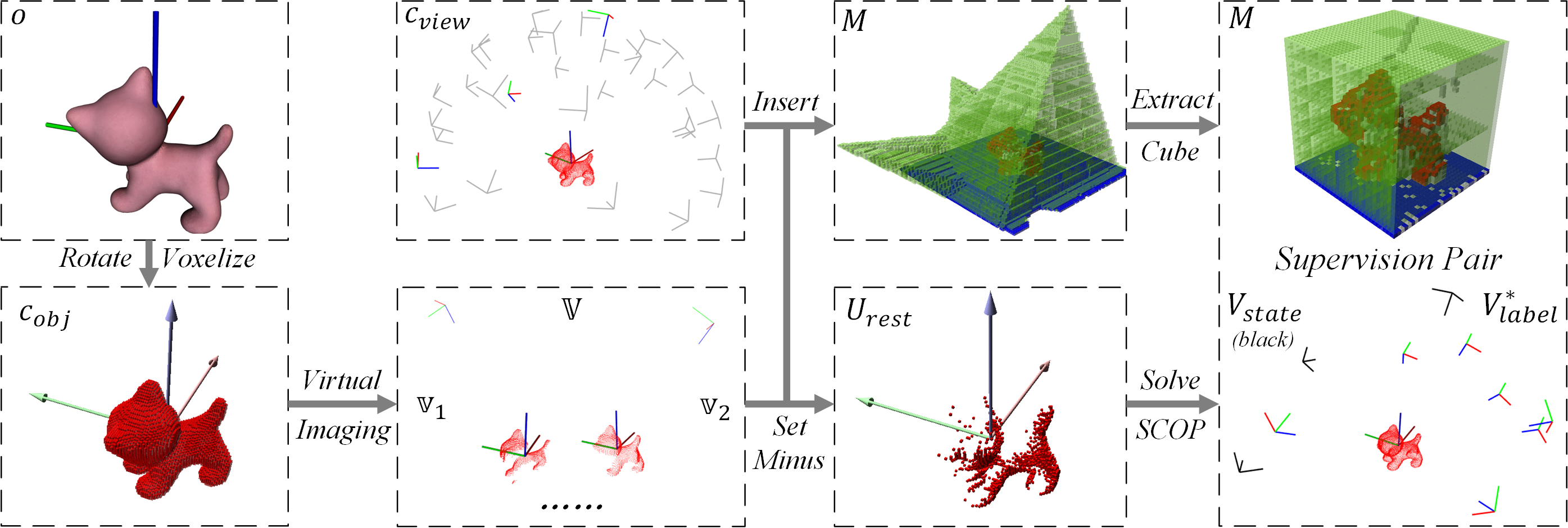}
\caption{
Illustration of the visual steps of view planning dataset generation for set covering. $c_{obj}$ is generated from the mesh model $o$ and a rotation angle $r$. $c_{view}$ is the selected views in the candidate view space and decompressed to $V_{state}$. Taking a certain input case of $(c_{obj},c_{view})$, the collection $\mathbb{V}$ of all view sets can be obtained by virtual imaging. The input OctoMap $M$ is the squared grid obtained from the observations of the selected views in $V_{state}$. The smallest set $V^\ast_{label}$ is solved by SCOP, covering the rest universe set $U_{rest}$ obtained by removing the covered voxels of selected views in $V_{state}$. 
} 
\label{fig_dataset}
\vspace{-0.5cm}
\end{figure*}

Algorithm~\ref{alg3} summarizes the automatic dataset generation process, which is visualized in Fig~\ref{fig_dataset}. The virtual imaging is performed for each candidate view to obtain the set $\mathbb{v}$ of object voxels observed from a view $v$ (line 4) and the set $U$ of all visible object voxels (line 5). The view state $V_{state}$ is easy to obtain by decompressing the bits of $c_{view}$ (line 6). The OctoMap $M$ for the network input is initialized to unknown voxels and inserted with a set of occupied voxels representing the tabletop (line 7). Then the OctoMap $M$ is inserted with the observed voxels from each selected view in $V_{state}$ (line 9). $U_{cover}$ can be obtained by the union of all observed object voxels (line 10). Next, the rest universe set $U_{rest}$ is all the uncovered object voxels removing those observed voxels (line 12). A collection of rest view sets $\mathbb{V}_{rest}$ is the union of each view set $\mathbb{v}$ removing those observed voxels (lines 14-15). Finally, the smallest set $V^\ast_{label}$ is solved by the set-covering optimizer (line 17), and the supervision pair $(M,V_{state},V_{label}^\ast)$ is stored in the view planning dataset (line 18). Note that Algorithm~\ref{alg3} is designed to generate a view planning dataset for set covering, but it is easy to modify lines 12-17 by $v^\ast \gets \arg \max_{v} |\mathbb{v} \setminus U_{cover}|$ to generate a view planning dataset for NBV planning. 

\subsubsection{Labeling through SCOP} 
\begin{figure}[!t]
\centering
\includegraphics[width=0.55\columnwidth]{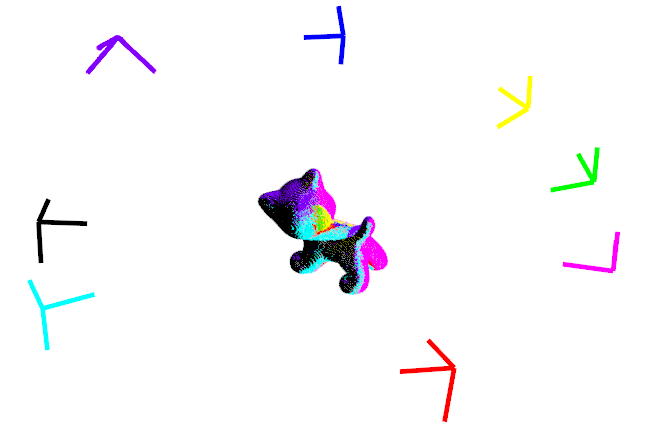}
\caption{
Set covering for view planning: colors represent views and the surfaces they cover.
} 
\label{fig_set_covering}
\vspace{-0.5cm}
\end{figure}

The function of the set-covering optimizer (line 17) is to find the smallest set $V^\ast_{label}$ that fully covers the remaining object surfaces $U_{rest}$. To give an intuitive understanding of the smallest set of views, Fig.~\ref{fig_set_covering} illustrates the surfaces of the Cat object, representing coverage from different views with distinct colors. It shows that eight views are required to cover all object surfaces. For example, if the black and red views are selected in the input case, the $V^\ast_{label}$ then becomes the set of six remaining views. The problem of finding the smallest set $V^\ast_{label}$ is transformed into a class of set covering optimization problems (SCOP):
\begin{definition} \label{def5}
Given the universe set $U_{rest}$ of all uncovered object voxels and the collection $\mathbb{V}_{rest}$ of all view sets removing observed voxels, SCOP for labeling one-shot view planning is to find the sub-collection of $\mathbb{V}_{rest}$ with the smallest number $w$ of view sets, whose union equals the universe $U_{rest}$.
\end{definition}

For instance, consider the universe set $U_{rest}=\left\{1,2,3,4,5\right\}$ and the collection $\mathbb{V}_{rest}=\{\mathbb{v}_1=\left\{1,2,3\right\},\mathbb{v}_2=\left\{2,3,4\right\},\mathbb{v}_3=\left\{1,4,5\right\}\}$. Clearly, the union of $\mathbb{V}_{rest}$ is $U_{rest}$ ($\mathbb{v}_1 \cup \mathbb{v}_2 \cup \mathbb{v}_3=U_{rest}$). However, we can cover all voxels with a smaller number of sets, $w=2$:$\{\mathbb{v}_2,\mathbb{v}_3\}$. Thus, $V^\ast_{label}=\{v_2,v_3\}$ can be obtained. 

SCOP has a long history in operations research~\cite{vazirani2013approximation}. It is an NP-hard problem and is usually defined as an integer linear program (ILP) expression:
\begin{equation}
\label{equ4}
\begin{aligned}
minimize:& \sum_{\mathbb{v}\in\mathbb{V}_{rest}} z_\mathbb{v} \\
\mathit{subject\ to}:&\ (a)\ \sum_{\mathbb{V}_{rest}:e\in\mathbb{v}} z_\mathbb{v}\geq1\ for\ all\ e\in U_{rest} \\
            &\ (b)\ z_\mathbb{v}\in\{0,1\}\ \ \ \ \ for\ all\ \mathbb{v}\in\mathbb{V}_{rest}
\end{aligned}
\end{equation}

The objective function $\sum_{\mathbb{v}\in\mathbb{V}_{rest}} z_\mathbb{v}$ is to minimize the number of chosen view sets. It is subject to (a) each voxel element $e\in U_{rest}$ must be covered by at least one chosen view that can observe this voxel ($e\in \mathbb{v}$), and (b) each view set is either in the collection of chosen sets or not.

The ILP expression of SCOP makes it solvable by a linear programming solver. We choose Gurobi Optimizer to solve SCOP, which takes the lead in the benchmark of linear programming solvers~\cite{mittelmann2018latest}. Since the problem size of SCOP is not too large (the number of voxels and views is not too many), Gurobi can solve SCOP in an acceptable time for labeling (less than 10 seconds).

\begin{figure}[!t]
\centering
\includegraphics[width=0.75\columnwidth]{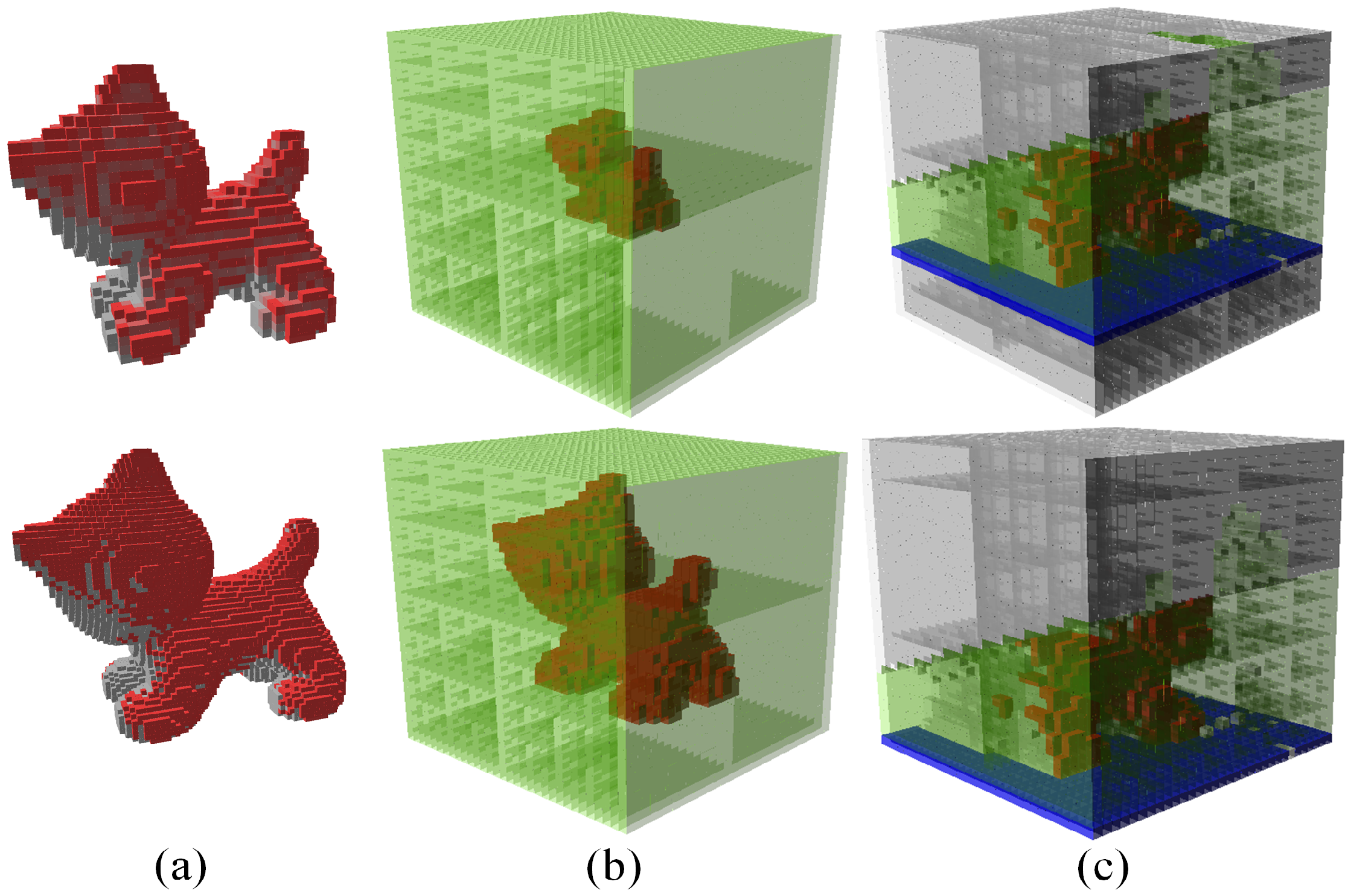}
\caption{
Sim-to-real gap improvements: (a) Imaging resolution 0.004\,m (top) \textit{vs}. 0.002\,m (bottom). (b) Fixed resolution (top) \textit{vs}. Dynamic (bottom). (c) Grid of the object center (top) \textit{vs}. Bottom of the tabletop (bottom).
} 
\label{fig_real_to_sim_gap}
\vspace{-0.5cm}
\end{figure}

\subsubsection{Sim-to-Real Gaps} To bridge the gap between the input cases in the simulation and the real-world situation, three issues should be considered: 

\textbf{Resolution for virtual imaging.} The resolution of the OctoMap for the virtual imaging determines the performance of SCOP because the $U_{rest}$ and $\mathbb{V}_{rest}$ depend on the voxel size. In general, the higher the resolution (the smaller the voxel size), the better the surface details as well as the more views in $V^\ast_{label}$. It is natural to use a high resolution since our view planning goal is to achieve an object model with complete surfaces. However, there is some noise in real point clouds from an RGB-D camera whose resolution is about 0.001\,m. Reducing the resolution of OctoMap can effectively reduce the noise in the point cloud. Therefore, we choose a balanced OctoMap resolution of 0.002\,m for virtual imaging instead of 0.004\,m in SCVP. As shown in Fig.~\ref{fig_real_to_sim_gap}(a), higher resolution OctoMap can capture more surface details.

\textbf{Different sizes of objects.} The size $o_{size}$ of an object is defined as the radius of the minimum bounding sphere of the object. In SCVP, we resized objects from 3D model datasets to a maximum size of 0.10\,m and used a fixed resolution of 0.00625\,m of the OctoMap $M$ for the network input (note that this is not the OctoMap used for virtual imaging). However, some objects with a small original size (such as 0.05\,m) can only occupy half of the OctoMap. In this study, we adopt a dynamic resolution of $M_{res} = 2 \times o_{size} / 32$. The resolution is calculated this way because the OctoMap $M$ is eventually divided into a cube grid whose edge length is twice the size $o_{size}$ of an object. As shown in Fig.~\ref{fig_real_to_sim_gap}(b), the small object will lose surface details in a fixed resolution of OctoMap. Furthermore, this allows our network inputs to no longer be limited by the size of the object.

\textbf{Tabletop.} In real-world situations, an object is placed on a tabletop. To combine the OctoMap $M$ with the tabletop information, we insert a tabletop plane immediately below the object. In SCVP, the center of the divided grid of the OctoMap $M$ is set to the object center. However, the tabletop will have an unfixed position since objects vary in height. In this work, we move down the grid center to make sure the tabletop is fixed at the bottom of the grid as shown in Fig.~\ref{fig_real_to_sim_gap}(c).

\vspace{-0.2cm}

\subsection{Multiview-Activated SCVP Network Architecture} \label{secIV.C}

\begin{figure*}[!t]
\centering
\includegraphics[width=1.0\textwidth]{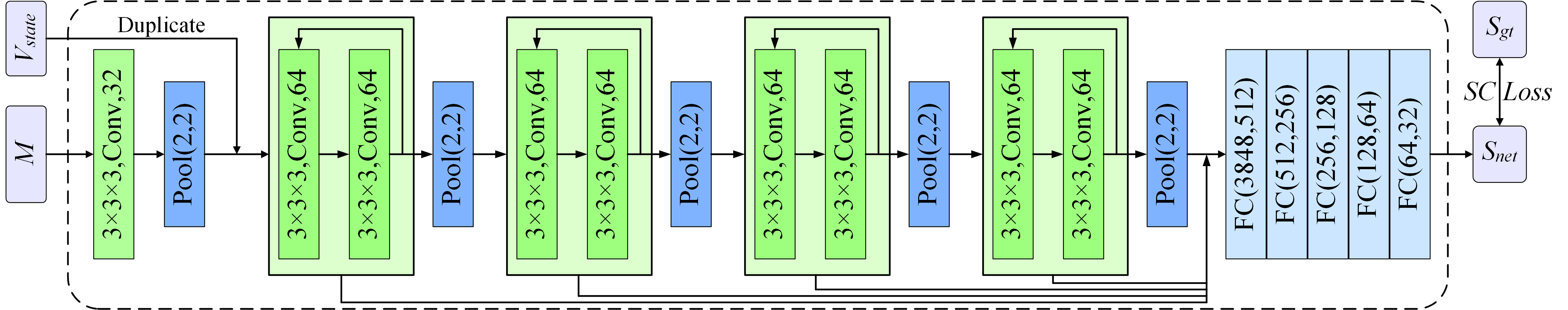}
\caption{MA-SCVP network architecture: Conv block denotes 3D convolution operation with Batch Normalization (BN)~\cite{ioffe2015batch} and Leaky Rectified Linear Units (ReLU)~\cite{maas2013rectifier}. Pool denotes Max-Pooling. FC denotes fully connected layers. The intersections of these arrows denote the concatenate operation of features.}
\label{fig_MA-SCVP_network}
\vspace{-0.5cm}
\end{figure*}

The function of the multiview-activated MA-SCVP network is a classic multilabel classification~\cite{herrera2016multilabel} function:
\begin{equation}
\label{equ5}
\mathit{f_{MA-SCVP}(M,V_{state})}: \mathbb{R}^{32\times32\times32} \times \{0,1\}^{32} \rightarrow \{0,1\}^{32}
\end{equation}
This function takes a $32\times32\times32$ occupancy grid and a vector of 32 bits as input and predicts a vector of 32 bits so that the $V_{cover}^\ast$ can be obtained.

We follow the setting in NBVNet~\cite{mendoza2020supervised} to construct the shape size of our input and output. Since convolution is easier to operate on data with equal dimensions, we extract a $32\times32\times32$ cubic bounding box of the object from the OctoMap $M$. We adopt the dynamic resolution of $M$ so that the shape size of $32\times 32\times 32$ is sufficient to generalize to different object sizes. In the real world, we assume $o_{size}$ is not greater than 15 cm, which can be predicted by solving the minimum bounding sphere (bounding box) in the point clouds to obtain the $M$. The view state vector $V_{state}$ is the same size as our candidate view space. A bit in our network output is bound to a certain candidate view because the SCOP is solved in such a fixed candidate view space. The challenge of a fixed view space with varying rotations of objects is addressed through training based on multiple rotations $r$.

\subsubsection{View State and MA-SCVP Network}

To learn from multiview input data, it is necessary to handle the view state vector input to the deep network. The view state inputs information about previously visited views into the network, reducing the likelihood of predicting views that have already been visited. As demonstrated in PCNBV~\cite{zeng2020pc}, using the view state as input leads to better predictions of views.

Fig.~\ref{fig_MA-SCVP_network} shows the architecture of the MA-SCVP network. After extracting features from the original grid once (1 Conv), we duplicate the view state vector 32 times before the concatenate operation to make the features structured, which is similar to PCNBV~\cite{zeng2020pc}. Then four convolution-based blocks are used to extract features at different resolutions. We apply residual learning to each block (2 Conv blocks have two skip connections) because it will be easy to optimize as shown in the literature~\cite{he2016deep}. Before putting the features into five fully connected layers, we concatenate the features extracted at four resolutions so that they will have a multi-resolution horizon. In particular, the 3D features are concatenated at channel dimension and then flattened to fully connected layers. Finally, the final score for each candidate view is computed into a vector of 32 real numbers, $S_{net}$.

\subsubsection{Loss Function}

The \textit{SCLoss} function is proposed to calculate the difference between $S_{net}$ and $S_{gt}$. The $S_{gt}=\{s_i|s_i=1\ if\ v_i\in V_{label}^\ast\ or\ s_i=0\ if\ v_i\notin V_{label}^\ast\}$ is the binary ground truth mask of $V^\ast_{label}$. We define the \textit{SCLoss} with aid of the cross-entropy loss term widely used in deep learning:
\begin{equation}
\label{equ6}
\begin{aligned}
CE_i &= s_i^{gt}logs_i^{net}+(1-s_i^{gt})log(1-s_i^{net})
\end{aligned}
\end{equation}
where $s_i^{net} \in S_{net}$ is the predicted score and $s_i^{gt} \in S_{gt}$ is the ground truth mask. As shown in SCVP~\cite{pan2022scvp}, the 1's and 0's in the ground truth mask $S_{gt}$ have different importance to the network training. We use a hyper-parameter $\lambda$ to balance them:
\begin{equation}
\label{equ7}
\begin{aligned}
SCLoss=-\frac{1}{32}\sum_{i}^{32}[(1-s_i^{gt})\times CE_i + \lambda \times s_i^{gt}\times CE_i]
\end{aligned}
\end{equation}
The value of $\lambda$ depends on the distribution of 1's in the ground truth masks of a certain view planning dataset.

\subsubsection{Network Output}

The output of the network is a vector of 32 bits computed by a parameter $\gamma$. For each view score $s_i\in S_{net}$, we mark the view score greater than $\gamma$ as 1, otherwise 0. Formally, the final predicted view subset $V_{cover}^\ast$ is defined as $\{v_i|s_i>\gamma \wedge v_i\in V\}$. It is natural to set $\gamma=0.5$ as a common confidence threshold for classification problems since we have no prior knowledge before training. Moreover, we experimentally validate parameters $\lambda$ and $\gamma$ in Appendix~\ref{Appendix_hyperparameter}.

\vspace{-0.2cm}

\section{Experimental Results} \label{secV}

Extensive simulation experiments are designed to support our claims that: (1)~the long-tail sampling method enhances both the network training (Sec.~\ref{secV.B}) and the reconstruction test (Sec.~\ref{secV.C}); (2)~the MA-SCVP network trained on our long-tail multiview dataset enhances the reconstruction quality with less number of required NBVs compared to the SCVP network trained on the singleview dataset (Sec.~\ref{secV.D}); (3)~our system achieves concurrently high-quality and high-efficiency reconstruction of unknown objects compared to state-of-the-art systems (Sec.~\ref{secV.E}). Next, we provide real-world experiments to validate the generalization and deployment of our system (Sec.~\ref{secV.F}). Furthermore, we analyze object test case complexity to elucidate how providing extra information can enhance the reconstruction quality (Sec.~\ref{secV.G}). The real-world demo video can be found on \url{https://youtu.be/gy2bHbBI168}.

\vspace{-0.2cm}

\subsection{Setup and Evaluation} \label{secV.A}

Our simulation is similar to the one released in our previous work SCVP~\cite{pan2022scvp}. The Sensor Perception module uses virtual imaging with a tabletop inserted. The images are $1280 \times 720$ in size and the camera parameters are copied from an Intel Realsense D435 camera. The OctoMap used in the Scene Updating module has the default settings and the dynamic resolution discussed in Sec.~\ref{secIV.B}. We train and evaluate the methods on a PC with an Intel Core i7-12700H CPU, 32 GB RAM, and an Nvidia GeForce RTX3090 GPU.

\begin{figure}[!t]
\centering
\includegraphics[width=0.9\columnwidth]{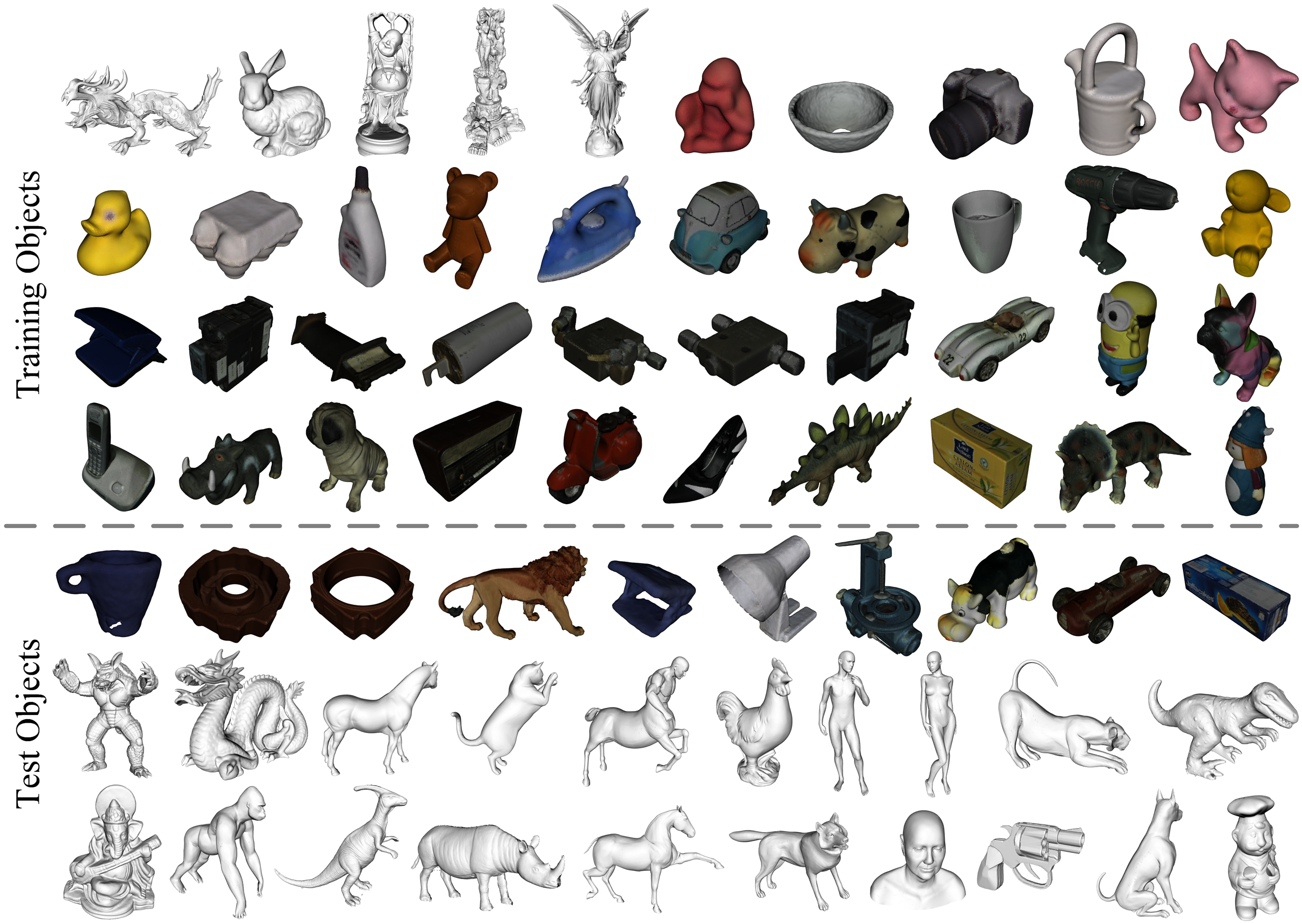}
\caption{Object 3D models: training and test objects.}
\label{fig_object_models}
\vspace{-0.5cm}
\end{figure}

\subsubsection{Object Datasets}

Scanned object 3D mesh models used in the simulation are from the Stanford 3D Scanning Repository~\cite{krishnamurthy1996fitting}, Linemod~\cite{hinterstoisser2012model}, HomebrewedDB~\cite{kaskman2019homebreweddb}, and Clutter~\cite{rodola2013scale}. The range of object sizes is augmented within the interval of [0.05, 0.15]\,m, and the radius of the view-space sphere is thereby set as 0.4\,m. Note that we omit some repetitive models. In total, there are 70 complex models and we split these models into 40 for training objects and 30 for test objects (unknown objects) as shown in Fig.~\ref{fig_object_models}.

% \footnote{\url{https://graphics.stanford.edu/data/3Dscanrep}}
% \footnote{\url{https://campar.in.tum.de/Main/StefanHinterstoisser}}
% \footnote{\url{https://campar.in.tum.de/personal/ilic/homebreweddb}}
% \footnote{\url{https://cvg.cit.tum.de/data/datasets/clutter}}

\subsubsection{Test Cases and Statistical Significance}

The view planning performance can be slightly sensitive to the object rotations and initial views. We randomly select five different object rotations and twenty initial views for each object model. In total, we have $30\times5\times20=3,000$ test cases. Importantly, all methods were evaluated on the same test cases, allowing for two-tailed statistical significance analysis using both the paired t-test and the Wilcoxon signed-rank test (due to some non-normally distributed data), with an alpha value of 0.05.

\subsubsection{Evaluation}

\begin{figure}[!t]
\centering
\includegraphics[width=0.75\columnwidth]{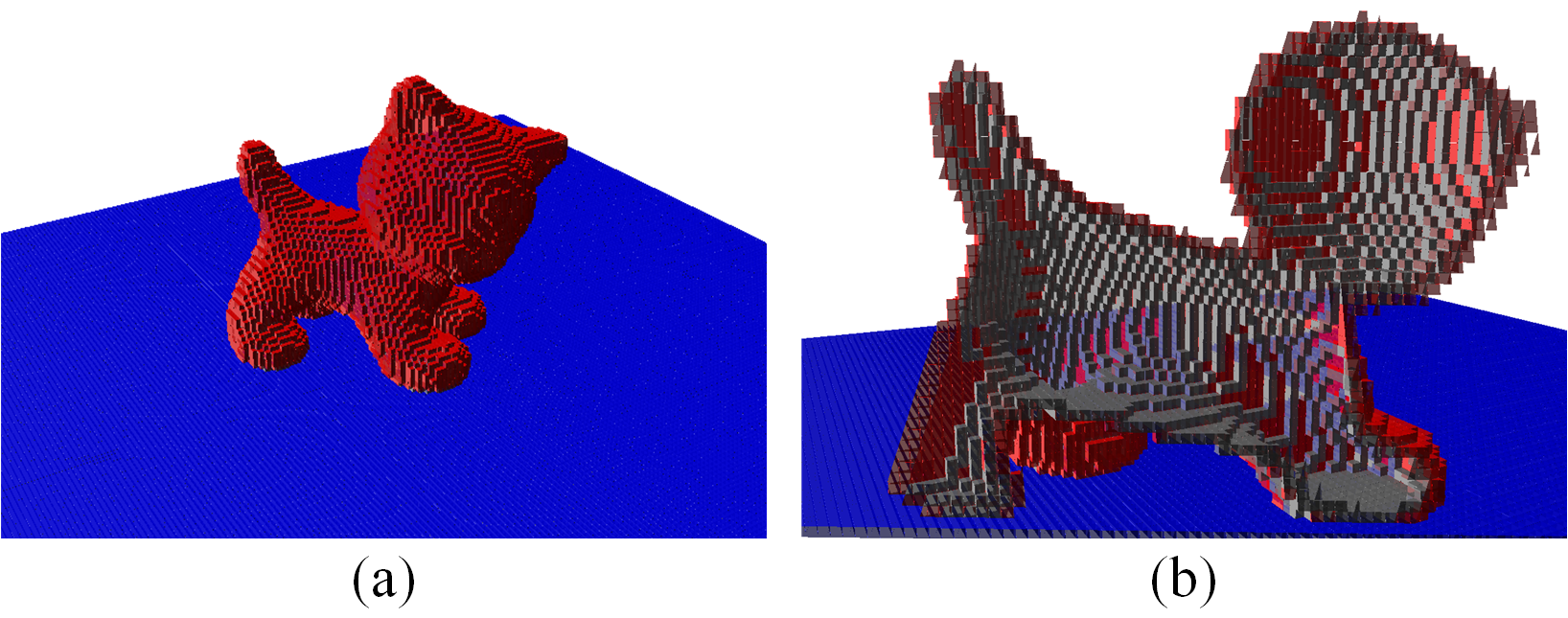}
\caption{Illustration of the invisible surfaces: (a) Outside perspective. (b) Inside perspective: bottom and inner surface voxels (gray).}
\label{fig_bottom_inner}
\vspace{-0.5cm}
\end{figure}

The reconstruction quality is quantified by the visible surface coverage \textbf{\textit{VSC}}:
\begin{equation}
\label{equ8}
\begin{aligned}
VSC = \frac{|U_{cover}|}{|U|}
\end{aligned}
\end{equation}
We use this type of surface coverage to highlight the difference because the bottom and inner surfaces cannot be seen by any of the possible candidate views as shown in Fig.~\ref{fig_bottom_inner}.

Reconstruction efficiency is evaluated by required views, movement cost, and reconstruction time. Since our network has the auto-stop criterion, it is worth evaluating how many views are generated in the whole pipeline. The number of required views \textbf{\textit{RV}} is the total number of views to finish the reconstruction including the initial view:
\begin{equation}
\label{equ9}
\begin{aligned}
RV = |V^\ast|
\end{aligned}
\end{equation}
The movement cost \textbf{\textit{MC}} is the total local path length of the optimal view sequence $V^\ast$:
\begin{equation}
\label{equ10}
\begin{aligned}
MC = \sum_{i=0}^{RV-2}Path(v_i,v_{i+1})
\end{aligned}
\end{equation}
Note that the number of local paths is $RV-1$. 
We assess the reconstruction time \textbf{\textit{RT}} that is the total execution time.

\subsubsection{Comparative Methods}

These \textit{NBV} planning methods are employed as baselines:
\begin{itemize}
\item APORA~\cite{daudelin2017adaptable}, RSE~\cite{delmerico2018comparison}: These voxel-based search methods consider local information gain.
\item MCMF~\cite{pan2022aglobal}, GMC~\cite{pan2023global}: These voxel-based search methods consider both local information gain and global coverage optimization.
\item NBVNet~\cite{mendoza2020supervised}: This voxel-based learning method considers a fixed number 14 of candidate views. We retrained the network under our settings.
\item PCNBV~\cite{zeng2020pc}: This point-cloud-based learning method considers a fixed number 33 of candidate views. We retrained the network under our settings.
\end{itemize}
This \textit{one-shot} planning method is employed as a baseline:
\begin{itemize}
\item SCVP~\cite{pan2022scvp}: Our previous one-shot view planning network considers the singleview input.
\end{itemize}
For our \textit{combined} pipeline, we give some notations for clarity:
\begin{itemize}
\item k-NBV+SCVP: These methods incorporate $k$ NBV before activating the SCVP network~\cite{pan2022scvp}.
\item k-NBV+MA-SCVP: These methods incorporate $k$ NBV before activating the our MA-SCVP network. For example, 1-NBV+MA-SCVP means that we plan a single NBV and then activate the MA-SCVP network.
\end{itemize}

\vspace{-0.3cm}

\subsection{Network Training} \label{secV.B}

As discussed in Algorithm~\ref{alg3}, we can generate view planning datasets for both set covering and NBV planning using 40 training objects. The datasets generated are randomly divided into training and validation sets in an 8:2 ratio.

\subsubsection{View Planning Datasets}

To explore the effectiveness of sampling methods on training, we examine two sampling methods (NBVR sampling and long-tail sampling) along with two levels of the total number of samples (20K and 8K). $n_{single}=32, 8$ is utilized to generate the total number of long-tail samples, resulting in 20,740 and 7,962 samples, denoted as Long-Tail-20K and Long-Tail-8K, respectively. As NBVR samples are dynamically obtained by the number of initial views, we generate a comparable number of samples, totaling 20,952 and 8,281, denoted as NBVR-20K and NBVR-8K, respectively. It takes around 25 hours to generate the Long-Tail-20K dataset. Our datasets can be found on \url{https://www.kaggle.com/datasets/sicongpan/ma-scvp-dataset}. 

\begin{table}[!t]
\centering
\resizebox{\columnwidth}{!}{%
\begin{tabular}{!{\vrule width0.5pt}c!{\vrule width0.5pt}c!{\vrule width0.5pt}c!{\vrule width0.5pt}c!{\vrule width0.5pt}c!{\vrule width0.5pt}} 
\hline
Network & Dataset & Recall\,(\%) & Precision\,(\%) & $F_1$\,(\%)\\
\Xhline{1pt}

\multirow{4}*{\makecell{OctreeNet~\cite{wang2019octreenet} \\ w/o view state}} & NBVR-8K & 75.87 & 69.39 & 72.49 \\
\cline{2-5}
~ & Long-Tail-8K & 68.13 & 72.60 & 70.29 \\
\cline{2-5}
~ & NBVR-20K & 80.50 & 74.54 & 77.41 \\
\cline{2-5}
~ & Long-Tail-20K & 81.48 & 82.45 & 81.96 \\
\Xhline{1pt}

\multirow{4}*{\makecell{OctreeNet~\cite{wang2019octreenet} \\ w/ view state}} & NBVR-8K & 77.10 & 81.00 & 79.00 \\
\cline{2-5}
~ & Long-Tail-8K & 67.84 & 70.65 & 69.22 \\
\cline{2-5}
~ & NBVR-20K & 79.98 & 77.44 & 78.69 \\
\cline{2-5}
~ & Long-Tail-20K & 82.40 & 81.45 & 81.92 \\
\Xhline{1pt}

\multirow{4}*{\makecell{MA-SCVP \\ w/o view state}} & NBVR-8K & 68.27 & 50.47 & 58.04 \\
\cline{2-5}
~ & Long-Tail-8K & 79.60 & 57.69 & 66.90 \\
\cline{2-5}
~ & NBVR-20K & 72.37 & 59.04 & 65.03 \\
\cline{2-5}
~ & Long-Tail-20K & \textbf{86.13} & 76.23 & 80.88 \\
\Xhline{1pt}

\multirow{4}*{\makecell{MA-SCVP \\ w/ view state}} & NBVR-8K & 73.62 & 75.03 & 74.32 \\
\cline{2-5}
~ & Long-Tail-8K & 74.31 & 74.21 & 74.26 \\
\cline{2-5}
~ & NBVR-20K & 80.26 & 84.29 & 82.22 \\
\cline{2-5}
~ & Long-Tail-20K & 83.95 & \textbf{85.76} & \textbf{84.84} \\
\Xhline{1pt}

\end{tabular}
}
\caption{Evaluation of network architecture and dataset on training metrics: Each metric value is computed on the validation set. \textbf{w/} stands for with, while \textbf{w/o} stands for without. As can be seen, our MA-SCVP with view state trained on the Long-Tail-20K dataset achieves the highest $F_1$ score, indicating superior network performance.
}
\label{tab_metrics}
\vspace{-0.5cm}
\end{table}

\subsubsection{Training Details}

To explore the effectiveness of multiview-activated architecture, we examine our MA-SCVP networks with and without the input of the view state. We employ the Adam optimizer~\cite{kingma2014adam} with a base learning rate set at 0.0004 and a mini-batch size of 8. The networks undergo training for 300 epochs. The parameters $\lambda$ and $\gamma$ are examined in Appendix~\ref{Appendix_hyperparameter}.

\subsubsection{Metrics and Results}

We used two metrics to evaluate the network performance:
\begin{equation}
\label{equ11}
Recall\ =\ \frac{|V_{true}^\ast|}{|V_{label}^\ast|}, \ 
Precision\ =\ \frac{|V_{true}^\ast|}{|V_{cover}^\ast|}
\end{equation}
where $V_{true}^\ast=\{v|v\in V_{cover}^\ast \wedge v\in V_{label}^\ast\}$ is the set of predicted views that are also correct in the label. A low-recall network will miss some views that greatly improve coverage while a low-precision network will have redundancy of predicted views. The $F_1$ score is used to balance these two metrics:
\begin{equation}
\label{equ12}
F_1\ =\ \frac{2\times Recall\times Precision}{Recall+Precision}
\end{equation}

In addition to our MA-SCVP networks, we conduct a comparison with OctreeNet~\cite{wang2019octreenet}, a recent architecture designed for voxel grid processing. The results are reported in Tabel~\ref{tab_metrics}. From the results, we have three major findings:
\begin{itemize}
\item The datasets with a larger sample size, 20K, generally outperform the datasets with a smaller number, 8K. This encourages us to maximize $n_{single}$ to 32.
\item The Long-Tail datasets generally outperform the NBVR datasets. This confirms our long-tail sampling method enhances the network training. 
\item The MA-SCVP with view state generally outperforms other architectures. This confirms our architecture enhances the network training. 
\end{itemize}

Based on these findings, we proceed with experiments using 20K datasets and architectures with view state. Therefore, the following MA-SCVP refers to MA-SCVP with view state, the following NBVR dataset refers to NBVR-20K, and the following Long-Tail dataset refers to Long-Tail-20K.

\begin{table}[!t]
\centering
\resizebox{1.0\columnwidth}{!}{%
\begin{tabular}{!{\vrule width0.5pt}c!{\vrule width0.5pt}cc!{\vrule width0.5pt}cc!{\vrule width0.5pt}}
\hline
\multirow{2}*{Iterations} & \multicolumn{2}{c!{\vrule width0.5pt}}{PCNBV~\cite{zeng2020pc}} & \multicolumn{2}{c!{\vrule width0.5pt}}{NBVNet~\cite{mendoza2020supervised}}\\
\cline{2-5}
~ & Long-Tail & NBVR & Long-Tail & NBVR \\
\Xhline{1pt}
% 0 & 38.16 ± 7.55 & 38.16 ± 7.55 & 38.16 ± 7.55 & 38.16 ± 7.55\\
% \cline{1-5}
1 & \textbf{71.90} ± 8.63 & 71.49 ± 8.77 & \textbf{71.01} ± 9.12 & 70.09 ± 9.57\\
\cline{1-5}
2 & \textbf{85.68} ± 5.98 & 85.32 ± 6.14 & \textbf{84.06} ± 7.29 & 82.83 ± 7.92\\
\cline{1-5}
3 & \textbf{91.40} ± 4.93 & 91.14 ± 4.87 & \textbf{89.47} ± 6.52 & 88.74 ± 6.99\\
\cline{1-5}
4 & \textbf{93.65} ± 4.03 & 93.83 ± 3.74 & \textbf{91.90} ± 5.84 & 91.44 ± 6.21\\
\cline{1-5}
5 & 94.74 ± 3.46 & \textbf{95.34} ± 3.04 & \textbf{93.41} ± 5.23 & 93.15 ± 5.54\\
\cline{1-5}
9 & 96.44 ± 2.51 & \textbf{97.20} ± 2.28 & 95.78 ± 3.97 & \textbf{96.01} ± 4.42\\
\cline{1-5}
14 & 97.20 ± 2.15 & \textbf{97.65} ± 2.05 & 96.44 ± 3.81 & \textbf{97.05} ± 4.00\\
\cline{1-5}
19 & 97.54 ± 1.98 & \textbf{97.85} ± 1.96 & 96.60 ± 3.78 & \textbf{97.29} ± 3.97\\
\Xhline{1pt}
\end{tabular}
}
\caption{Evaluation of NBV networks trained on datasets using different sampling methods: Each value is the mean VSC value with its standard deviation, computed over 3,000 tests. VSC stands for visible surface coverage. The initial VSC (iteration 0) is the same for all methods, with a value of 38.16 ± 7.55. As can be seen, NBV networks trained on the Long-Tail dataset achieve higher coverage in early iterations.
}
\vspace{-0.2cm}
\label{tab_iterative_network}
\end{table}

\begin{table}[!t]
\centering
\resizebox{0.8\columnwidth}{!}{%
\begin{tabular}{!{\vrule width0.5pt}c!{\vrule width0.5pt}ccc!{\vrule width0.5pt}} 
\hline
Dataset & RV & VSC\,(\%) & MC\,(m)\\
\Xhline{1pt}
NBVR & 14.22 ± 2.13 & 98.06 ± 4.50 & 2.83 ± 0.33\\
\cline{1-4}
Long-Tail & \textbf{13.13} ± 1.43 & \textbf{98.10} ± 3.73 & \textbf{2.59} ± 0.23\\
\Xhline{1pt}
\end{tabular}
}
\caption{Evaluation of our MA-SCVP networks trained on datasets using different sampling methods: Each value is the mean value and its standard deviation, computed over 3,000 tests. \textbf{RV}, \textbf{VSC}, and \textbf{MC} stand for the number of required views, visible surface coverage, and movement cost, respectively. As can be seen, MA-SCVP trained on the Long-Tail dataset achieves better VSC, while notably reducing RV and MC, especially smaller standard deviation.
}
\label{tab_final_network}
\vspace{-0.5cm}
\end{table}

\vspace{-0.2cm}

\subsection{Reconstruction Study on Sampling Method} \label{secV.C}

To further study the effectiveness of sampling methods, we conduct reconstruction tests on two NBV networks (PCNBV and NBVNet) and our MA-SCVP network. The NBV networks are trained following the parameters in their paper.

\subsubsection{Performance of NBV Networks}

As shown in Fig.~\ref{fig_multiview_features}, all objects received full coverage below 20 ground-truth NBVs. To test NBV networks thoroughly, Table~\ref{tab_iterative_network} reports the results of NBV networks for the major iterations where $RV=20$. From the results, we find that, in early iterations (1-4 for PCNBV and 1-5 for NBVNet), networks trained on the Long-Tail dataset exhibit higher surface coverage than those trained on the NBVR dataset. In subsequent iterations, the opposite trend is observed.

Based on this finding, we conclude that the long-tail sampling method is well-suited for NBV networks in our system, particularly since we utilize NBVs in early iterations. Therefore, the NBV networks in the following study on combined pipelines are trained on the Long-Tail dataset.

\subsubsection{Performance of MA-SCVP Network}

Since our MA-SCVP network has the auto-stop criterion, the results of the final reconstruction are reported in Table~\ref{tab_final_network}. From the results, we find that MA-SCVP trained on the Long-Tail dataset achieves higher quality and efficiency of reconstruction than that trained on the NBVR dataset.

Based on this finding, we conclude that the long-tail sampling method improves the reconstruction performance of our MA-SCVP network, enhancing our system performance.

\vspace{-0.2cm}

\subsection{Reconstruction Study on Combined Pipeline} \label{secV.D}

To establish our combined pipeline, three components require examination: (1) the selected NBV method for our NBV submodule, (2) the selected network for our one-shot submodule, and (3) the number $k$ of NBVs combined.

\begin{table*}[!t]
\centering
\resizebox{0.75\textwidth}{!}{%
\begin{tabular}{!{\vrule width0.5pt}c!{\vrule width0.5pt}c!{\vrule width0.5pt}c!{\vrule width0.5pt}c!{\vrule width0.5pt}c!{\vrule width0.5pt}c!{\vrule width0.5pt}c!{\vrule width0.5pt}} 
\hline
Iterations & PCNBV~\cite{zeng2020pc} & NBVNet~\cite{mendoza2020supervised} & RSE~\cite{delmerico2018comparison} & APORA~\cite{daudelin2017adaptable} & MCMF~\cite{pan2022aglobal} & GMC~\cite{pan2023global} \\
\Xhline{1pt}
% 0 & 38.16 ± 7.55 & 38.16 ± 7.55 & 38.16 ± 7.55 & 38.16 ± 7.55 & 38.16 ± 7.55 & 38.16 ± 7.55\\
% \cline{1-7}
1 & \textbf{71.90} ± 8.63 & 71.01 ± 9.12 & 68.29 ± 10.37 & 64.67 ± 9.34 & 64.12 ± 10.09 & 64.19 ± 10.07\\
\cline{1-7}
2 & \textbf{85.68} ± 5.98 & 84.06 ± 7.29 & 77.09 ± 9.32 & 76.02 ± 8.00 & 80.86 ± 8.70 & 80.90 ± 8.63\\
\cline{1-7}
3 & \textbf{91.40} ± 4.93 & 89.47 ± 6.52 & 84.09 ± 7.93 & 83.90 ± 6.67 & 88.23 ± 6.53 & 88.43 ± 6.20\\
\cline{1-7}
4 & \textbf{93.65} ± 4.03 & 91.90 ± 5.84 & 88.32 ± 6.68 & 88.98 ± 5.67 & 92.18 ± 4.97 & 92.41 ± 4.65\\
\cline{1-7}
5 & \textbf{94.74} ± 3.46 & 93.41 ± 5.23 & 91.52 ± 5.86 & 91.86 ± 4.70 & 94.42 ± 3.83 & 94.65 ± 3.50 \\
\cline{1-7}
6 & 95.39 ± 3.08 & 94.33 ± 4.80 & 93.68 ± 5.27 & 93.71 ± 4.11 & 95.85 ± 2.86 & \textbf{96.03} ± 2.61 \\
\cline{1-7}
9 & 96.44 ± 2.51 & 95.78 ± 3.97 & 96.82 ± 3.21 & 96.74 ± 2.83 & 97.93 ± 1.68 & \textbf{97.95} ± 1.58 \\
\cline{1-7}
14 & 97.20 ± 2.15 & 96.44 ± 3.81 & 98.26 ± 1.64 & 98.59 ± 1.36 & \textbf{99.02} ± 1.07 & 98.93 ± 1.09 \\
\cline{1-7}
19 & 97.54 ± 1.98 & 96.60 ± 3.78 & 98.95 ± 0.98 & 99.24 ± 0.84 & \textbf{99.48} ± 0.75 & 99.33 ± 0.81\\
\Xhline{1pt}
\end{tabular}
}
\caption{Evaluation of different NBV methods to identify the most effective approach in early iterations: Each value is the mean VSC value with its standard deviation, computed over 3,000 tests. VSC stands for visible surface coverage. The initial VSC (iteration 0) is the same for all methods, with a value of 38.16 ± 7.55. As can be seen, PCNBV exhibits superior performance in the early iterations (1-5), making it our final choice.}
\label{tab_iterative_NBV}
\vspace{-0.2cm}
\end{table*}

\begin{table*}[!t]
\centering
\resizebox{0.70\textwidth}{!}{%
\begin{tabular}{!{\vrule width0.5pt}c!{\vrule width0.5pt}ccc!{\vrule width0.5pt}ccc!{\vrule width0.5pt}} 
\hline
\multirow{2}*{\makecell{$k$ NBV\\Combined}} & \multicolumn{3}{c!{\vrule width0.5pt}}{SCVP  (singleview dataset)~\cite{pan2022scvp}} & \multicolumn{3}{c!{\vrule width0.5pt}}{MA-SCVP (long-tail multiview dataset)} \\
\cline{2-7}
~ & RV & VSC (\%) & MC (m) & RV & VSC (\%) & MC (m) \\
\Xhline{1pt}
0 & \textbf{9.61} ± 0.96 & 96.12 ± 6.29 & \textbf{2.28} ± 0.15 & 13.13 ± 1.43 & 98.10 ± 3.73 & \textbf{2.59} ± 0.23 \\
\cline{1-7}
1 & 10.24 ± 1.03 & 96.99 ± 4.66 & 2.83 ± 0.13 & 13.01 ± 1.46 & 98.41 ± 2.67 & 3.07 ± 0.24 \\
\cline{1-7}
2 & 10.74 ± 1.01 & 97.28 ± 4.42 & 3.45 ± 0.15 & 12.93 ± 1.43 & 98.40 ± 2.79 & 3.67 ± 0.23 \\
\cline{1-7}
3 & 11.26 ± 0.96 & 97.52 ± 4.28 & 4.10 ± 0.19 & \textbf{12.90} ± 1.44 & 98.40 ± 2.79 & 4.29 ± 0.25\\
\cline{1-7}
4 & 11.91 ± 0.92 & 97.96 ± 3.53 & 4.72 ± 0.25 & 13.06 ± 1.51 & 98.56 ± 2.35 & 4.86 ± 0.31 \\
\cline{1-7}
5 & 12.65 ± 0.90 & \textbf{98.27} ± 2.91 & 5.21 ± 0.30 & 13.29 ± 1.56 & \textbf{98.60} ± 2.25 & 5.30 ± 0.35\\
\Xhline{1pt}
\end{tabular}
}
\caption{Evaluation of combined pipelines on SCVP trained on the singleview dataset~\cite{pan2022scvp} and MA-SCVP trained on the long-tail multiview dataset: Each value is the mean value and its standard deviation, computed over 3,000 tests. \textbf{RV}, \textbf{VSC}, and \textbf{MC} stand for the number of required views, visible surface coverage, and movement cost, respectively. Note that $k=0$ means the input of only the initial view (singleview-activated) while $k\geq1$ means the input of the initial view and the additional $k$ NBVs from \textbf{PCNBV} (multiview-activated). As can be seen, MA-SCVP achieves a sufficient VSC with only a single NBV ($k=1$), exhibiting a plateau across $k=1,2,3$ NBVs combined as confirmed in Table~\ref{tab_statistical_test}. Furthermore, although adding more NBVs ($k=4,5$) improves VSC, it also substantially increases the MC. To achieve a better trade-off, we selected 1-NBV+MA-SCVP as our final choice.
}
\label{tab_pipeline}
\vspace{-0.5cm}
\end{table*}

\subsubsection{Determination of NBV Submodule}

Table~\ref{tab_iterative_NBV} reports the iterative reconstruction results of two NBV networks and four search methods in terms of VSC. From the results, we have two major findings:
\begin{itemize}
\item PCNBV outperforms the other methods achieving higher coverage in the early iterations (1-5).
\item Learning-based methods exhibit inferior VSC compared to search-based methods in later iterations, highlighting challenges in high-quality \textit{unknown} object reconstruction.
\end{itemize}

Based on these findings, we finally select the best-performing PCNBV as our NBV submodule because we only employ NBVs in the early iterations.

\subsubsection{Determination of One-Shot Submodule and Number k}

Table~\ref{tab_pipeline} reports the final reconstruction results of our combined pipelines with $k=0-5$ NBVs from PCNBV on both the SCVP network trained on the singleview dataset~\cite{pan2022scvp} and our MA-SCVP network trained on the long-tail multiview dataset. From the results, we have four major findings:
\begin{itemize}
\item Increasing the number $k$ of NBVs incorporated into SCVP enhances VSC but simultaneously leads to an increase in RV and MC. This suggests that incorporating \textit{online} surface information from NBV perception data enhances the performance of the one-shot network. However, training it solely on the singleview dataset renders it unsuitable for our pipeline as reflected in the increased RV and MC.
\item Without incorporating NBVs ($k=0$), MA-SCVP trained on the long-tail multiview dataset exhibits a significantly higher VSC compared to SCVP trained on the singleview dataset. This suggests that our long-tail multiview sampling and multiview-activated architecture \textit{offline} provide more geometric knowledge to the network.
\item When incorporating NBVs into MA-SCVP, a single NBV ($k=1$) proves adequate to achieve a high VSC. Adding a few more NBVs ($k=2,3$) yields no significant benefit for VSC, as a stable plateau is observed. This trend is confirmed by significance tests presented in Table~\ref{tab_statistical_test}.
\item While incorporating a larger number of NBVs ($k=4,5$) into MA-SCVP leads to a significant increase in VSC, it also results in a substantial increase in MC (58\% and 73\%). This is logical, as adding more NBVs introduces additional local paths and more online information (pure NBVs can achieve a very high VSC as in Table~\ref{tab_iterative_NBV}).
\end{itemize}

\begin{table}[!t]
\centering
\resizebox{1.0\columnwidth}{!}{%
\begin{tabular}{!{\vrule width0.5pt}c!{\vrule width0.5pt}c!{\vrule width0.5pt}c!{\vrule width0.5pt}c!{\vrule width0.5pt}c!{\vrule width0.5pt}} 
\hline
\multirow{2}*{\makecell{Method}} & \multirow{2}*{\makecell{Reference Method}} & \multicolumn{2}{c!{\vrule width0.5pt}}{Significance of VSC} & \multirow{2}*{\makecell{Growth Rate \\ of MC}}\\
\cline{3-4}
~ & ~ & t-test & Wilcoxon & ~ \\
\Xhline{1pt}
0-NBV+MA-SCVP & 0-NBV+SCVP & $\mathbb{\star\star\star}$ & $\mathbb{\star\star\star}$ & \textbf{14\%} \\
\cline{1-5}
1-NBV+MA-SCVP & 0-NBV+MA-SCVP & $\mathbb{\star\star\star}$ & $\mathbb{\star}$ & 19\% \\
\cline{1-5}
2-NBV+MA-SCVP & 1-NBV+MA-SCVP & n.s. & n.s. & 20\% \\
\cline{1-5}
3-NBV+MA-SCVP & 1-NBV+MA-SCVP & n.s. & n.s. & 40\% \\
% \cline{1-5}
% 4-NBV+MA-SCVP & 1-NBV+MA-SCVP & $\star\star\star$ & $\star\star\star$ & 58\% \\
% \cline{1-5}
% 5-NBV+MA-SCVP & 1-NBV+MA-SCVP & $\star\star\star$ & $\star\star\star$ & 73\% \\
\Xhline{1pt}
\end{tabular}
}
\caption{Statistical significance analysis using both paired t-test and Wilcoxon signed-rank test on different combined pipelines in Table~\ref{tab_pipeline}. $\mathbb{\star\star\star}$ denotes statistical significance at $p<0.001$, $\star$ denotes significance at $p<0.05$, and n.s. indicates non-significance. \textbf{VSC} and \textbf{MC} stand for visible surface coverage and movement cost. As can be seen, (1) 0-NBV+MA-SCVP exhibits a significant VSC increase compared to 0-NBV+SCVP, indicating the effectiveness of our long-tail multiview sampling; (2) 1-NBV+MA-SCVP exhibits a significant VSC increase compared to 0-NBV+MA-SCVP, indicating the effectiveness of our combined pipeline with a single NBV; (3) the rest $k$-NBV+MA-SCVP ($k=2,3$) do not exhibit a significant VSC increase compared to 1-NBV+MA-SCVP but they require considerably more MC.
}
\label{tab_statistical_test}
\vspace{-0.5cm}
\end{table}

\begin{table*}[!t]
\centering
\resizebox{0.70\textwidth}{!}{%
\begin{tabular}{!{\vrule width0.5pt}c!{\vrule width0.5pt}c!{\vrule width0.5pt}c!{\vrule width0.5pt}ccc!{\vrule width0.5pt}} 
\hline
System Type             & Stop Criterion               & Method              & RV & VSC\,(\%) & MC\,(m) \\ 
\Xhline{1pt}
\multirow{20}{*}{Iterative NBV} & \multirow{10}{*}{\makecell{$m$-based\\($<0.5\%$)}}  & PCNBV~\cite{zeng2020pc} & 6.96 ± 1.64  & 95.86 ± 2.90 & 3.74 ± 0.88 \\ 
\cline{3-6} 
                          &                             & NBVNet~\cite{mendoza2020supervised}              & \textbf{6.94} ± 1.90    & 93.61 ± 5.71          & 3.70 ± 1.03       \\ 
\cline{3-6} 
                          &                             & RSE~\cite{delmerico2018comparison}                 & 7.33 ± 2.33    & 93.39 ± 6.06          & 3.22 ± 1.10       \\ 
\cline{3-6} 
                          &                             & APORA~\cite{daudelin2017adaptable}               & 7.13 ± 2.09    & 92.52 ± 6.72          & 3.34 ± 1.27       \\ 
\cline{3-6} 
                          &                             & MCMF~\cite{pan2022aglobal}                & 7.06 ± 1.88    & 95.34 ± 4.33          & 3.54 ± 1.07       \\ 
\cline{3-6} 
                          &                             & GMC~\cite{pan2023global}                 & 7.06 ± 1.84    & 95.59 ± 3.83          & 3.54 ± 1.04       \\ 
\cline{3-6} 
                          &                             & RSE+Mov~\cite{delmerico2018comparison}             & 7.13 ± 2.38    & 90.66 ± 7.50          & 2.70 ± 1.11       \\ 
\cline{3-6} 
                          &                             & APORA+Mov~\cite{daudelin2017adaptable}           & 7.13 ± 2.06    & 92.49 ± 6.71          & 3.30 ± 1.23       \\ 
\cline{3-6} 
                          &                             & MCMF+Mov~\cite{pan2022aglobal}             & 8.02 ± 2.33    & 92.60 ± 10.12         & 2.40 ± 0.91       \\ 
\cline{3-6} 
                          &                             & GMC+Mov~\cite{pan2023global}             & 6.89 ± 1.86    & 93.73 ± 4.98          & 2.80 ± 0.94       \\ 
\Xcline{2-6}{1pt}
                                & \multirow{10}{*}{\makecell{$m$-based\\($<0.1\%$)}} & PCNBV~\cite{zeng2020pc} & 10.40 ± 3.30 & 97.07 ± 2.33 & 5.21 ± 1.36 \\ 
\cline{3-6} 
                          &                             & NBVNet~\cite{mendoza2020supervised}              & 10.37 ± 3.55   & 95.62 ± 4.67          & 5.35 ± 1.64       \\ 
\cline{3-6} 
                          &                             & RSE~\cite{delmerico2018comparison}                 & 11.74 ± 3.99   & 97.32 ± 2.87          & 4.62 ± 1.46       \\ 
\cline{3-6} 
                          &                             & APORA~\cite{daudelin2017adaptable}               & 11.76 ± 3.92   & 96.44 ± 4.43          & 5.52 ± 1.96       \\ 
\cline{3-6} 
                          &                             & MCMF~\cite{pan2022aglobal}                 & 11.39 ± 3.83   & 97.85 ± 2.40          & 5.73 ± 1.94       \\ 
\cline{3-6} 
                          &                             & GMC~\cite{pan2023global}                 & 11.27 ± 3.82   & 97.86 ± 2.29          & 5.57 ± 1.80       \\ 
\cline{3-6} 
                          &                             & RSE+Mov~\cite{delmerico2018comparison}             & 11.70 ± 4.22   & 95.50 ± 4.97          & 3.91 ± 1.50       \\ 
\cline{3-6} 
                          &                             & APORA+Mov~\cite{daudelin2017adaptable}           & 11.74 ± 3.91   & 96.41 ± 4.42          & 5.37 ± 1.89       \\ 
\cline{3-6} 
                          &                             & MCMF+Mov~\cite{pan2022aglobal}             & 12.47 ± 3.79   & 97.46 ± 4.65          & 4.15 ± 1.46       \\ 
\cline{3-6} 
                          &                             & GMC+Mov~\cite{pan2023global}             & 11.46 ± 4.18   & 96.80 ± 3.60          & 4.28 ± 1.62       \\ 
\Xhline{1pt}
One-Shot                          & Auto-Stop                   & SCVP~\cite{pan2022scvp}                & 9.61 ± 0.96    & 96.12 ± 6.29          & \textbf{2.28} ± 0.15       \\
\Xhline{1pt}
Combined\,(Ours)                  & Auto-Stop                   & 1-NBV+MASCVP & 13.01 ± 1.46   & $\mathbb{^{\star\star\star}}$\,\textbf{98.41} ± 2.67 & 3.07 ± 0.24       \\ 
\Xhline{1pt}
\end{tabular}
}
\caption{Evaluation of three types of view planning systems: iterative NBV, one-shot, and combined. Each value is the mean value and its standard deviation, computed over 3,000 tests. $m$-based stop criterion~\cite{yervilla2019optimal,yervilla2022bayesian} finishes the reconstruction when the variation in the number of frontier voxels becomes smaller than a threshold for consecutive $m=3$ sensing operations. The threshold is set to 0.1\% (strict) or 0.5\% (lenient) of the total voxels. \textbf{+Mov} stands for the movement-cost weighted NBV evaluation. \textbf{RV}, \textbf{VSC}, and \textbf{MC} stand for the number of required views, visible surface coverage, and movement cost, respectively. $\mathbb{^{\star\star\star}}$ indicates a consistent significance at $p<0.001$ between our method and all baselines (\ie, several paired tests), as determined by both paired t-test and the Wilcoxon signed-rank test. As can be seen, our 1-NBV+MASCVP achieves a significantly higher VSC than all baselines, while reducing MC by 45\% compared to the second-highest VSC method (GMC with $<0.1\%$).
}
\label{tab_system_compare}
\vspace{-0.5cm}
\end{table*}

Based on these findings, we consider the trade-off between achieving a high VSC and using fewer NBVs and less MC, opting for MA-SCVP with $k=1$ as the number of combined NBVs for the one-shot submodule. It is worth mentioning that 1-NBV+MA-SCVP demonstrates an even higher VSC (98.41) than 5-NBV+SCVP (98.27), accompanied by a substantial MC reduction of 41\%. This supports our claim that our long-tail multiview sampling effectively reduces the required NBVs and MC. The following 1-NBV+MA-SCVP stands for this determined pipeline. 

To offer a more thorough comprehension of our pipeline, we introduce an analysis of how different NBV submodules influence MA-SCVP performance in Appendix~\ref{Appendix_NBVsubmodule}.

\vspace{-0.2cm}

\subsection{Reconstruction Study on System Performance} \label{secV.E}

A typical view planning system needs two major setups: (1)~the view planning method, and (2)~the stop criterion.

\subsubsection{System Setup} Our \textit{combined} system utilizes the 1-NBV+MA-SCVP method with an auto-stop criterion. 

For the \textit{one-shot} system, the SCVP method~\cite{pan2022scvp} also contains an auto-stop criterion.

For \textit{iterative} NBV systems, most view planning methods are designed to maximize surface coverage, \ie, enhancing reconstruction quality. Delmerico \etal~\cite{delmerico2018comparison} propose a simple way to consider reconstruction efficiency simultaneously for search-based methods, using the movement-cost weighted NBV evaluation:
\begin{equation}
\mathit{utility(v)} = \frac{gain(v)}{\sum_V gain} - \frac{cost(v)}{\sum_V cost}
\end{equation}
where $gain(v)$ is the information gain defined for the search and $cost(v)$ is the movement cost of a candidate view. This cost is defined in such a way that the farther away a candidate view is from the view position in the current iteration, the higher the cost value, resulting in a lower $\mathit{utility(v)}$ for that candidate view. We incorporate this weighting approach for search-based methods, with the exception of GMC~\cite{pan2023global}, which has its own movement-cost weight. To distinguish them, we append the tag \textbf{+Mov} after the names of the search-based methods.

\begin{table}[!t]
\centering
\resizebox{0.85\columnwidth}{!}{%
\begin{tabular}{!{\vrule width0.5pt}c!{\vrule width0.5pt}c!{\vrule width0.5pt}c!{\vrule width0.5pt}} 
\hline
Method & Inference Time\,(s) & Inference Number \\
\cline{1-3} 
RSE~\cite{delmerico2018comparison} & 1.134 ± 0.178          & \multirow{6}{*}{\makecell{About 7\,(lenient)\\or 11\,(strict)}} \\
\cline{1-2} 
APORA~\cite{daudelin2017adaptable} & 1.141 ± 0.175          &  \\
\cline{1-2} 
MCMF~\cite{pan2022aglobal} & 1.274 ± 0.312 &  \\
\cline{1-2} 
GMC~\cite{pan2023global} & 1.518 ± 1.155 &  \\
\cline{1-2}
PCNBV~\cite{zeng2020pc} & 0.099 ± 0.019         &  \\
\cline{1-2} 
NBVNet~\cite{mendoza2020supervised} & \textbf{0.006} ± 0.001          & \\
\cline{1-3} 
SCVP~\cite{pan2022scvp} & 0.112 ± 0.121          & \textbf{1} \\
\cline{1-3} 
1-NBV+MASCVP\,(Ours) & 0.119 ± 0.038          & 2 \\
\Xhline{1pt}
\end{tabular}
}
\caption{Evaluation of inference time of different methods: The mean time for inferring view planning module once and its standard deviation are reported, computed over 3,000 tests. The number of inferences for NBV methods depends on the stop criterion, which is approximately 7 or 11. As can be seen, learning-based methods are much faster than search-based methods.
}
\label{tab_infer_time}
\vspace{-0.5cm}
\end{table}

It is important to note that our auto-stop criterion is only achievable by running our network, which is not feasible for NBV methods. As demonstrated in the literature for NBV stopping tests~\cite{yervilla2019optimal,yervilla2022bayesian}, the $m$-based stop criterion exhibits the highest reconstruction quality among different criteria. For unknown scenarios, this criterion finishes the reconstruction when the variation in the number of frontier voxels becomes smaller than a threshold for consecutive $m$ sensing operations. A frontier voxel is an unknown voxel and has at least one free and one occupied neighbor. $m$ is set to 3. We explore both a strict and lenient threshold, computed based on the number of voxels representing 0.1\% and 0.5\% of the total within the $32\times32\times32$ object bounding box, respectively.

\subsubsection{Comparison of System Performance}

Table~\ref{tab_system_compare} reports the comparison of different types of view planning systems. Note that the results of PCNBV and NBVNet trained on the NBVR dataset are reported here, as they demonstrate better performance in the later iterations. From the results, we have four major findings:
\begin{itemize}
\item With a lenient stop criterion, iterative NBV systems achieve a comparable MC against our combined system but exhibit much lower VSC.
\item With a strict stop criterion, iterative NBV systems achieve a comparable VSC against our combined system but exhibit much higher MC.
\item Our combined system achieves a significant improvement in VSC compared to all iterative NBV systems and the one-shot system.
\item Our combined system reduces MC by 45\% compared to the second-highest VSC method (GMC with $<0.1\%$).
\end{itemize}

Besides, the results also validate that our system provides an effective stop criterion. We further analyze it in Appendix~\ref{Appendix_stopcriterion}.

\subsubsection{Inference and Update Time} To have a more comprehensive analysis, we break down the reconstruction time into inference time and OctoMap update time. Table~\ref{tab_infer_time} reports the time required for inferring view planning once. As a learning-based method, our system achieves a short inference time. The OctoMap update time is measured at 0.163 ± 0.111 seconds for a single update. OctoMap-based NBV methods need to be updated at each iteration (about 7 or 11 in total), while our system requires only two updates (one for the initial view and one for NBV). This also indicates that the remaining number of required views does not reduce the efficiency of our system. Overall, our system exhibits high computational efficiency, totaling around 0.564 seconds.

Based on all these comparisons, we now conclude that our system achieves both high-quality and high-efficiency reconstruction of unknown objects. 

\vspace{-0.2cm}

\subsection{Real-World Experiments} \label{secV.F}

The hardware of our real-world system is a 6-DOF UR5 arm with an Intel Realsense D435 camera mounted on its end-effector. Experiments are performed on the ROS platform~\cite{quigley2009ros}. The physical constraints of the joints in our robot arm limit the workspace and movement. Some candidate view positions may be unreachable due to these constraints. To address this, we adjust the candidate view positions by moving them up by 0.008\,m to ensure reachability. Additionally, the computed cartesian trajectories occasionally fail to follow our planned paths. To mitigate this, we lower the object center to the table height, allowing for less constrained view poses. This adjustment has minimal impact on the reconstruction since the object remains within the camera field of view. If these adjustments still result in path failures, we directly solve the joint inverse solution for the target view. Importantly, these adjustments are consistent across all methods, ensuring a valid and fair comparison.

\subsubsection{Ground Truth Model and Noise}

\begin{figure}[!t]
\centering
\includegraphics[width=0.80\columnwidth]{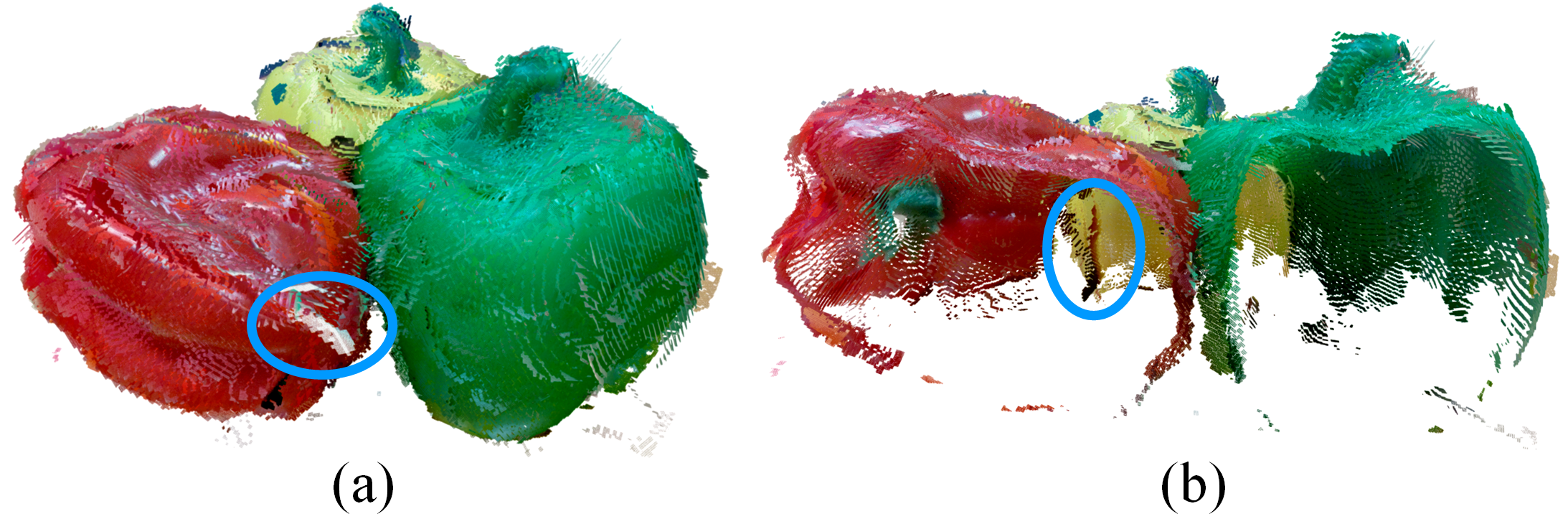}
\caption{
An example illustrating real-world noise in the ground truth point cloud: (a) Outside perspective. (b) Inside perspective (cross-section). The areas marked in blue highlight some of the noise spots on the surface and noise inside the object.
} 
\label{fig_noise}
\vspace{-0.2cm}
\end{figure}

To calculate the surface coverage, having a ground truth model of the object is essential. This is achieved by collecting point clouds from the entire view space and voxelizing them to 0.002\,m, serving as the ground truth model. It is worth mentioning that slight calibration errors may result in a small offset in the point cloud for each view. To address this issue, for a voxel on the ground truth model, we search for its corresponding neighbors. If any neighboring voxels are covered, it is included in the coverage count. However, achieving 100\% surface coverage in a real reconstruction trail is challenging due to sensor noise in the real world. Factors such as lighting and environmental conditions can affect the accuracy of depth camera perception, introducing some noise points. Fig.~\ref{fig_noise} illustrates real-world noise. While noise points are inevitable (less than 1\%), the comparison remains valid and fair as all methods utilize the same camera configuration.

\subsubsection{Dynamic Object Bounding Box and View Space}

\begin{figure}[!t]
\centering
\includegraphics[width=1.0\columnwidth]{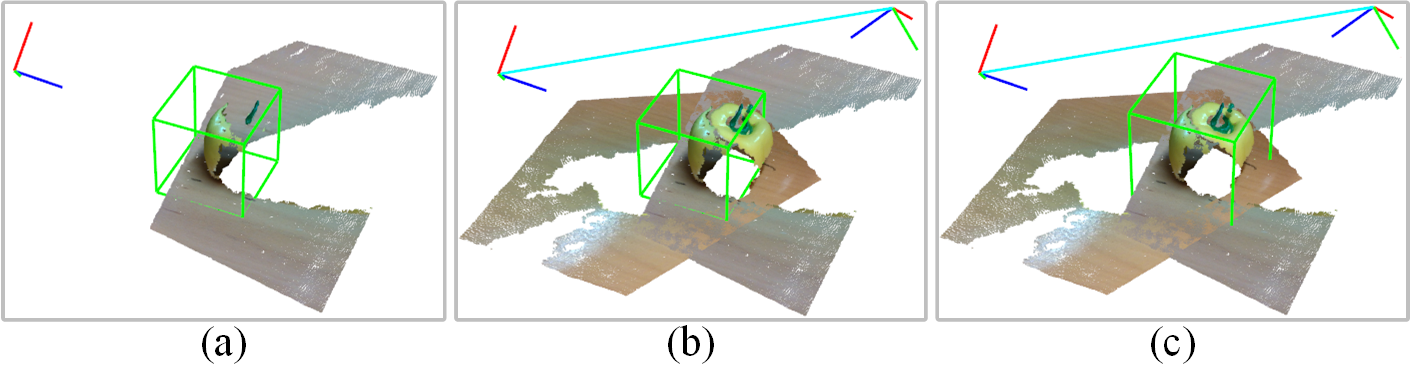}
\caption{
Dynamic computation of the object center and size: partially reconstructed scenes, paths (cyan), planned views (red-green-blue), and the bounding box (green cube). (a) The bounding box is computed from the initial view. (b) Upon the arrival of the point cloud from the second view, it becomes evident that the existing bounding box is insufficient to encompass the entire object. (c) Subsequently, the bounding box is updated based on the new point cloud.
} 
\label{fig_dynamic_bbx}
\vspace{-0.2cm}
\end{figure}

The object center and size are known in advance in simulation, \ie, the object bounding box with an edge length equal to two times the size is determined. However, in the real world, this information is unknown. Therefore, we employ a dynamic strategy to obtain the bounding box of an object. As shown in Fig.~\ref{fig_dynamic_bbx}, the bounding box is dynamically updated based on the partially reconstructed point clouds of the object. Therefore, the center of the view space undergoes a corresponding change along with the object center as discussed in Sec.~\ref{secIII.C}. The sphere radius of the view space is set to the computed $o_{size}$ plus 0.15\,m because the camera needs a certain distance to obtain correct depth images. 

\begin{table}[!t]
\centering
\resizebox{1.0\columnwidth}{!}{%
\begin{tabular}{!{\vrule width0.5pt}c!{\vrule width0.5pt}cccc!{\vrule width0.5pt}} 
\hline
Method & RV & VSC\,(\%) & MC\,(m) & RT\,(s)\\
\Xhline{1pt}
Ours with GT & \textbf{12.78} ± 1.09 & \textbf{99.36} ± 0.32 & \textbf{1.71} ± 0.19 & \textbf{102.5} ± 14.8 \\
\cline{1-5}
Ours Dynamic & 12.93 ± 1.27 & 99.35 ± 0.29 & 1.73 ± 0.22 & 105.3 ± 12.6 \\
\Xhline{1pt}
\end{tabular}
}
\caption{Evaluation of real-world generalization performance for unknown object center and size: \textbf{Ours with GT} denotes reconstruction based on the ground truth object center and size, whereas \textbf{Ours Dynamic} denotes reconstruction with dynamically updated object center and size. Each value is the mean value and its standard deviation, computed over 60 tests. \textbf{RV}, \textbf{VSC}, \textbf{MC}, and \textbf{RT} stand for the number of required views, visible surface coverage, movement cost, and reconstruction time, respectively. As can be seen, our system is not sensitive to deviations in the object center and size.
}
\label{tab_generalization_study}
\vspace{-0.5cm}
\end{table}

\subsubsection{Object Test Cases}

We test three objects under different sizes: a single crop measuring 0.056\,m (Fig.~\ref{fig_dynamic_bbx}), cluttered crops measuring 0.093\,m (Fig.~\ref{fig_noise}), and a complex cup measuring 0.060\,m (Fig.~\ref{fig_complex_cup}). We randomly assign two different object rotations, five initial views, and two repetitions for each object. In total, we conduct $3\times2\times5\times2=60$ test cases to mitigate the influence of potential noise.

\subsubsection{Generalization Study on Our System}

Since we need to employ the dynamic strategy in real-world scenarios, we investigate the sensitivity of our system to deviations in the object center and size. Table~\ref{tab_generalization_study} reports the results with and without ground truth. From the results, we confirm that (1) our system exhibits robust generalization capability to unknown object centers and sizes, and (2) our networks, trained solely in simulation, demonstrate satisfactory performance when directly deployed in a real-world environment.

\subsubsection{Comparison of System Performance} 

Based on the results presented in Table~\ref{tab_system_compare}, we select three baselines for comparison: the one-shot system SCVP and two iterative NBV systems of the best-performing GMC among NBV methods without movement weights and the best-performing MCMF+Mov among NBV methods with movement weights. For NBV systems, the stop criterion utilized is the $m$-based criterion ($m=3$) with a threshold of 0.4\% instead of 0.1\% to achieve a similar performance as in the simulation. Note that all these systems are employed with the dynamic strategy. Table~\ref{tab_real_system_compare} presents the comparison to the three baselines. The results confirm that our system achieves both high-quality and efficient reconstruction of unknown objects in the real world.

\begin{table}[!t]
\centering
\resizebox{1.0\columnwidth}{!}{%
\begin{tabular}{!{\vrule width0.5pt}c!{\vrule width0.5pt}cccc!{\vrule width0.5pt}} 
\hline
Method & RV & VSC\,(\%) & MC\,(m) & RT\,(s)\\
\Xhline{1pt}
GMC~\cite{pan2023global}  & 12.22 ± 2.63 & 98.57 ± 1.36 & 2.92 ± 0.87 & 143.1 ± 31.7 \\
\cline{1-5}
MCMF+Mov~\cite{pan2022aglobal} & 12.77 ± 2.36 & 97.02 ± 3.04 & 1.44 ± 0.42 & 102.4 ± 18.9 \\
\cline{1-5}
SCVP~\cite{pan2022scvp} & \textbf{10.20} ± 0.95 & 95.88 ± 4.28 & \textbf{1.28} ± 0.15 & \textbf{82.2} ± 11.7 \\
\cline{1-5}
Ours & 12.93 ± 1.27 & $\mathbb{^{\star\star\star}}$\,\textbf{99.35} ± 0.29 & 1.73 ± 0.22 & 105.3 ± 12.6 \\
\Xhline{1pt}
\end{tabular}
}
\caption{Evaluation of view planning systems with the dynamic strategy in the real world: GMC and MCMF+Mov adopt the $m$-based stop criterion ($m=3$) with a threshold 0.4\%. Each value is the mean value and its standard deviation, computed over 60 tests. \textbf{RV}, \textbf{VSC}, \textbf{MC}, and \textbf{RT} stand for the number of required views, visible surface coverage, movement cost, and reconstruction time, respectively. $\mathbb{^{\star\star\star}}$ indicates a consistent significance at $p<0.001$ between our method and all baselines (\ie, several paired tests), as determined by both paired t-test and the Wilcoxon signed-rank test. As can be seen, our system achieves a significantly higher VSC than baselines, while also exhibiting substantial reductions in MC compared to the second-highest VSC method (GMC).
}
\label{tab_real_system_compare}
\vspace{-0.2cm}
\end{table}

\begin{table}[!t]
\centering
\resizebox{0.85\columnwidth}{!}{%
\begin{tabular}{!{\vrule width0.5pt}c!{\vrule width0.5pt}ccc!{\vrule width0.5pt}} 
\hline
Method & VSC\,(\%) & MC\,(m) & RT\,(s)\\
\Xhline{1pt}
GMC~\cite{pan2023global}  & 98.97 ± 0.60  & 3.12 ± 0.70  & 150.0 ± 23.5 \\
\cline{1-4}
MCMF+Mov~\cite{pan2022aglobal} & 97.52 ± 1.85 & \textbf{1.45} ± 0.36 & \textbf{104.2} ± 15.3 \\
\cline{1-4}
Ours & $\mathbb{^{\star\star\star}}$\,\textbf{99.35} ± 0.29 & 1.73 ± 0.22 & 105.3 ± 12.6 \\
\Xhline{1pt}
\end{tabular}
}
\caption{Evaluation of applying our stop criterion to NBV systems in the real world: Each value is the mean value and its standard deviation, computed over 60 tests. RV (the number of required views) of NBV systems is 12.93 ± 1.27, which is the same as our system. \textbf{VSC}, \textbf{MC}, and \textbf{RT} stand for visible surface coverage, movement cost, and reconstruction time, respectively. $\mathbb{^{\star\star\star}}$ indicates a consistent significance at $p<0.001$ between our method and all baselines (\ie, several paired tests), as determined by both paired t-test and the Wilcoxon signed-rank test. As can be seen, while our stop criterion markedly enhances NBV systems by increasing VSC and reducing standard deviation, our system still achieves a significantly higher VSC.
}
\label{tab_real_stop_criterion}
\vspace{-0.5cm}
\end{table}

\subsubsection{Stop Criterion Analysis}

In the case of NBV systems, adjusting the threshold of the stop criterion might require intervention from a human expert in the real world. Otherwise, it could stop prematurely or fail to stop altogether. In contrast, our system provides an auto-stop criterion that proves to be both sufficient and stable, as demonstrated in Table~\ref{tab_real_stop_criterion}.

\vspace{-0.2cm}

\subsection{Object Test Case Complexity Analysis} \label{secV.G}

To gain insight into how the provision of extra information enhances reconstruction quality, we analyze our systems in terms of object test case complexity. The complexity of a given object test case is influenced by several factors, including size, rotation, the difficulty of the initial view, and surface complexity. To quantify the complexity of a specific object test case, we employ the SCOP solution, as defined in Sec.~\ref{secIV.B}. The resulting number of covering views from SCOP serves as an indicator of complexity. This is intuitive, since the more complex the situation, the more views are required to cover the remaining object surfaces. Table~\ref{tab_object_complexity} presents an analysis of object test case complexity within our systems in simulation. From the results, we confirm that (1) the offline training on our long-tail multiview dataset enhances overall performance (when comparing 0-NBV+MA-SCVP to SCVP); and (2) the online incorporation of a single NBV improves performance, particularly in more complex cases (when comparing 1-NBV+MA-SCVP to 0-NBV+MA-SCVP).

\begin{figure}[!t]
\centering
\includegraphics[width=0.90\columnwidth]{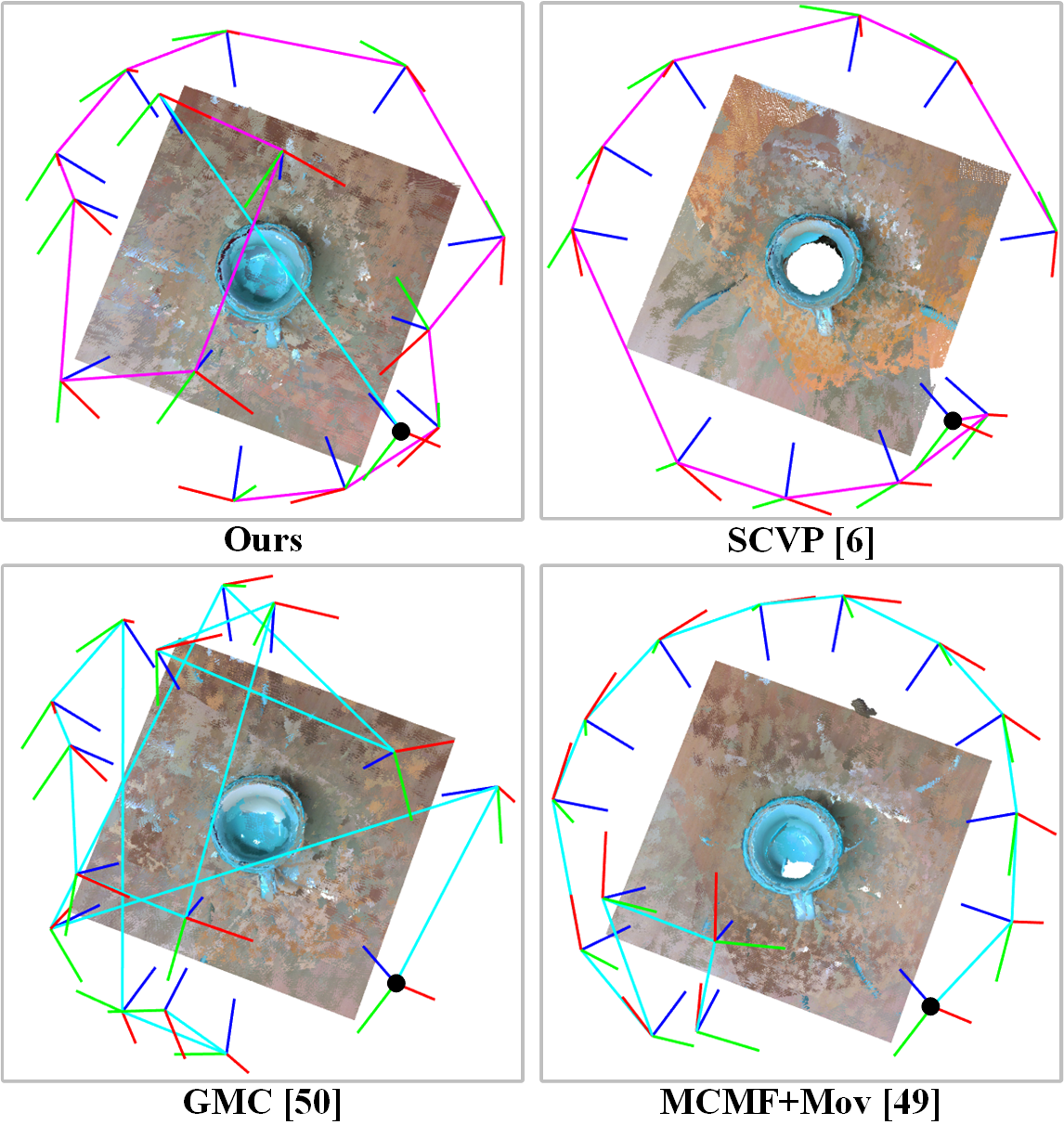}
\caption{Comparison of reconstruction process on a complex cup: reconstructed scenes and 3D models, local paths (cyan), global paths (purple), planned views (red-green-blue), and the same initial view (black circle). As can be seen, (1) SCVP fails to capture the missing middle surfaces, confirming the enhanced quality of complex cases in our system; and (2) our system reconstructed more surfaces (in comparison to MCMF+Mov) within shorter paths (in comparison to GMC).
} 
\label{fig_complex_cup}
\vspace{-0.5cm}
\end{figure}

To validate this finding beyond simulation, we conduct a comparison in the real world using a complex cup. This cup features continuous surfaces with large variations in curvature, requiring top views to observe the middle surfaces. As shown in Fig.~\ref{fig_complex_cup}, our system successfully plans top views to observe the missing surfaces, in contrast to SCVP. The full comparative reconstruction processes are illustrated in our demo video.

\begin{table}[!t]
\centering
\resizebox{1.0\columnwidth}{!}{%
\begin{tabular}{!{\vrule width0.5pt}c!{\vrule width0.5pt}c!{\vrule width0.5pt}c!{\vrule width0.5pt}c!{\vrule width0.5pt}} 
\hline
\makecell{Number of Covering Views\\(Number of Test Cases)} & \makecell{4-7\\(314)} & \makecell{8-12\\(1729)} & \makecell{13-16\\(957)} \\
\Xhline{1pt}
0-NBV+SCVP~\cite{pan2022scvp} & 99.67 ± 0.41 & 97.85 ± 2.75 & 91.83 ± 9.08 \\
\cline{1-4}
0-NBV+MA-SCVP & 99.95 ± 0.11 & 98.99 ± 1.73 & 95.87 ± 5.55 \\
\cline{1-4}
1-NBV+MA-SCVP\,(Ours) & 99.95 ± 0.14 & \textbf{99.01} ± 1.48 & \textbf{96.84} ± 3.80 \\
\cline{1-4}
2-NBV+MA-SCVP & \textbf{99.96} ± 0.06 & 98.99 ± 1.56 & 96.83 ± 4.02 \\
\Xhline{1pt}
\end{tabular}
}
\caption{Evaluation of object test case complexity in simulation involves grouping the number of covering views into three categories: simple (4-7), medium (8-12), and complex (13-16). Each value is the mean VSC value with its standard deviation, computed over the number of test cases in its category. VSC stands for visible surface coverage. As can be seen, 1-NBV+MA-SCVP achieves the highest VSC in medium and complex cases, while maintaining a similar VSC in simple cases. This indicates that providing extra information to one-shot networks contributes to improved performance.
}
\label{tab_object_complexity}
\vspace{-0.2cm}
\end{table}

\vspace{-0.2cm}

\section{Discussion} \label{secVI}

Although our system can achieve compelling results in experiments, there are still some important aspects left for future work. First, the VSC is only 96.84 ± 3.80 for those complex cases, as shown in Table~\ref{tab_object_complexity}. We observe that some objects in the simulation datasets have holes on their surfaces, \ie, discontinuous surfaces. For instance, the Cheff object (the last one in the bottom-right corner in Fig.~\ref{fig_object_models}) has two holes on its surfaces. However, this type of complexity falls outside the scope of this paper since the majority of real-world objects are solid inside. Specifically, this complexity arises from the need for specific views to observe surfaces inside the object. Predicting these specific views might pose a challenge for convolutional networks, given that the assumption of continuous and watertight surfaces is common in surface feature extraction~\cite{boulch2022poco}. In particular, predicting the number and the shape of holes in unknown areas is challenging without real perception. A possible solution is to detect these holes after the initial reconstruction using our system and then plan additional views to observe the inner surfaces along these holes.

Second, the candidate view space is pre-defined, and the network output is bound to a certain candidate view. This does not allow the system to drop some unreachable candidate views, such as due to obstacles and robot workspace. A possible solution is using a regression network~\cite{vasquez2021next} and training it with the state of unreachable views.

Third, we assume that there is only one object in the workspace. Therefore, reconstructing multiple objects or an object with multiple regions of interest requires manual separations. This might be improved by point-cloud segmentation methods~\cite{nguyen20133d} or the detection pipeline~\cite{zaenker2023graph}.

Fourth, although objects of any size can be reconstructed as long as the size does not exceed a maximum size bound, the system cannot handle larger objects such as chairs and large plants. A possible solution is to detect the object size in advance, which can be solved by some exploration methods~\cite{bircher2018receding}.

Finally, the bottom surfaces of an object cannot be reconstructed on a tabletop. Although such a model is sufficient for many applications such as grasping and phenotyping, it is better to have a watertight and enclosed 3D model. There are some solutions to address the lack of bottom surfaces. Wu \etal~\cite{wu2014quality} placed the object in the air with a stand. Krainin \etal~\cite{krainin2011autonomous} lifted the object with a robotic arm to scan the object from all directions. Another possible solution is to pick up and put down the object to observe the previous bottom surface.

\vspace{-0.2cm}

\section{Conclusion} \label{secVII}

In this work, we present a novel combined system that integrates one-shot view planning with a single next-best view through long-tail multiview sampling. Our long-tail sampling method addresses unbalanced importance among multiview inputs, presenting an efficient training dataset. The trained MA-SCVP network provides effective view planning with an auto-stop criterion learned from the set covering. Facing unknown environments, we demonstrate that both offline and online providing extra information improves surface coverage, especially in complex object cases. As a result, our system achieves both high-quality and high-efficiency reconstruction compared to state-of-the-art view planning systems. Real-world experiments demonstrate the robust generalization and deployment of our system.

\vspace{-0.2cm}

\section{Acknowledgements} \label{secVIII}
We thank Yiyun He for the fruitful discussions. 

\appendices

\vspace{-0.2cm}

\section{NBV Reconstruction and Sampling Space} \label{Appendix_NBVRsampling}

To generate the whole sampling space $C_{whole}$, we performed NBV reconstruction trails in the simulation as summarized in Algorithm~\ref{alg4}. The reconstruction needs the object mesh model set $O$, rotation set $R$, and candidate view space $V$ as input. It is worth mentioning that the candidate view space is pre-defined and our network output is bound to a certain candidate view. This implies the inputs to the network have lost the information of rotational invariance and symmetry in the hemispherical view space. Consequently, we incorporate $R$ and $V$ to enable the network to generalize across various object rotations and initial views during the reconstruction process. An object case $c_{obj}=(o,r)$ is an object $o \in O$ from 3D model datasets (as described in Sec.~\ref{secV.A}) with its rotation set $r \in R$ (line 2). The number $|O|$ of 3D models used for training is 40. We rotate the model around the Z+ (the vertically upward direction of the table) by a degree varying from $0^\circ$ to $360^\circ$ in steps of $45^\circ$, \ie, the number $|R|$ of the rotation is 8. The total number of reconstruction trails is $40 \times 8 \times 32 = 10,240$.

\begin{figure}[!t]
\vspace{-0.5cm}
\begin{algorithm}[H]
\caption{NBV Reconstruction with Ground Truth} 
\label{alg4}
\vspace{-0.1cm}
\begin{algorithmic}[1]
\REQUIRE $\mathit{Object\ Set\ O,\ Rotation\ Set\ R,\ View\ Space\ V}$
\FORALL{$o \in O\ and\ r \in R$}
    \STATE $c_{obj} \gets (o,r)$
    \FORALL{$v \in V$}
        \STATE $\mathbb{v} \gets VirtualImaging(c_{obj},v)$
        \STATE $U \gets U \cup \mathbb{v}$
    \ENDFOR
    \FORALL{$v \in V$}
        \STATE $U_{cover} \gets \mathbb{v}$
        \STATE $c_{view} \gets \mathit{BitInit(v)}$
        \STATE $C_{whole} \gets C_{whole} \cup \{(c_{obj},c_{view})\}$
        \STATE $\mathit{VSC} \gets \frac{|U_{cover}|}{|U|}$
        \WHILE {$\mathit{VSC} \neq 100\%$}
            \STATE $v^\ast \gets \arg \max_{v} |\mathbb{v} \setminus U_{cover}|$
            \STATE $U_{cover} \gets U_{cover} \cup \mathbb{v^\ast}$
            \STATE $n_{select} \gets \mathit{PopCount(c_{view})}$
            \STATE $NS(c_{obj},n_{select}) \gets \mathit{Update\ with\ \frac{|U_{cover}|}{|U|}-\mathit{VSC}}$
            \STATE $c_{view} \gets \mathit{BitUpdate(v^\ast)}$
            \STATE $C_{whole} \gets C_{whole} \cup \{(c_{obj},c_{view})\}$
            \STATE $\mathit{VSC} \gets \frac{|U_{cover}|}{|U|}$
        \ENDWHILE
    \ENDFOR
\ENDFOR
\RETURN $C_{whole}$
\end{algorithmic} 
\end{algorithm}
\vspace{-0.6cm}
\end{figure}

\begin{figure}[!t]
\centering
\includegraphics[width=0.70\columnwidth]{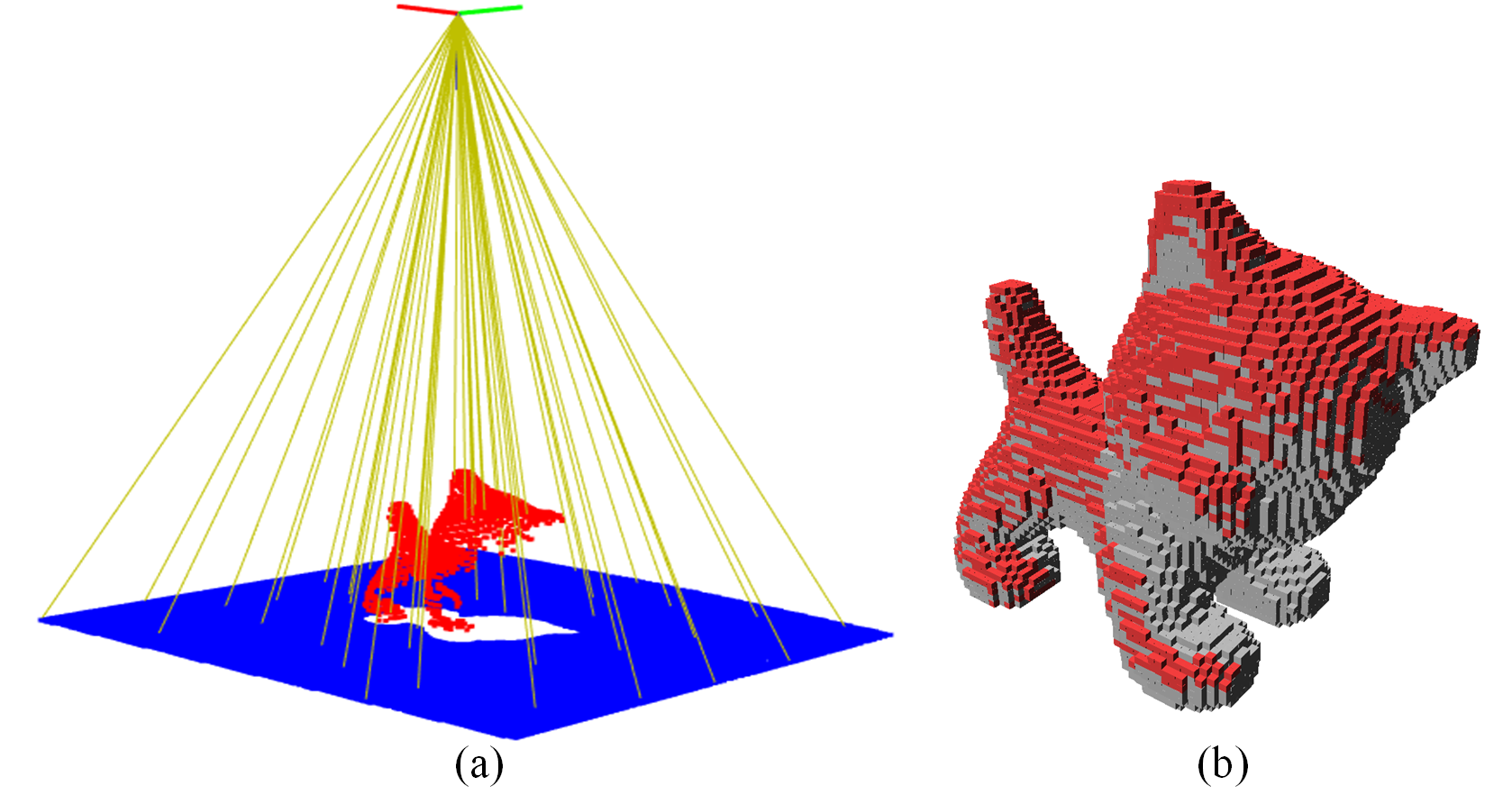}
\caption{
Virtual imaging: (a) Some rays of a view $v$ in point-cloud space (yellow). (b) The set $\mathbb{v}$ of object voxels (red) and uncovered voxels (gray) in the ground truth model.
} 
\label{fig_ray_casting}
\vspace{-0.5cm}
\end{figure}

Before the reconstruction trails, we perform virtual imaging for all candidate views on an object case $c_{obj}$ (lines 3-6). The set $\mathbb{v}$ of object surface voxels that can be observed from a view $v\in V$ is obtained by virtual imaging (line 4). Since the bottom and inner voxels are completely invisible, we generate a universe set $U=\bigcup_{v\in V}\mathbb{v}$ of all visible surface voxels as the ground truth (line 5). As shown in Fig.~\ref{fig_ray_casting}, virtual imaging in our simulated reconstruction is implemented with the ray-casting operation of OctoMap, which is common in voxel-based view planning~\cite{delmerico2018comparison}. We insert an object case into an OctoMap and get visible voxels in it from the cast rays of a given view. A ray is cast by the de-projecting pixels of the virtual image plane until it hits an occupied voxel. The object model is converted into a dense point cloud before insertion into the OctoMap to ensure that the object surface is closed. 

A reconstruction trail starts with an initial view from the candidate view space (lines 8-11). $U_{cover}$ stands for the current number of covered object voxels (lines 8 and 14), $c_{view}$ is initialized and updated by the view bit (lines 9 and 17), and $\mathit{VSC}$ stands for current visible surface coverage (lines 11 and 19). The NBV can be easily determined from the view that observes the most uncovered surface voxels in the ground truth (line 13). $\mathit{PopCount}$ calculates the number of 1 bit, \ie, the number of visited views (line 15). $NS$ is a hash table that maps the paired object case and the number $n_{select}$ of selected views to the total averaged surface coverage gain. $\mathit{Update}$ will replace the old average value with the new one calculated from the current surface coverage gain $|U_{cover}|/|U|-\mathit{VSC}$ (line 16). The input cases for each iteration are stored to generate the whole sampling space $C_{whole}$ (lines 10 and 18). Traditional sampling methods~\cite{mendoza2020supervised,zeng2020pc} can be regarded as modifying $v\in V$ in line 7 to $v\in V_{ran}$, where $V_{ran}$ is a random subset of $V$.

\vspace{-0.3cm}

\section{Supplementary Simulation Experiments}

\vspace{-0.1cm}

\subsection{Ablation Study on MA-SCVP Parameters} \label{Appendix_hyperparameter}

We determine the MA-SCVP parameter values based on the highest $F_1$ score, following the strategy outlined in SCVP~\cite{pan2022scvp}. Table~\ref{tab_parameter} reports the ablation study. The results indicate that $\lambda=2.0$ and $\gamma=0.5$ are valid for the MA-SCVP network.

\begin{table}[!t]
\centering
\resizebox{0.67\columnwidth}{!}{%
\begin{tabular}{!{\vrule width0.5pt}c!{\vrule width0.5pt}c!{\vrule width0.5pt}c!{\vrule width0.5pt}c!{\vrule width0.5pt}c!{\vrule width0.5pt}c!{\vrule width0.5pt}} 
\Xhline{1pt}

$\lambda$\,($\gamma=0.5$) & 1.0 & 1.5 & \textbf{2.0} & 2.5 & 3.0 \\
\Xhline{1pt}
Recall\,(\%) & 62.18 & 67.49 & 83.95 & 86.45 & \textbf{86.71} \\
\cline{1-6}
Precision\,(\%) & \textbf{87.29} & 80.00 & 85.76 & 64.95 & 62.44 \\
\cline{1-6}
$F_1$\,(\%) & 72.63 & 73.21 & \textbf{84.84} & 74.17 & 72.60 \\
\Xhline{1pt}

$\gamma$\,($\lambda=2.0$) & 0.1 & 0.3 & \textbf{0.5} & 0.7 & 0.9 \\
\Xhline{1pt}
Recall\,(\%) & \textbf{100.00} & 99.71 & 83.95 & 69.76 & 39.57 \\
\cline{1-6}
Precision\,(\%) & 49.90 & 66.97 & 85.76 & 88.51 & \textbf{90.33} \\
\cline{1-6}
$F_1$\,(\%) & 66.58 & 80.12 & \textbf{84.84} & 78.02 & 55.04 \\
\Xhline{1pt}

\end{tabular}
}
\caption{Ablation study on MA-SCVP parameters: Each value is computed on the validation set. As can be seen, $\lambda=2.0$ and $\gamma=0.5$ leads to the highest $F_1$ score.
}
\label{tab_parameter}
\vspace{-0.2cm}
\end{table}

\begin{table}[!t]
\centering
\resizebox{1.0\columnwidth}{!}{%
\begin{tabular}{!{\vrule width0.5pt}c!{\vrule width0.5pt}c!{\vrule width0.5pt}ccc!{\vrule width0.5pt}} 
\hline
1-NBV Submodule & First Iteration VSC\,(\%) & RV & VSC\,(\%) & MC\,(m)\\
\Xhline{1pt}
PCNBV~\cite{zeng2020pc} & \textbf{71.90} ± 8.63 & 13.01 ± 1.46 & \textbf{98.41} ± 2.67 & 3.07 ± 0.24 \\
\cline{1-5}
NBVNet~\cite{mendoza2020supervised} & 71.01 ± 9.12 & \textbf{12.95} ± 1.46 & 98.32 ± 2.84 & 3.06 ± 0.23 \\
\cline{1-5}
RSE~\cite{delmerico2018comparison} & 68.29 ± 10.37 & 13.07 ± 1.43 & 98.16 ± 3.56 & 3.09 ± 0.24 \\
\cline{1-5}
APORA~\cite{daudelin2017adaptable} & 64.67 ± 9.34 & 13.63 ± 1.39 & 98.38 ± 3.23 & 3.08 ± 0.21 \\
\cline{1-5}
MCMF~\cite{pan2022aglobal} & 64.12 ± 10.09 & 13.20 ± 1.47 & 98.31 ± 3.19 & 3.01 ± 0.25 \\
\cline{1-5}
GMC~\cite{pan2023global} & 64.19 ± 10.07 & 13.19 ± 1.49 & 98.30 ± 3.25 & 3.01 ± 0.25 \\
\cline{1-5}
Random & 58.23 ± 10.84 & 13.41 ± 1.45 & 98.22 ± 3.48 & \textbf{2.93} ± 0.26 \\
\Xhline{1pt}
\end{tabular}
}
\caption{Evaluation of different NBV submodules within our 1-NBV+MA-SCVP pipeline: Each value is the mean value and its standard deviation, computed over 3,000 tests. \textbf{RV}, \textbf{VSC}, and \textbf{MC} stand for the number of required views, visible surface coverage, and movement cost, respectively. As can be seen, since PCNBV achieves the highest VSC of the first NBV iteration, our 1-NBV+MA-SCVP pipeline achieves the highest VSC and lowest standard deviation when PCNBV is used as the NBV submodule.
}
\label{tab_NBV_module}
\vspace{-0.5cm}
\end{table}

\vspace{-0.3cm}

\subsection{NBV Submodule Analysis} \label{Appendix_NBVsubmodule}

As discussed in Sec.~\ref{secIII.B}, our system allows for integration with various NBV methods. To examine the correlation between NBV method performance and our MA-SCVP network, we replace PCNBV with alternative NBV methods, including a random method, in our 1-NBV+MA-SCVP pipeline. Table~\ref{tab_NBV_module} reports the results of different NBV submodules. From the results, we confirm that (1) VSC increases when any NBV method is applied in our 1-NBV+MA-SCVP pipeline, compared to 0-NBV+MA-SCVP (98.10 ± 3.73); and (2) there is a generally positive correlation between the performance of the NBV method and the performance of our MA-SCVP network, \ie, the higher the performance of the NBV method, the higher the performance of our MA-SCVP network (higher VSC with lower standard deviation).

\vspace{-0.3cm}

\subsection{Stop Criterion Analysis} \label{Appendix_stopcriterion}

To further validate that our 1-NBV+MA-SCVP provides an efficient stop criterion, we conduct two additional experiments. First, we allow NBV baselines to run until they reach the same number of views as our system. Table~\ref{tab_stop_criterion} reports the results of applying our stop criterion to different NBV methods. From the results, we observe that (1)~our stop criterion provides a sufficient number of required views; (2)~MCMF+Mov achieves a similar VSC as ours but requires 40\% more MC, highlighting the efficiency of our system; and (3)~MCMF and GMC achieve a slightly higher VSC as ours but demand 110\% more MC, which is not worthwhile.

\begin{table}[!t]
\centering
\resizebox{0.65\columnwidth}{!}{%
\begin{tabular}{!{\vrule width0.5pt}c!{\vrule width0.5pt}cc!{\vrule width0.5pt}} 
\hline
Method & VSC\,(\%) & MC\,(m)\\
\Xhline{1pt}
PCNBV~\cite{zeng2020pc} & 97.51 ± 2.09          & 6.14 ± 1.03 \\
\cline{1-3} 
NBVNet~\cite{mendoza2020supervised} & 96.79 ± 4.02          & 6.68 ± 1.15 \\
\cline{1-3} 
RSE~\cite{delmerico2018comparison} & 97.89 ± 1.88          & 4.92 ± 0.84 \\
\cline{1-3} 
APORA~\cite{daudelin2017adaptable} & 98.00 ± 2.00          & 6.12 ± 1.00 \\
\cline{1-3} 
MCMF~\cite{pan2022aglobal} & \textbf{98.70} ± 1.27 & 6.57 ± 0.97 \\
\cline{1-3} 
GMC~\cite{pan2023global} & 98.65 ± 1.21 & 6.42 ± 0.93 \\
\cline{1-3} 
RSE+Mov~\cite{delmerico2018comparison} & 97.06 ± 2.67          & 4.22 ± 0.87 \\
\cline{1-3} 
APORA+Mov~\cite{daudelin2017adaptable} & 97.96 ± 2.01          & 5.95 ± 0.95 \\
\cline{1-3} 
MCMF+Mov~\cite{pan2022aglobal} & 98.42 ± 1.58 & 4.43 ± 0.93 \\
\cline{1-3}  
GMC+Mov~\cite{pan2023global} & 98.26 ± 1.50          & 4.85 ± 0.94 \\
\cline{1-3} 
1-NBV+MASCVP\,(Ours) & 98.41 ± 2.67          & \textbf{3.07} ± 0.24 \\
\Xhline{1pt}
\end{tabular}
}
\caption{Evaluation of applying our stop criterion to different NBV methods: Each value is the mean value and its standard deviation, computed over 3,000 tests. RV (the number of required views) of all methods is 13.01 ± 1.46, which is the same as our system. \textbf{VSC}, and \textbf{MC} stand for the number of required views, visible surface coverage, and movement cost, respectively. As can be seen, our stop criterion enhances the VSC for all NBV methods but also demands a much higher MC.
}
\label{tab_stop_criterion}
\vspace{-0.2cm}
\end{table}

\begin{table}[!t]
\centering
\resizebox{0.86\columnwidth}{!}{%
\begin{tabular}{!{\vrule width0.5pt}c!{\vrule width0.5pt}ccc!{\vrule width0.5pt}} 
\hline
Method & RV & VSC\,(\%) & MC\,(m)\\
\Xhline{1pt}
\multirow{6}{*}{\makecell{Random\\with Global Path Planning}} & \textbf{12}             & 95.96 ± 3.11          & \textbf{2.64} ± 0.18 \\
\cline{2-4}
 & 13             & 96.55 ± 2.76          & 2.79 ± 0.17 \\
\cline{2-4}
 & 14             & 97.02 ± 2.47          & 2.94 ± 0.16 \\
\cline{2-4}
 & 15             & 97.42 ± 2.20          & 3.08 ± 0.15 \\
\cline{2-4}
 & 16             & 97.77 ± 1.98          & 3.22 ± 0.15 \\
\cline{2-4}
 & 13.01 ± 1.46 & 96.42 ± 3.00 & 2.79 ± 0.28 \\
\cline{1-4}
1-NBV+MASCVP\,(Ours) & 13.01 ± 1.46 & $\mathbb{^{\star\star\star}}$\,\textbf{98.41} ± 2.67          & 3.07 ± 0.24\\
\Xhline{1pt}
\end{tabular}
}
\caption{Evaluation of random methods with a fixed number of views and our stop criterion: Each value is the mean value and its standard deviation, computed over 3,000 tests. \textbf{RV}, \textbf{VSC}, and \textbf{MC} stand for the number of required views, visible surface coverage, and movement cost, respectively. $\mathbb{^{\star\star\star}}$ indicates a consistent significance at $p<0.001$ between our method and all random methods (\ie, several paired tests), as determined by both paired t-test and the Wilcoxon signed-rank test. As can be seen, our 1-NBV+MASCVP achieves a significantly higher VSC against random methods, indicating our effective stop criterion.
}
\label{tab_random_oneshot}
\vspace{-0.5cm}
\end{table}

Second, we randomly select a fixed number of views and utilize our global path planning for comparison. The fixed number of views is determined by prior knowledge regarding how many NBVs are required to fully cover an object, specifically 12-16, as shown in Fig.~\ref{fig_multiview_features}. It is important to note that these values are derived from training objects, and acquiring this knowledge in advance for unknown objects in real testing is unrealistic. Table~\ref{tab_random_oneshot} reports the results of random methods. From the results, we observe that (1)~random methods require more RV and MC to achieve a comparable VSC than our method; and (2)~the random method with our stop criterion confirms that our method not only predicts a sufficient RV but also effectively anticipates the view configuration, \ie, which candidate views are needed to be selected.

\vspace{-0.2cm}

\bibliographystyle{IEEEtran}
\bibliography{refs}

\begin{thebibliography}{10}
\providecommand{\url}[1]{#1}
\csname url@samestyle\endcsname
\providecommand{\newblock}{\relax}
\providecommand{\bibinfo}[2]{#2}
\providecommand{\BIBentrySTDinterwordspacing}{\spaceskip=0pt\relax}
\providecommand{\BIBentryALTinterwordstretchfactor}{4}
\providecommand{\BIBentryALTinterwordspacing}{\spaceskip=\fontdimen2\font plus
\BIBentryALTinterwordstretchfactor\fontdimen3\font minus \fontdimen4\font\relax}
\providecommand{\BIBforeignlanguage}[2]{{%
\expandafter\ifx\csname l@#1\endcsname\relax
\typeout{** WARNING: IEEEtran.bst: No hyphenation pattern has been}%
\typeout{** loaded for the language `#1'. Using the pattern for}%
\typeout{** the default language instead.}%
\else
\language=\csname l@#1\endcsname
\fi
#2}}
\providecommand{\BIBdecl}{\relax}
\BIBdecl

\bibitem{zaenker2021viewpoint}
T.~Zaenker, C.~Smitt, C.~McCool, and M.~Bennewitz, ``Viewpoint planning for fruit size and position estimation,'' in \emph{2021 IEEE/RSJ International Conference on Intelligent Robots and Systems (IROS)}.\hskip 1em plus 0.5em minus 0.4em\relax IEEE, 2021, pp. 3271--3277.

\bibitem{dengler2023viewpoint}
N.~Dengler, S.~Pan, V.~Kalagaturu, R.~Menon, M.~Dawood, and M.~Bennewitz, ``Viewpoint push planning for mapping of unknown confined spaces,'' \emph{2023 IEEE/RSJ International Conference on Intelligent Robots and Systems (IROS)}, 2023.

\bibitem{breyer2022closed}
M.~Breyer, L.~Ott, R.~Siegwart, and J.~J. Chung, ``Closed-loop next-best-view planning for target-driven grasping,'' in \emph{2022 IEEE/RSJ International Conference on Intelligent Robots and Systems (IROS)}.\hskip 1em plus 0.5em minus 0.4em\relax IEEE, 2022, pp. 1411--1416.

\bibitem{chen2011active}
S.~Chen, Y.~Li, and N.~M. Kwok, ``Active vision in robotic systems: A survey of recent developments,'' \emph{The International Journal of Robotics Research}, vol.~30, no.~11, pp. 1343--1377, 2011.

\bibitem{zeng2020view}
R.~Zeng, Y.~Wen, W.~Zhao, and Y.-J. Liu, ``View planning in robot active vision: A survey of systems, algorithms, and applications,'' \emph{Computational Visual Media}, pp. 1--21, 2020.

\bibitem{pan2022scvp}
S.~Pan, H.~Hu, and H.~Wei, ``Scvp: Learning one-shot view planning via set covering for unknown object reconstruction,'' \emph{IEEE Robotics and Automation Letters}, vol.~7, no.~2, pp. 1463--1470, 2022.

\bibitem{delmerico2018comparison}
J.~Delmerico, S.~Isler, R.~Sabzevari, and D.~Scaramuzza, ``A comparison of volumetric information gain metrics for active 3d object reconstruction,'' \emph{Autonomous Robots}, vol.~42, no.~2, pp. 197--208, 2018.

\bibitem{zeng2020pc}
R.~Zeng, W.~Zhao, and Y.-J. Liu, ``Pc-nbv: A point cloud based deep network for efficient next best view planning,'' in \emph{2020 IEEE/RSJ International Conference on Intelligent Robots and Systems (IROS)}.\hskip 1em plus 0.5em minus 0.4em\relax Las Vegas, NV, USA: IEEE, 2020, pp. 7050--7057.

\bibitem{zhang2021deep}
Y.~Zhang, B.~Kang, B.~Hooi, S.~Yan, and J.~Feng, ``Deep long-tailed learning: A survey,'' \emph{IEEE Transactions on Pattern Analysis and Machine Intelligence}, vol.~45, no.~9, pp. 10\,795--10\,816, 2023.

\bibitem{mendoza2020supervised}
M.~Mendoza, J.~I. Vasquez-Gomez, H.~Taud, L.~E. Sucar, and C.~Reta, ``Supervised learning of the next-best-view for 3d object reconstruction,'' \emph{Pattern Recognition Letters}, vol. 133, pp. 224--231, 2020.

\bibitem{connolly1985determination}
C.~Connolly, ``The determination of next best views,'' in \emph{1985 IEEE international conference on robotics and automation}, vol.~2.\hskip 1em plus 0.5em minus 0.4em\relax St. Louis, MO, USA: IEEE, 1985, pp. 432--435.

\bibitem{tarabanis1995survey}
K.~A. Tarabanis, P.~K. Allen, and R.~Y. Tsai, ``A survey of sensor planning in computer vision,'' \emph{IEEE transactions on Robotics and Automation}, vol.~11, no.~1, pp. 86--104, 1995.

\bibitem{scott2003view}
W.~R. Scott, G.~Roth, and J.-F. Rivest, ``View planning for automated three-dimensional object reconstruction and inspection,'' \emph{ACM Computing Surveys (CSUR)}, vol.~35, no.~1, pp. 64--96, 2003.

\bibitem{maboudi2022review}
M.~Maboudi, M.~Homaei, S.~Song, S.~Malihi, M.~Saadatseresht, and M.~Gerke, ``A review on viewpoints and path planning for uav-based 3d reconstruction,'' \emph{IEEE Journal of Selected Topics in Applied Earth Observations and Remote Sensing}, 2023.

\bibitem{bircher2018receding}
A.~Bircher, M.~Kamel, K.~Alexis, H.~Oleynikova, and R.~Siegwart, ``Receding horizon path planning for 3d exploration and surface inspection,'' \emph{Autonomous Robots}, vol.~42, pp. 291--306, 2018.

\bibitem{monica2019humanoid}
R.~Monica, J.~Aleotti, and D.~Piccinini, ``Humanoid robot next best view planning under occlusions using body movement primitives,'' in \emph{2019 IEEE/RSJ International Conference on Intelligent Robots and Systems (IROS)}.\hskip 1em plus 0.5em minus 0.4em\relax IEEE, 2019, pp. 2493--2500.

\bibitem{song2020online}
S.~Song, D.~Kim, and S.~Jo, ``Online coverage and inspection planning for 3d modeling,'' \emph{Autonomous Robots}, vol.~44, no.~8, pp. 1431--1450, 2020.

\bibitem{song2021view}
S.~Song, D.~Kim, and S.~Choi, ``View path planning via online multiview stereo for 3-d modeling of large-scale structures,'' \emph{IEEE Transactions on Robotics}, 2021.

\bibitem{vasquez2013hierarchical}
J.~I. Vasquez-Gomez, L.~E. Sucar, and R.~Murrieta-Cid, ``Hierarchical ray tracing for fast volumetric next-best-view planning,'' in \emph{2013 International Conference on Computer and Robot Vision}.\hskip 1em plus 0.5em minus 0.4em\relax Regina, CANADA: IEEE, 2013, pp. 181--187.

\bibitem{aleotti2014global}
J.~Aleotti, D.~L. Rizzini, R.~Monica, and S.~Caselli, ``Global registration of mid-range 3d observations and short range next best views,'' in \emph{2014 IEEE/RSJ International Conference on Intelligent Robots and Systems}.\hskip 1em plus 0.5em minus 0.4em\relax IEEE, 2014, pp. 3668--3675.

\bibitem{monica2018contour}
R.~Monica and J.~Aleotti, ``Contour-based next-best view planning from point cloud segmentation of unknown objects,'' \emph{Autonomous Robots}, vol.~42, pp. 443--458, 2018.

\bibitem{peng2020viewpoints}
W.~Peng, Y.~Wang, Z.~Miao, M.~Feng, and Y.~Tang, ``Viewpoints planning for active 3-d reconstruction of profiled blades using estimated occupancy probabilities (eop),'' \emph{IEEE Transactions on Industrial Electronics}, vol.~68, no.~5, pp. 4109--4119, 2020.

\bibitem{burusa2022attention}
A.~K. Burusa, E.~J. van Henten, and G.~Kootstra, ``Attention-driven next-best-view planning for efficient reconstruction of plants and targeted plant parts,'' \emph{Biosystems Engineering}, vol. 246, pp. 248--262, 2024.

\bibitem{wong1999next}
L.~M. Wong, C.~Dumont, and M.~A. Abidi, ``Next best view system in a 3d object modeling task,'' in \emph{Proceedings 1999 IEEE International Symposium on Computational Intelligence in Robotics and Automation. CIRA'99 (Cat. No. 99EX375)}.\hskip 1em plus 0.5em minus 0.4em\relax IEEE, 1999, pp. 306--311.

\bibitem{zaenker2021combining}
T.~Zaenker, C.~Lehnert, C.~McCool, and M.~Bennewitz, ``Combining local and global viewpoint planning for fruit coverage,'' in \emph{2021 European Conference on Mobile Robots (ECMR)}.\hskip 1em plus 0.5em minus 0.4em\relax IEEE, 2021, pp. 1--7.

\bibitem{wu2019plant}
C.~Wu, R.~Zeng, J.~Pan, C.~C. Wang, and Y.-J. Liu, ``Plant phenotyping by deep-learning-based planner for multi-robots,'' \emph{IEEE Robotics and Automation Letters}, vol.~4, no.~4, pp. 3113--3120, 2019.

\bibitem{lauri2020multi}
M.~Lauri, J.~Pajarinen, J.~Peters, and S.~Frintrop, ``Multi-sensor next-best-view planning as matroid-constrained submodular maximization,'' \emph{IEEE Robotics and Automation Letters}, vol.~5, no.~4, pp. 5323--5330, 2020.

\bibitem{keselman2017intel}
L.~Keselman, J.~Iselin~Woodfill, A.~Grunnet-Jepsen, and A.~Bhowmik, ``Intel realsense stereoscopic depth cameras,'' in \emph{Proceedings of the IEEE conference on computer vision and pattern recognition workshops}, 2017, pp. 1--10.

\bibitem{border2018surface}
R.~Border, J.~D. Gammell, and P.~Newman, ``Surface edge explorer (see): Planning next best views directly from 3d observations,'' in \emph{2018 IEEE International Conference on Robotics and Automation (ICRA)}.\hskip 1em plus 0.5em minus 0.4em\relax IEEE, 2018, pp. 6116--6123.

\bibitem{wu20153d}
Z.~Wu, S.~Song, A.~Khosla, F.~Yu, L.~Zhang, X.~Tang, and J.~Xiao, ``3d shapenets: A deep representation for volumetric shapes,'' in \emph{Proceedings of the IEEE conference on computer vision and pattern recognition}, 2015, pp. 1912--1920.

\bibitem{pito1999solution}
R.~Pito, ``A solution to the next best view problem for automated surface acquisition,'' \emph{IEEE Transactions on pattern analysis and machine intelligence}, vol.~21, no.~10, pp. 1016--1030, 1999.

\bibitem{wu2014quality}
S.~Wu, W.~Sun, P.~Long, H.~Huang, D.~Cohen-Or, M.~Gong, O.~Deussen, and B.~Chen, ``Quality-driven poisson-guided autoscanning,'' \emph{ACM Transactions on Graphics}, vol.~33, no.~6, 2014.

\bibitem{hornung2013octomap}
A.~Hornung, K.~M. Wurm, M.~Bennewitz, C.~Stachniss, and W.~Burgard, ``Octomap: An efficient probabilistic 3d mapping framework based on octrees,'' \emph{Autonomous robots}, vol.~34, no.~3, pp. 189--206, 2013.

\bibitem{massios1998best}
N.~A. Massios, R.~B. Fisher \emph{et~al.}, \emph{A best next view selection algorithm incorporating a quality criterion}.\hskip 1em plus 0.5em minus 0.4em\relax Citeseer, 1998, vol.~2.

\bibitem{menon2022viewpoint}
R.~Menon, T.~Zaenker, and M.~Bennewitz, ``Nbv-sc: Next best view planning based on shape completion for fruit mapping and reconstruction,'' \emph{2023 IEEE/RSJ International Conference on Intelligent Robots and Systems (IROS)}, 2023.

\bibitem{monica2018surfel}
R.~Monica and J.~Aleotti, ``Surfel-based next best view planning,'' \emph{IEEE Robotics and Automation Letters}, vol.~3, no.~4, pp. 3324--3331, 2018.

\bibitem{mildenhall2021nerf}
B.~Mildenhall, P.~P. Srinivasan, M.~Tancik, J.~T. Barron, R.~Ramamoorthi, and R.~Ng, ``Nerf: Representing scenes as neural radiance fields for view synthesis,'' \emph{Communications of the ACM}, vol.~65, no.~1, pp. 99--106, 2021.

\bibitem{pan2022activenerf}
X.~Pan, Z.~Lai, S.~Song, and G.~Huang, ``Activenerf: Learning where to see with uncertainty estimation,'' in \emph{European Conference on Computer Vision}.\hskip 1em plus 0.5em minus 0.4em\relax Springer, 2022, pp. 230--246.

\bibitem{ran2023neurar}
Y.~Ran, J.~Zeng, S.~He, J.~Chen, L.~Li, Y.~Chen, G.~Lee, and Q.~Ye, ``Neurar: Neural uncertainty for autonomous 3d reconstruction with implicit neural representations,'' \emph{IEEE Robotics and Automation Letters}, 2023.

\bibitem{jin2023neu}
L.~Jin, X.~Chen, J.~R{\"u}ckin, and M.~Popovi{\'c}, ``Neu-nbv: Next best view planning using uncertainty estimation in image-based neural rendering,'' in \emph{2023 IEEE/RSJ International Conference on Intelligent Robots and Systems (IROS)}, 2023.

\bibitem{border2020proactive}
R.~Border and J.~D. Gammell, ``Proactive estimation of occlusions and scene coverage for planning next best views in an unstructured representation,'' in \emph{2020 IEEE/RSJ International Conference on Intelligent Robots and Systems (IROS)}.\hskip 1em plus 0.5em minus 0.4em\relax IEEE, 2020, pp. 4219--4226.

\bibitem{border2022surface}
R.~{}Border and J.~D. Gammell, ``The surface edge explorer (see): A measurement-direct approach to next best view planning,'' \emph{The International Journal of Robotics Research}, p. 02783649241230098, 2024.

\bibitem{kriegel2011surface}
S.~Kriegel, T.~Bodenm{\"u}ller, M.~Suppa, and G.~Hirzinger, ``A surface-based next-best-view approach for automated 3d model completion of unknown objects,'' in \emph{2011 IEEE International Conference on Robotics and Automation}.\hskip 1em plus 0.5em minus 0.4em\relax IEEE, 2011, pp. 4869--4874.

\bibitem{lee2020automatic}
I.~D. Lee, J.~H. Seo, Y.~M. Kim, J.~Choi, S.~Han, and B.~Yoo, ``Automatic pose generation for robotic 3-d scanning of mechanical parts,'' \emph{IEEE Transactions on Robotics}, vol.~36, no.~4, pp. 1219--1238, 2020.

\bibitem{krainin2011autonomous}
M.~Krainin, B.~Curless, and D.~Fox, ``Autonomous generation of complete 3d object models using next best view manipulation planning,'' in \emph{2011 IEEE International Conference on Robotics and Automation}.\hskip 1em plus 0.5em minus 0.4em\relax Shanghai, PEOPLES R CHINA: IEEE, 2011, pp. 5031--5037.

\bibitem{vasquez2014volumetric}
J.~I. Vasquez-Gomez, L.~E. Sucar, R.~Murrieta-Cid, and E.~Lopez-Damian, ``Volumetric next-best-view planning for 3d object reconstruction with positioning error,'' \emph{International Journal of Advanced Robotic Systems}, vol.~11, no.~10, p. 159, 2014.

\bibitem{vasquez2017view}
J.~I. Vasquez-Gomez, L.~E. Sucar, and R.~Murrieta-Cid, ``View/state planning for three-dimensional object reconstruction under uncertainty,'' \emph{Autonomous Robots}, vol.~41, no.~1, pp. 89--109, 2017.

\bibitem{daudelin2017adaptable}
J.~Daudelin and M.~Campbell, ``An adaptable, probabilistic, next-best view algorithm for reconstruction of unknown 3-d objects,'' \emph{IEEE Robotics and Automation Letters}, vol.~2, no.~3, pp. 1540--1547, 2017.

\bibitem{pan2022aglobal}
S.~Pan and H.~Wei, ``A global max-flow-based multi-resolution next-best-view method for reconstruction of 3d unknown objects,'' \emph{IEEE Robotics and Automation Letters}, vol.~7, no.~2, pp. 714--721, 2022.

\bibitem{pan2023global}
S.~{}Pan and H.~Wei, ``A global generalized maximum coverage-based solution to the non-model-based view planning problem for object reconstruction,'' \emph{Computer Vision and Image Understanding}, vol. 226, p. 103585, 2023.

\bibitem{lee2022uncertainty}
S.~Lee, L.~Chen, J.~Wang, A.~Liniger, S.~Kumar, and F.~Yu, ``Uncertainty guided policy for active robotic 3d reconstruction using neural radiance fields,'' \emph{IEEE Robotics and Automation Letters}, vol.~7, no.~4, pp. 12\,070--12\,077, 2022.

\bibitem{sunderhauf2023density}
N.~S{\"u}nderhauf, J.~Abou-Chakra, and D.~Miller, ``Density-aware nerf ensembles: Quantifying predictive uncertainty in neural radiance fields,'' in \emph{2023 IEEE International Conference on Robotics and Automation (ICRA)}.\hskip 1em plus 0.5em minus 0.4em\relax IEEE, 2023, pp. 9370--9376.

\bibitem{peralta2020next}
D.~Peralta, J.~Casimiro, A.~M. Nilles, J.~A. Aguilar, R.~Atienza, and R.~Cajote, ``Next-best view policy for 3d reconstruction,'' in \emph{2020 European Conference on Computer Vision}.\hskip 1em plus 0.5em minus 0.4em\relax Glasgow, UK: Springer, 2020, pp. 558--573.

\bibitem{zeng2022deep}
X.~Zeng, T.~Zaenker, and M.~Bennewitz, ``Deep reinforcement learning for next-best-view planning in agricultural applications,'' in \emph{2022 International Conference on Robotics and Automation (ICRA)}.\hskip 1em plus 0.5em minus 0.4em\relax IEEE, 2022, pp. 2323--2329.

\bibitem{zhou2018voxelnet}
Y.~Zhou and O.~Tuzel, ``Voxelnet: End-to-end learning for point cloud based 3d object detection,'' in \emph{Proceedings of the IEEE conference on computer vision and pattern recognition}, 2018, pp. 4490--4499.

\bibitem{vasquez2021next}
J.~I. Vasquez-Gomez, D.~Troncoso, I.~Becerra, E.~Sucar, and R.~Murrieta-Cid, ``Next-best-view regression using a 3d convolutional neural network,'' \emph{Machine Vision and Applications}, vol.~32, no.~2, pp. 1--14, 2021.

\bibitem{qi2017pointnet}
C.~R. Qi, H.~Su, K.~Mo, and L.~J. Guibas, ``Pointnet: Deep learning on point sets for 3d classification and segmentation,'' in \emph{Proceedings of the IEEE conference on computer vision and pattern recognition}, 2017, pp. 652--660.

\bibitem{han2022double}
Y.~Han, I.~H. Zhan, W.~Zhao, and Y.-J. Liu, ``A double branch next-best-view network and novel robot system for active object reconstruction,'' in \emph{2022 International Conference on Robotics and Automation (ICRA)}.\hskip 1em plus 0.5em minus 0.4em\relax IEEE, 2022, pp. 7306--7312.

\bibitem{monica2021probabilistic}
R.~Monica and J.~Aleotti, ``A probabilistic next best view planner for depth cameras based on deep learning,'' \emph{IEEE Robotics and Automation Letters}, vol.~6, no.~2, pp. 3529--3536, 2021.

\bibitem{peuzin2021survey}
M.~Peuzin-Jubert, A.~Polette, D.~Nozais, J.-L. Mari, and J.-P. Pernot, ``Survey on the view planning problem for reverse engineering and automated control applications,'' \emph{Computer-Aided Design}, vol. 141, p. 103094, 2021.

\bibitem{kaba2017reinforcement}
M.~D. Kaba, M.~G. Uzunbas, and S.~N. Lim, ``A reinforcement learning approach to the view planning problem,'' in \emph{2017 IEEE Conference on Computer Vision and Pattern Recognition (CVPR)}.\hskip 1em plus 0.5em minus 0.4em\relax Honolulu, HI, USA: IEEE, 2017, pp. 5094--5102.

\bibitem{hepp2018plan3d}
B.~Hepp, M.~Nie{\ss}ner, and O.~Hilliges, ``Plan3d: Viewpoint and trajectory optimization for aerial multi-view stereo reconstruction,'' \emph{ACM Transactions on Graphics (TOG)}, vol.~38, no.~1, pp. 1--17, 2018.

\bibitem{jing2018computational}
W.~Jing, C.~F. Goh, M.~Rajaraman, F.~Gao, S.~Park, Y.~Liu, and K.~Shimada, ``A computational framework for automatic online path generation of robotic inspection tasks via coverage planning and reinforcement learning,'' \emph{IEEE Access}, vol.~6, pp. 54\,854--54\,864, 2018.

\bibitem{patidar2016survey}
V.~Patidar and R.~Tiwari, ``Survey of robotic arm and parameters,'' in \emph{2016 International conference on computer communication and informatics (ICCCI)}.\hskip 1em plus 0.5em minus 0.4em\relax IEEE, 2016, pp. 1--6.

\bibitem{chitta2016moveit}
S.~Chitta, ``Moveit!: an introduction,'' \emph{Robot Operating System (ROS) The Complete Reference (Volume 1)}, pp. 3--27, 2016.

\bibitem{enebuse2021comparative}
I.~Enebuse, M.~Foo, B.~K.~K. Ibrahim, H.~Ahmed, F.~Supmak, and O.~S. Eyobu, ``A comparative review of hand-eye calibration techniques for vision guided robots,'' \emph{IEEE Access}, 2021.

\bibitem{han2017review}
X.-F. Han, J.~S. Jin, M.-J. Wang, W.~Jiang, L.~Gao, and L.~Xiao, ``A review of algorithms for filtering the 3d point cloud,'' \emph{Signal Processing: Image Communication}, vol.~57, pp. 103--112, 2017.

\bibitem{conway2013sphere}
J.~H. Conway and N.~J.~A. Sloane, \emph{Sphere packings, lattices and groups}.\hskip 1em plus 0.5em minus 0.4em\relax Springer Science \& Business Media, 2013, vol. 290.

\bibitem{held1962dynamic}
M.~Held and R.~M. Karp, ``A dynamic programming approach to sequencing problems,'' \emph{Journal of the Society for Industrial and Applied mathematics}, vol.~10, no.~1, pp. 196--210, 1962.

\bibitem{vazirani2013approximation}
V.~V. Vazirani, ``Approximation algorithms (springer science \& business media,'' 2013.

\bibitem{mittelmann2018latest}
H.~Mittelmann, ``Latest benchmark results,'' in \emph{Proceedings of the INFORMS Annual Conference, Phoenix, AZ, USA}, 2018, pp. 4--7.

\bibitem{ioffe2015batch}
S.~Ioffe and C.~Szegedy, ``Batch normalization: Accelerating deep network training by reducing internal covariate shift,'' in \emph{International conference on machine learning}.\hskip 1em plus 0.5em minus 0.4em\relax pmlr, 2015, pp. 448--456.

\bibitem{maas2013rectifier}
A.~L. Maas, A.~Y. Hannun, A.~Y. Ng \emph{et~al.}, ``Rectifier nonlinearities improve neural network acoustic models,'' in \emph{Proc. icml}, vol.~30, no.~1.\hskip 1em plus 0.5em minus 0.4em\relax Atlanta, Georgia, USA, 2013, p.~3.

\bibitem{herrera2016multilabel}
F.~Herrera, F.~Charte, A.~J. Rivera, M.~J. Del~Jesus, F.~Herrera, F.~Charte, A.~J. Rivera, and M.~J. del Jesus, \emph{Multilabel classification}.\hskip 1em plus 0.5em minus 0.4em\relax Springer, 2016.

\bibitem{he2016deep}
K.~He, X.~Zhang, S.~Ren, and J.~Sun, ``Deep residual learning for image recognition,'' in \emph{Proceedings of the IEEE conference on computer vision and pattern recognition}, 2016, pp. 770--778.

\bibitem{krishnamurthy1996fitting}
V.~Krishnamurthy and M.~Levoy, ``Fitting smooth surfaces to dense polygon meshes,'' in \emph{Proceedings of the 23rd annual conference on Computer graphics and interactive techniques}, 1996, pp. 313--324.

\bibitem{hinterstoisser2012model}
S.~Hinterstoisser, V.~Lepetit, S.~Ilic, S.~Holzer, G.~Bradski, K.~Konolige, and N.~Navab, ``Model based training, detection and pose estimation of texture-less 3d objects in heavily cluttered scenes,'' in \emph{Asian conference on computer vision}.\hskip 1em plus 0.5em minus 0.4em\relax Springer, 2012, pp. 548--562.

\bibitem{kaskman2019homebreweddb}
R.~Kaskman, S.~Zakharov, I.~Shugurov, and S.~Ilic, ``Homebreweddb: Rgb-d dataset for 6d pose estimation of 3d objects,'' in \emph{Proceedings of the IEEE/CVF International Conference on Computer Vision Workshops}, 2019, pp. 0--0.

\bibitem{rodola2013scale}
E.~Rodola, A.~Albarelli, F.~Bergamasco, and A.~Torsello, ``A scale independent selection process for 3d object recognition in cluttered scenes,'' \emph{International journal of computer vision}, vol. 102, pp. 129--145, 2013.

\bibitem{wang2019octreenet}
F.~Wang, Y.~Zhuang, H.~Gu, and H.~Hu, ``Octreenet: A novel sparse 3-d convolutional neural network for real-time 3-d outdoor scene analysis,'' \emph{IEEE Transactions on Automation Science and Engineering}, vol.~17, no.~2, pp. 735--747, 2019.

\bibitem{kingma2014adam}
D.~P. Kingma and J.~Ba, ``Adam: {A} method for stochastic optimization,'' in \emph{Proceedings of the 3rd International Conference on Learning Representations, {ICLR}}, 2015.

\bibitem{yervilla2019optimal}
H.~Yervilla-Herrera, J.~I. Vasquez-Gomez, R.~Murrieta-Cid, I.~Becerra, and L.~E. Sucar, ``Optimal motion planning and stopping test for 3-d object reconstruction,'' \emph{Intelligent Service Robotics}, vol.~12, pp. 103--123, 2019.

\bibitem{yervilla2022bayesian}
H.~Yervilla-Herrera, I.~Becerra, R.~Murrieta-Cid, L.~E. Sucar, and E.~F. Morales, ``Bayesian probabilistic stopping test and asymptotic shortest time trajectories for object reconstruction with a mobile manipulator robot,'' \emph{Journal of Intelligent \& Robotic Systems}, vol. 105, no.~4, p.~82, 2022.

\bibitem{quigley2009ros}
M.~Quigley, K.~Conley, B.~Gerkey, J.~Faust, T.~Foote, J.~Leibs, R.~Wheeler, A.~Y. Ng \emph{et~al.}, ``Ros: an open-source robot operating system,'' in \emph{ICRA workshop on open source software}, vol.~3, no. 3.2.\hskip 1em plus 0.5em minus 0.4em\relax Kobe, Japan, 2009, p.~5.

\bibitem{boulch2022poco}
A.~Boulch and R.~Marlet, ``Poco: Point convolution for surface reconstruction,'' in \emph{Proceedings of the IEEE/CVF Conference on Computer Vision and Pattern Recognition}, 2022, pp. 6302--6314.

\bibitem{nguyen20133d}
A.~Nguyen and B.~Le, ``3d point cloud segmentation: A survey,'' in \emph{2013 6th IEEE conference on robotics, automation and mechatronics (RAM)}.\hskip 1em plus 0.5em minus 0.4em\relax IEEE, 2013, pp. 225--230.

\bibitem{zaenker2023graph}
T.~Zaenker, J.~R{\"u}ckin, R.~Menon, M.~Popovi{\'c}, and M.~Bennewitz, ``Graph-based view motion planning for fruit detection,'' \emph{2023 IEEE/RSJ International Conference on Intelligent Robots and Systems (IROS)}, 2023.

\end{thebibliography}

% \vspace{-14cm}
\begin{IEEEbiography}[{\includegraphics[width=1in,height=1.25in,clip,keepaspectratio]{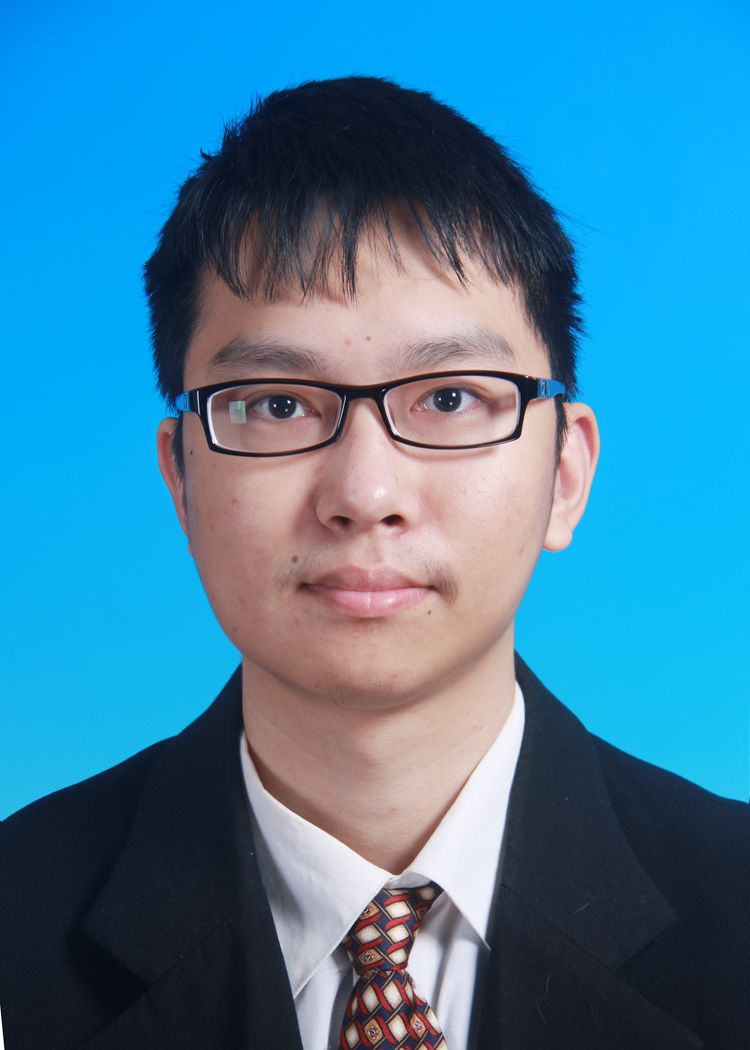}}]{Sicong Pan} is a Ph.D. student in Computer Science at the Humanoid Robots Lab headed by Prof. Maren Bennewitz at the University of Bonn, Germany. He obtained his B.Eng. degree in Computer Science \& Technology from Shanghai University, China in 2018 and received his M.Sc. degree in Computer Science from Fudan University, China in 2022. His current research interests include active and interactive perception, viewpoint planning, and 3D reconstruction.
\end{IEEEbiography}

% \vspace{-14cm}
\begin{IEEEbiography}[{\includegraphics[width=1in,height=1.25in,clip,keepaspectratio]{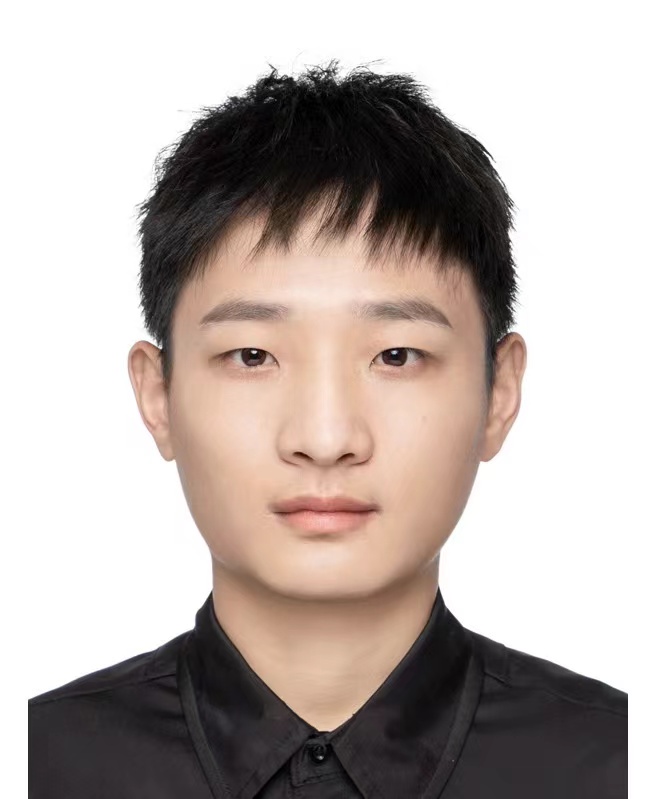}}]{Hao Hu} will start pursuing the Ph.D. degree in Computer Science at the Robot Perception and Learning Lab headed by Jun.-Prof. Hermann Blum at the University of Bonn, Germany in 2025. He obtained his B.Eng. degree in Bioengineering from Nanjing Agricultural University, China in 2019 and received the M.Sc. degree in Computer Science from Fudan University, China, in 2023. His current research interests include robotic vision, 3D reconstruction, AI-generated content, and large language models.
\end{IEEEbiography}

% \vspace{-14cm}
\begin{IEEEbiography}[{\includegraphics[width=1in,height=1.25in,clip,keepaspectratio]{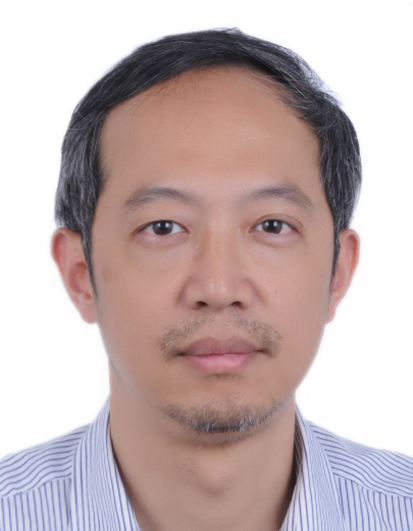}}]{Hui Wei} received the Ph.D. degree from the Department of Computer Science, Beijing University of Aeronautics and Astronautics, China in 1998. From 1998 to 2000, he was a Post-Doctoral Fellow with the Department of Computer Science and the Institute of Artificial Intelligence, Zhejiang University, China. Since November 2000, he has been with the Department of Computer Science and Engineering, Fudan University, China. His research interests include artificial intelligence and cognitive science.
\end{IEEEbiography}

% \vspace{-14cm}
\begin{IEEEbiography}[{\includegraphics[width=1in,height=1.25in,clip,keepaspectratio]{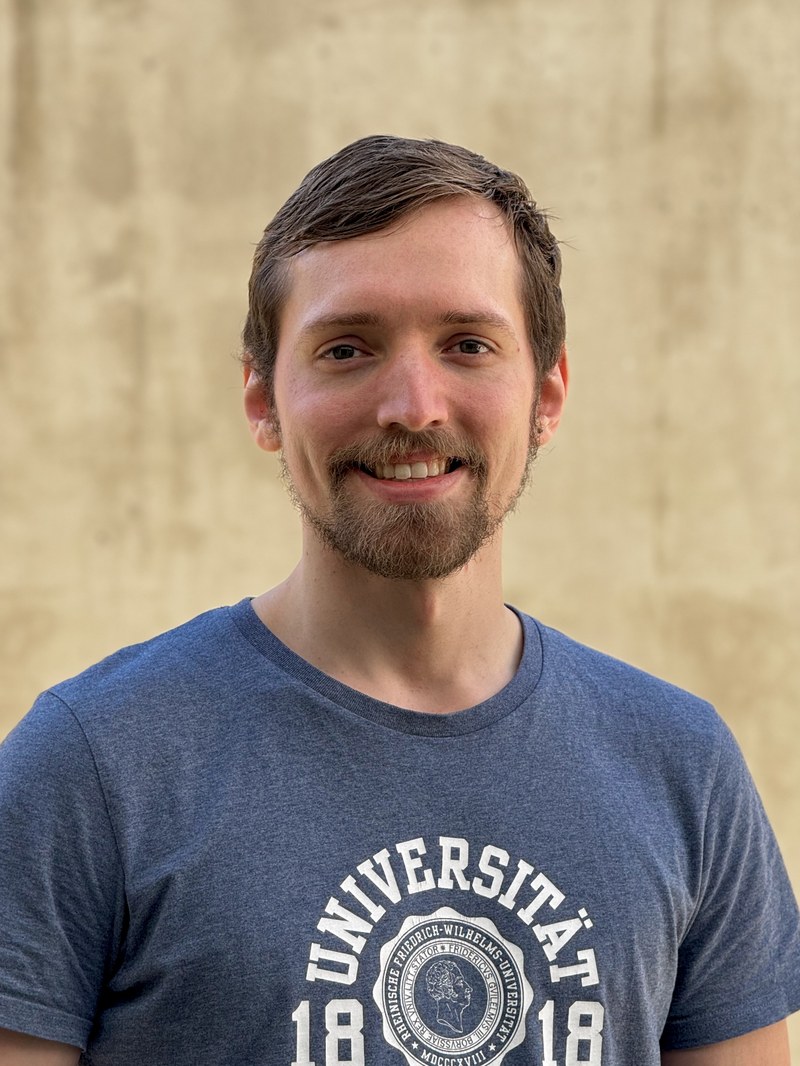}}]{Nils Dengler} is a Ph.D. student in Computer Science at the Humanoid Robots Lab headed by Prof. Maren Bennewitz at the University of Bonn, Germany. He obtained his B.Sc. and M.Sc. degree in Computer Science from the University of Bonn, Germany in
2018 and 2021. His current research interests include active and interactive perception, viewpoint planning, and non-prehensile manipulation.
\end{IEEEbiography}

% \vspace{-14cm}
\begin{IEEEbiography}[{\includegraphics[width=1in,height=1.25in,clip,keepaspectratio]{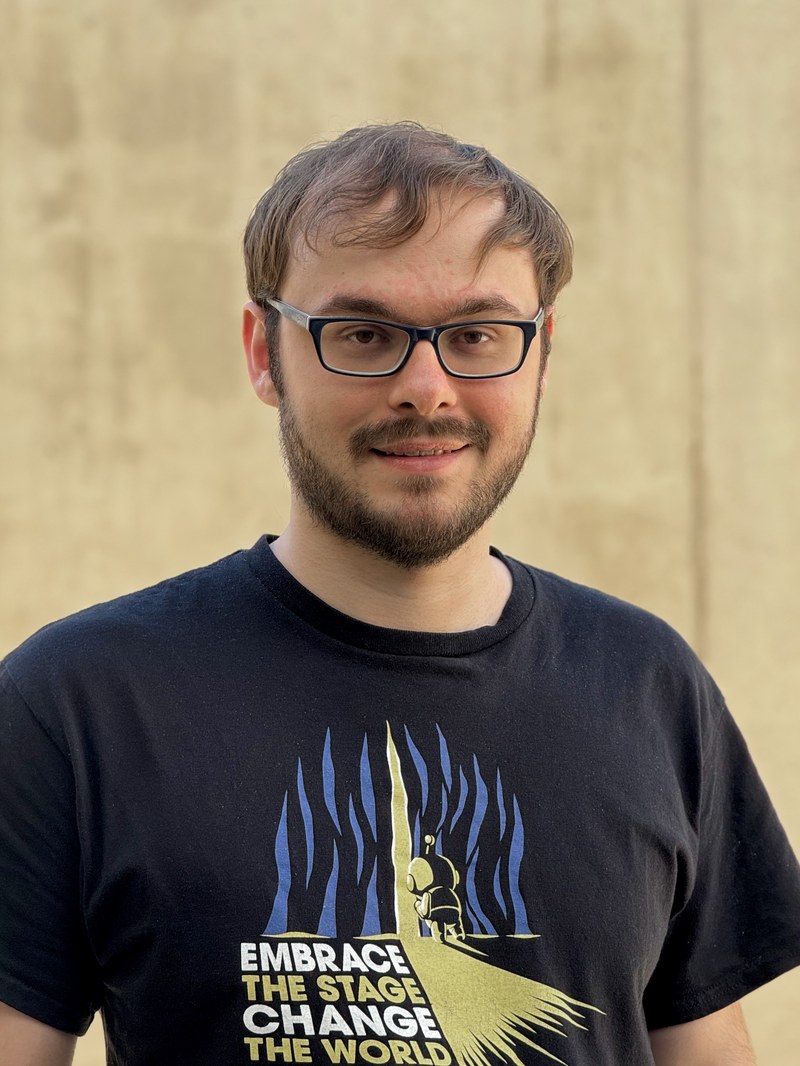}}]{Tobias Zaenker} is a Ph.D. student in Computer Science at the Humanoid Robots Lab headed by Prof. Maren Bennewitz at the University of Bonn, Germany. He obtained his B.Sc. degree in Aerospace Computer Science from the University of Würzburg, Germany in 2017 and received his M.Sc. degree in Space Science and Technology from Aalto University, Finland in 2019. His current research interests include viewpoint planning, fruit detection, and semantic mapping.
\end{IEEEbiography}

% \vspace{-14cm}
\begin{IEEEbiography}[{\includegraphics[width=1in,height=1.25in,clip,keepaspectratio]{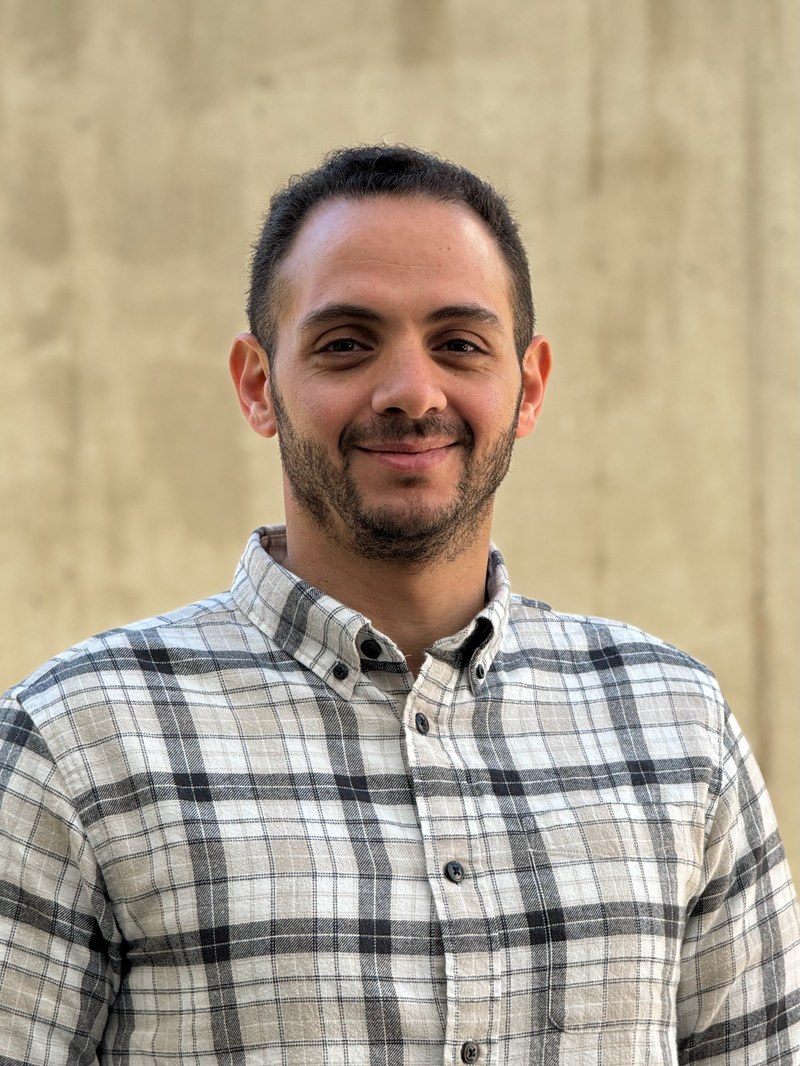}}]{Murad Dawood} is a Ph.D. student in Computer Science at the Humanoid Robots Lab headed by Prof. Maren Bennewitz at the University of Bonn, Germany. He obtained his B.Sc. degree in Mechatronics Engineering from Ain Shams University, Egypt in 2013 and received his M.Sc. degree in Mechatronics Engineering from Ain Shams University, Egypt in 2021. His current research interests include model predictive control, multi-agent reinforcement learning, and safe reinforcement learning in robotics.
\end{IEEEbiography}

\vspace{-15.3cm}
\begin{IEEEbiography}[{\includegraphics[width=1in,height=1.25in,clip,keepaspectratio]{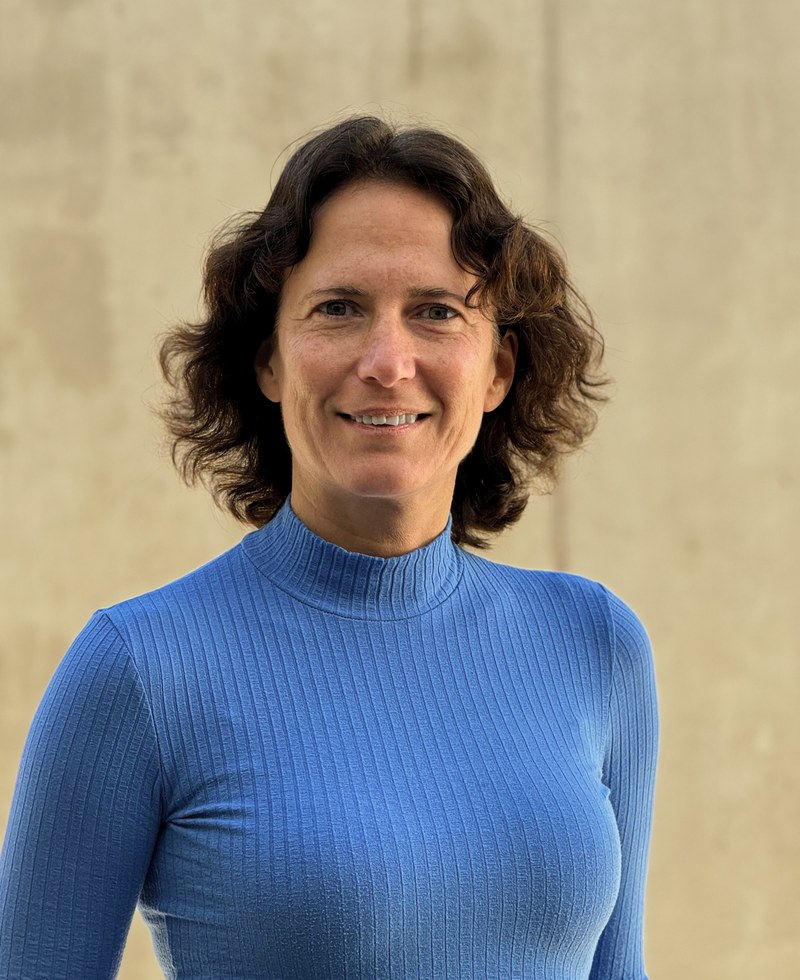}}]{Maren Bennewitz} is a full professor at the University of Bonn, Germany and head of the Humanoid Robots Lab. She is additionally with the Lamarr Institute for Machine Learning and Artificial Intelligence, a member of the executive board and steering committee of the Cluster of Excellence PhenoRob, and a founding member and steering committee member of the Center for Robotics, Germany. Her current research interests include robot navigation, active perception, intelligent manipulation, and personalized human robot-interaction.
\end{IEEEbiography}

\end{document}